\documentclass{article}

     \PassOptionsToPackage{numbers, compress}{natbib}

     \usepackage[final]{RumiGAN_NIPS2020}

\usepackage[utf8]{inputenc} 
\usepackage[T1]{fontenc}    
\usepackage{hyperref}       
\usepackage{url}            
\usepackage{booktabs}       
\usepackage{amsfonts}       
\usepackage{nicefrac}       
\usepackage{microtype}      

\usepackage{microtype}
\usepackage{graphicx}
\usepackage{rotating} 
\usepackage{subfigure}
\usepackage{booktabs} 
\usepackage{floatrow} 
\usepackage{multirow}

\usepackage{amsmath,amssymb}
\usepackage{amsthm}
\usepackage{amsfonts}
\usepackage{physics} 
\usepackage{bm} 
\usepackage{hyperref}
\usepackage{url}
\usepackage{booktabs}
\usepackage{graphicx} 
\usepackage{caption}

\DeclareMathOperator{\E}{\mathbb{E}}
\newcommand{\ds}{\displaystyle}
\newcommand{\loss}{\mathcal{L}}

\newcommand{\pd}{p_d}
\newcommand{\pdp}{p_d^+}
\newcommand{\pdn}{p_d^-}
\newcommand{\bp}{b^+}
\newcommand{\bn}{b^-}
\newcommand{\alphap}{\alpha^+}
\newcommand{\alphan}{\alpha^-}
\newcommand{\betap}{\beta^+}
\newcommand{\betan}{\beta^-}
\newcommand{\gammap}{\gamma^+}
\newcommand{\gamman}{\gamma^-}

\newcommand{\mcalX}{\mathcal{X}}

\newcommand{\pg}{p_g}

\newcommand{\x}{\bm{x}}

\DeclareMathOperator*{\argmax}{arg\,max}
\DeclareMathOperator*{\argmin}{arg\,min}

\newcommand{\fracpartial}[2]{\frac{\partial #1}{\partial  #2}}
\newcommand{\qedwhite}{\hfill \ensuremath{\Box}} 

\newtheorem{theorem}{Theorem}[section]

\newtheorem{lemma}[theorem]{Lemma}
\usepackage{hyperref}

\usepackage{algorithm}
\usepackage{algorithmic}

\usepackage{array}
\newcolumntype{P}[1]{>{\centering\arraybackslash}p{#1}}

\title{Teaching a GAN What Not to Learn}

\author{%
  Siddarth Asokan\thanks{Corresponding Author} \\
  Robert Bosch Center for Cyber-Physical Systems\\
  Indian Institute of Science\\
  Bangalore, India \\
  \texttt{siddartha@iisc.ac.in} \\
  \And Chandra Sekhar Seelamantula \\
  Department of Electrical Engineering\\
  Indian Institute of Science\\
  Bangalore, India \\
  \texttt{css@iisc.ac.in} \\
}

\begin{document}

\maketitle


\begin{abstract}
Generative adversarial networks (GANs) were originally envisioned as unsupervised generative models that learn to follow a target distribution. Variants such as conditional GANs, auxiliary-classifier GANs (ACGANs) project GANs on to supervised and semi-supervised learning frameworks by providing labelled data and using multi-class discriminators. In this paper, we approach the supervised GAN problem from a different perspective, one that is motivated by the philosophy of the famous Persian poet Rumi who said, \textit{``The art of knowing is knowing what to ignore.''} In the GAN framework, we not only provide the GAN \textit{positive} data that it must learn to model, but also present it with so-called \textit{negative} samples that it must learn to avoid --- we call this {\it the Rumi framework}. This formulation allows the discriminator to represent the underlying target distribution better by learning to penalize generated samples that are undesirable --- we show that this capability accelerates the learning process of the generator. We present a reformulation of the standard GAN (SGAN) and least-squares GAN (LSGAN) within the Rumi setting. The advantage of the reformulation is demonstrated by means of experiments conducted on MNIST, Fashion MNIST, CelebA, and CIFAR-10 datasets. Finally, we consider an application of the proposed formulation to address the important problem of learning an under-represented class in an unbalanced dataset. The Rumi approach results in substantially lower FID scores than the standard GAN frameworks while possessing better generalization capability.
\end{abstract}

\section{Introduction and Related Work}
Generative adversarial networks (GANs), originally proposed by Goodfellow {\it et al.}~\cite{SGAN14}, are unsupervised deep learning machines, wherein a generator learns to mimic a target data distribution, whereas the discriminator attempts to distinguish between real data and the samples coming from the generator.  The original GAN optimization and subsequent flavors such as the least-squares GAN (LSGAN) \cite{LSGAN} and \(f\)-GANs could be viewed as performing divergence minimization, where the optimal generator is the minimizer of a chosen divergence metric between the generated and true data distributions.
\par

The generator in the standard GAN formulation transforms input noise \( \bm{z} \sim p_Z \), typically a standard multivariate Gaussian, to the output \( G(\bm{z}) \) with distribution \( \pg(\x) \). The target data is sampled from an underlying distribution \( \pd(\x)\). The discriminator \(D(\x)\) predicts the probability of its input coming from \(\pd\). This is formulated as a min-max game between the generator and the discriminator:
\begin{align*}
\min_{\pg} \max_{D(\x)}\quad \E_{\x \sim \pd} \left[ \log D(\x)\right] + \E_{\x \sim p_g} \left[ \log(1 - D(\x))\right],
\end{align*}
where the optimal discriminator \( D^*(\x) = \frac{\pd}{\pd +\pg} \) was shown to be the one that measures the odds of a sample coming from the data distribution, and the optimal generator was shown to be the minimizer of the Jensen-Shannon divergence between \(\pd\) and \(\pg\). In the LSGAN formulation \cite{LSGAN}, the discriminator executes a regression task, i.e., to assign a class label \(a\) to the generated samples, and a class label \(b\) to the real ones. The generator learns to confuse the discriminator by creating samples that get classified by the discriminator as belonging to another category labelled \(c\). Mathematically stated, the LSGAN optimization comprises the following problems:
\begin{align*}
\min_{D(\x)} \quad &\E_{\x \sim \pd} \left[  (D(\x) - b)^2 \right] + \E_{\x \sim p_g} \left[ (D(\x) - a)^2\right],~\text{and} \\ 
\min_{\pg} \quad &\E_{\x \sim \pd} \left[  (D(\x) - c)^2 \right] + \E_{\x \sim p_g} \left[ (D(\x) - c)^2\right].
\end{align*}
The optimal generator in this case is the minimizer of the Pearson-\(\chi^2\) divergence between $2p_g$ and the sum of $p_d$ and $p_g$. 

Conditional GANs (CGANs)~\cite{cGAN} are a supervised extension of GANs and require labelled input data. The objective in CGANs is to control the class of the generated sample. Conditional GANs augment the inputs to both the generator and the discriminator with a one-hot encoding of the class label. Subsequent works have improved upon the class-conditional image generation in CGANs by introducing the label information into the hidden layers of the discriminator~\cite{Text2Im16} or by means of an inner-product with the class label introduced into the pre-final layer of the discriminator (CGAN-PD)~\cite{CGANPD18}. Conditional GANs and their variants have found applications in text-to-image synthesis \cite{Text2Im16,StackPP18} and image-to-image translation tasks \cite{MMPix2Pix18,Pix2Pix17,DivPix2Pix18,ModeSeeking19}. They have also been modified to include multi-class discriminators~\cite{VACGAN18}, and perform auxiliary classification (ACGAN)~\cite{ACGAN17} to improve the performance of the conditional generator. Twin auxiliary classifier GANs (TACGANs)~\cite{TACGAN19} improve the multi-class performance of ACGANs by introducing a data balancing term in the generator loss. Balancing GANs (BAGANs)~\cite{BAGAN18} perform data balancing by deploying an autoencoder to initialize the ACGANs. Both CGANs and ACGANs have been successful in medical image data augmentation applications~\cite{GANLivers18,ColabFiltGANs19}, where a converged generator is used to output samples from an under-represented class. \par
Another line of work involves splitting data into positive and negative samples for discriminative learning. In metric learning~\cite{ContrastiveLoss06,TripletLoss10,SimilarityMetric05,DistanceMetricLearning09} and representation learning~\cite{ContrastiveCNN14,ContrastiveRepLearning18}, the relative distances between samples are used to train a neural network. The contrastive loss~\cite{ContrastiveLoss06} compares pairs of samples, and assigns positive weights to similar/desired pairs and negative weights to dissimilar/undesired ones. As an extension, in triplet loss~\cite{TripletLoss10}, the distance of the target from the positive class is minimized, and that from the negative class is maximized. \par
In complementary CGAN (CCGAN)~\cite{CCGAN20}, a CGAN is trained on data where each sample is associated with a \textit{complementary label}, which is a yes/no tag, corresponding to a randomly selected class. Generative positive-unlabelled learning (GenPU)~\cite{GenPU18} involves training semi-supervised GANs with a mixture of positive, negative, and unlabelled samples, with the goal of obtaining a classifier that separates the positive unlabelled samples from the negative ones. Unlike the standard GAN, where the optimal generator is of ultimate interest, in CCGAN and GenPU, the end product is the optimized discriminator. \par


\section{Our Contribution}
\begin{figure*}[t!]
\begin{center}
  \begin{tabular}[b]{c|c|c}
    \includegraphics[height=.4\linewidth]{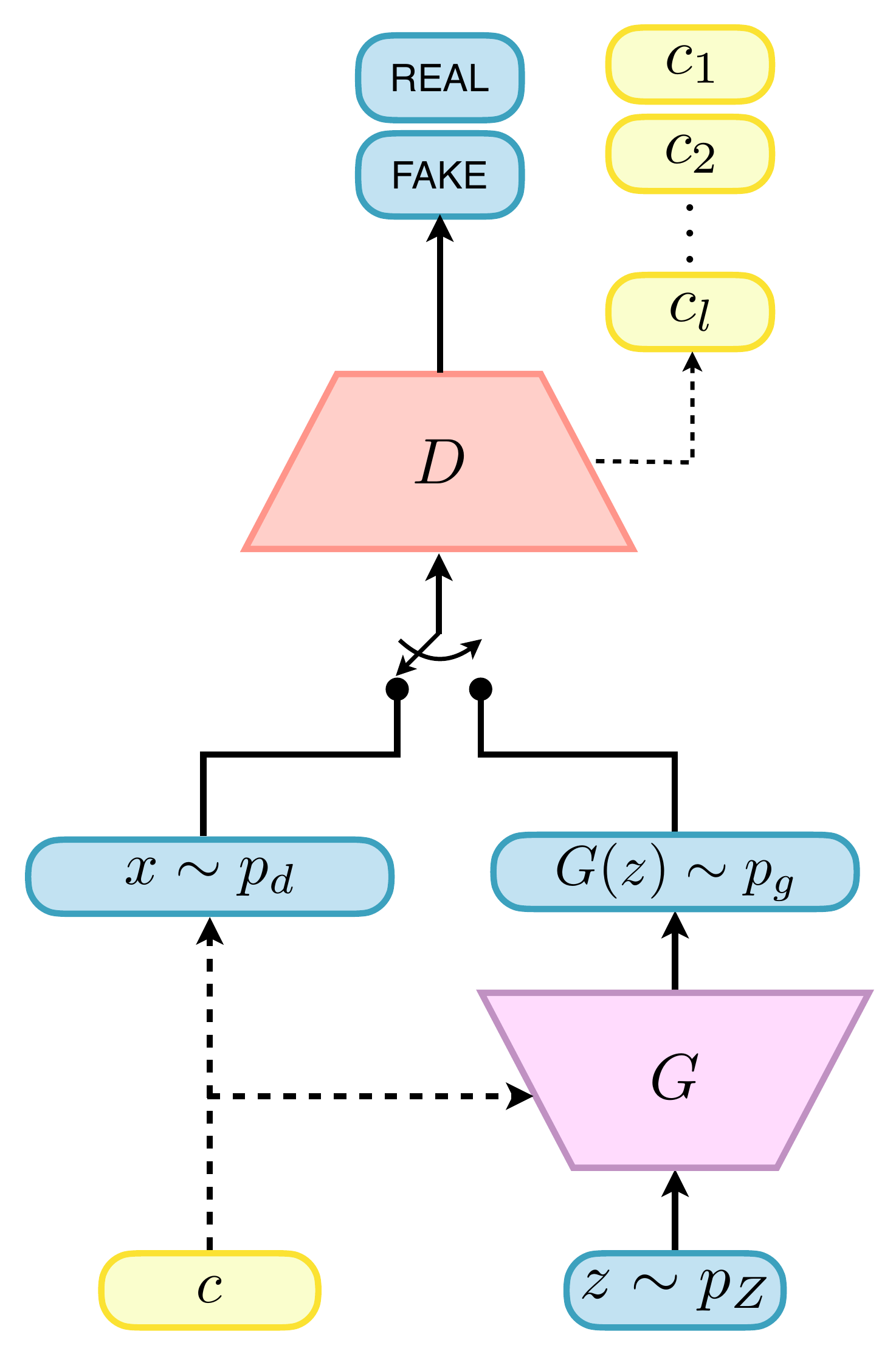} & \includegraphics[height=.4\linewidth]{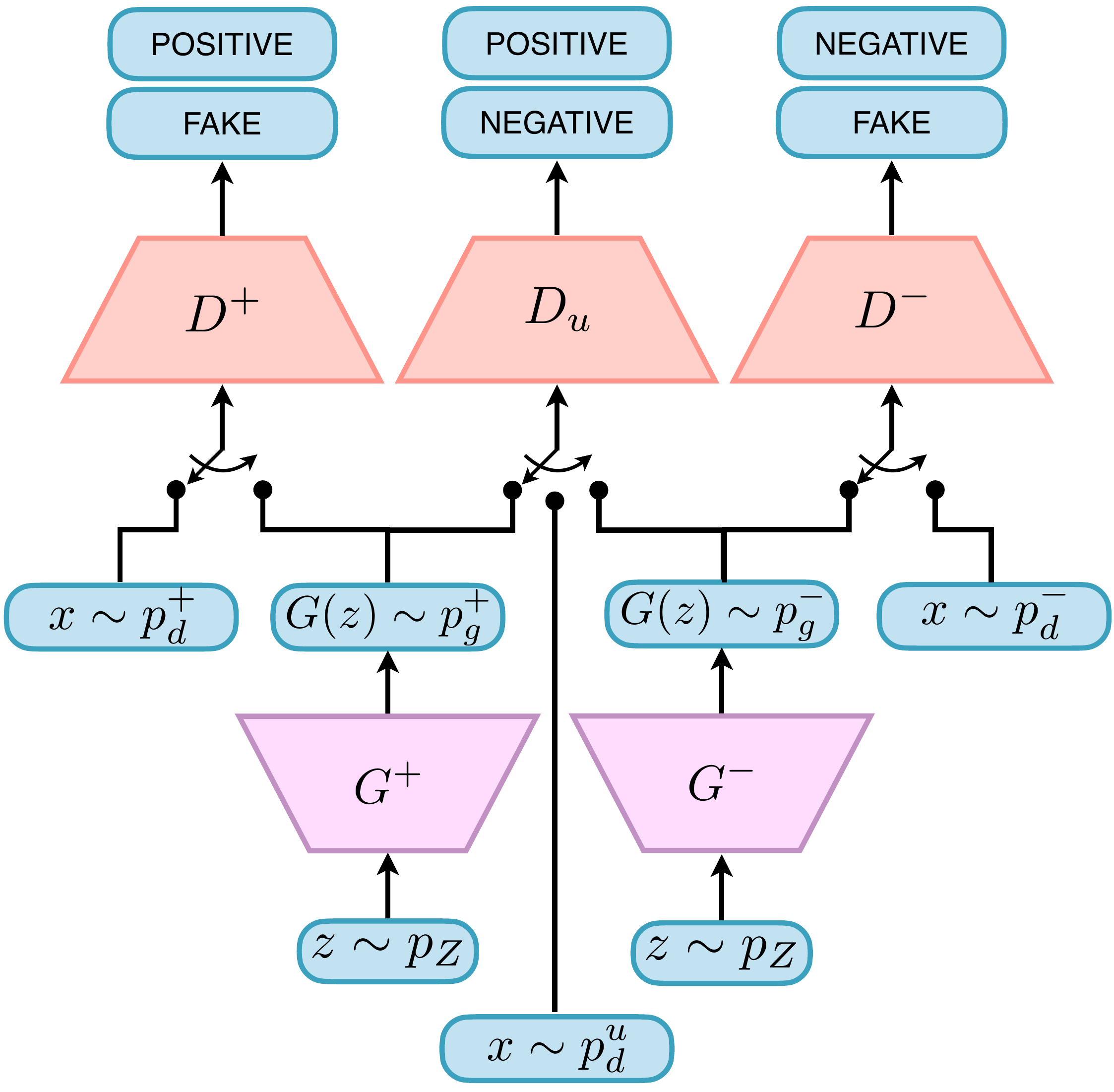} & \includegraphics[height=.4\linewidth]{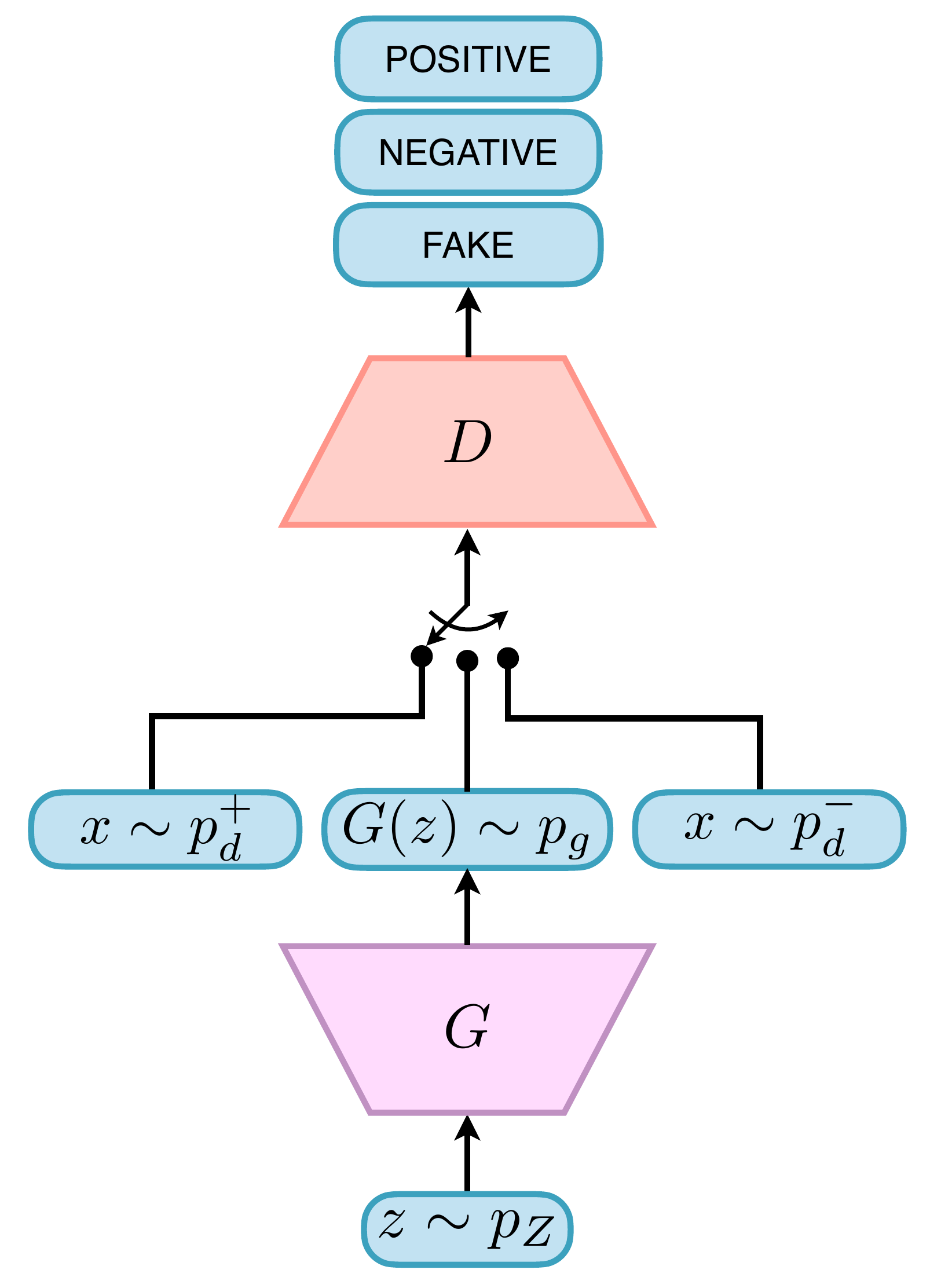} \\
    (a) ACGAN & (b) GenPU & (c) \textbf{Rumi-GAN}
  \end{tabular} 
  \caption[Comparison ACGAN, GenPU, and our proposed Rumi-GAN models]{ (\includegraphics[height=0.014\textheight]{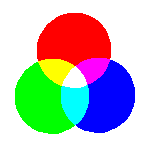} Color Online) Comparison of the pipelines of ACGAN and GenPU vis-\`a-vis the proposed Rumi-GAN.} 
  \label{Fig:RumiGAN_Block}
  \end{center}
\end{figure*}

The approach that we advocate in this paper is motivated by the famous Persian poet Rumi's philosophy, \textit{``The art of knowing is knowing what to ignore.''} In the context of machine learning, we interpret it as empowering models to \textit{learn by ignoring}. Formally, this represents learning both from examples as well as counterexamples. We refer to our formulation as \textit{The Rumi Formulation}. We take the ``middle-ground'' between the conditional data generation capability of ACGANs, and the positive, unlabelled, and negative data classification feature of models such as GenPU. Figure~\ref{Fig:RumiGAN_Block} brings out the differences between these approaches and the proposed one. \par
 
In Rumi-GAN, the discriminator learns to bin the samples it receives into one of three classes: (1) Positives, representing samples from the target distribution; (2) Negatives, representing samples from the same underlying background distribution as that of the target data, but from a subset that must be avoided; and (3) Fakes, which are the samples drawn from the generator. The generator, on the other hand, is tasked with learning the distribution of only the positive ones by simultaneously learning to avoid the negative ones. Effectively, the Rumi-GAN philosophy is a unique case of \textit{complementary learning}, where the negative class of exemplars serve the purpose of boosting the positive-class learning capability of the generator. \par 
This paper is structured as follows. In Section~\ref{Sec:Rumi_formulation}, we first reformulate the standard GAN (SGAN) and LSGAN within the Rumi framework, giving rise to Rumi-SGAN and Rumi-LSGAN, respectively, and derive the optimal distribution learnt in each case. Any other flavor of GAN could also be accommodated within the Rumi framework.  In Section~\ref{Sec:Experimental_val}, we demonstrate the learning capabilities of Rumi-LSGAN on MNIST and Fashion-MNIST datasets and show improvements over the baseline LSGAN and ACGAN approaches. In Section~\ref{Sec:Unbalanced_data}, considering the particular case of training on unbalanced datasets, we simulate minority classes in MNIST, CelebA and CIFAR-10 datasets, and show that Rumi-GAN learns a better representation of the target distribution and has a lower generator network complexity when compared with the state-of-the-art ACGAN and CGAN architectures. Although we demonstrate the applicability of Rumi-GANs on unbalanced datasets, the formulation is extendable to any complementary learning task.

\section{The Rumi Formulation of GANs} \label{Sec:Rumi_formulation}

Here, we consider a scenario where the real data distribution consists of either overlapping or non-overlapping subsets, one labeled positive and the other negative. In the Rumi formulation of SGAN and LSGAN, the discriminator is converted into a three-class classifier giving scalar outputs. The data distribution \(\pd\) is split into two: (i) the target distribution that the GAN is required to learn (\(\pdp\)); and (ii) the distribution of samples that it must avoid (\(\pdn\)). Unlike ACGANs, in the Rumi formulation, we weigh the generator samples from the desired and undesired classes differently. In principle, one could develop the Rumi counterpart of any known \(f\)-divergence~\cite{fGAN16} or integral probability metric based GAN~\cite{WGAN17,MMDGAN17}. The details are deferred to the supplementary (cf. Appendices~\ref{AppSec: Rumi-LSGAN} -~\ref{AppSec: Rumi-WGAN}). In the following, we present the optimal discriminator and generator functions obtained for Rumi-SGAN and Rumi-LSGAN. 

\subsection{The Rumi-SGAN} \label{Rumi-GAN}
The Rumi-SGAN discriminator loss comprises three terms: the expected cross-entropy between (a)  \( [1, 0]^\mathrm{T} \) and \( [D(\x), 1 - D(\x)]^\mathrm{T} \) for the positive data; (b)  \( [0, 1]^\mathrm{T} \) and \( [D(\x), 1 - D(\x)]^\mathrm{T}\) for the generator samples; and (c) \( [0, 1]^\mathrm{T} \) and \( [D(\x), 1 - D(\x)]^\mathrm{T}\) for samples drawn from the negative class, and is given as:
\begin{align}
\loss^{S}_D &= - \left( \alphap\E_{\x \sim \pdp} \left[ \log D(\x)\right] + \E_{\x \sim p_g} \left[ \log(1 - D(\x))\right] + \alphan\E_{\x \sim \pdn} \left[ \log(1 - D(\x))\right] \right),
\label{S Rumi-GAN discriminator}
\end{align}
where \(\alphap\) and \(\alphan\) are the weights attached to the losses associated with the positive and negative subsets, respectively. The min-max generator considers \( \loss_G = - \loss_D \), subject to the integral constraint \(\ds \Omega_{\pg}: \int_{\mathcal{X}\subseteq\mathbb{R}^n} \pg(\x)~ \mathrm{d}\x = 1 \), and the non-negativity constraint \( \ds \Phi_{\pg} : \pg(\x) \geq 0, ~\forall~ \x\). Incorporating the constraints, we consider the Lagrangian
\begin{align}
&\loss^S_G(D^*(\x)) = - \loss^S_D(D^*(\x)) + \lambda_p \left( \int_\mathcal{X} \pg(\x)~\mathrm{d}\x - 1 \right) + \int_{\mcalX} \mu_p(\x) \pg(\x)\,\mathrm{d}\x,
\label{S Rumi-GAN generator}
\end{align}
where \( \lambda_p\) and \(\mu_p(\x) \) are the Karush-Kuhn-Tucker (KKT) multipliers with \(\mu_p(\x) \leq 0,~\forall~ \x \in \mcalX\), and \(\mu_p(\x)p_g^*(\x) = 0,~\forall~\x \in \mcalX\). The optimal KKT multipliers will have to be determined.
\begin{lemma}{\textbf{The optimal Rumi-SGAN}: \label{Optimal Rumi-GAN}} Consider the GAN optimization defined through Equations \eqref{S Rumi-GAN discriminator} and \eqref{S Rumi-GAN generator}. The optimal discriminator \(D^*(\x) \) and the optimal generator density \(\pg^*(\x)\) are given as
\begin{align*}
D^*(\x) = \frac{\alphap\pdp(\x)}{\alphap\pdp(\x) + \pg(\x) + \alphan\pdn(\x)} \quad \text{and} \quad \pg^*(\x) = (1 + \alphan)\pdp(\x) - \alphan\pdn(\x),
\end{align*} 
respectively, where \( \alphan \geq \alphap - 1,\) and \( \alphap \in [0, 1]\). The optimal KKT multipliers are \( \mu_p^*(\x) := 0,~\forall~\x \in \mcalX\), and \( \lambda^* = \log(\frac{1 + \alphap + \alphan}{1 + \alphan}) \).
\end{lemma}
\textbf{Proof}: The cost functions in Equations \eqref{S Rumi-GAN discriminator} and \eqref{S Rumi-GAN generator} could be minimized point-wise. The solution to \eqref{S Rumi-GAN discriminator} is relatively straightforward since there are no constraints. Minimization of \eqref{S Rumi-GAN generator} subject to \( \Omega_{\pg} \) and \(\Phi_{\pg}\) gives \( \pg^* =  \left(\frac{\kappa(\x)}{1 - \kappa(\x)}\right) \alphap\pdp(\x) - \alphan \pdn(\x) \), where \( \kappa(\x) = e^{-\lambda_p - \mu_p(\x)}\). Enforcing the integral and non-negativity constraints, \(\Omega_{\pg}\) and \(\Phi_{\pg}\), respectively, yields the optimal solution. \qedwhite

\subsection{The Rumi-LSGAN} \label{Rumi-LSGAN}
The Rumi-LSGAN loss minimizes the least-squares distance between the discriminator output \(D(\x)\) and: (a) the class label \(\bp\) for positive samples; (b) the class label \(\bn\) for negative samples; and (c) the class label \(a\) for samples coming from the generator. Simultaneously, the generator minimizes the least-squares distance between \(D(\x)\) and a class label \(c\) for all the samples it generates. The weighted loss functions become
\begin{align}
\loss^{LS}_{D} &= \betap \E_{\x \sim \pdp} \left[  \left( D(\x) - \bp \right)^2 \right] + \betan \E_{\x \sim \pdn} \left[  \left( D(\x) - \bn \right)^2 \right] + \E_{\x \sim \pg} \left[ \left( D(\x) - a\right)^2 \right]\label{LS Rumi-GAN discriminator},\\
\loss^{LS}_{G} &= \betap \E_{\x \sim \pdp} \left[  \left( D^*(\x) - c \right)^2 \right] + \betan \E_{\x \sim \pdn} \left[  \left( D^*(\x) - c \right)^2 \right] + \E_{\x \sim \pg} \left[ \left( D^*(\x) - c\right)^2 \right], \label{LS Rumi-GAN generator}
\end{align}
where \(\loss_G^{LS}\) is subjected to the integral and non-negativity constraints \( \Omega_{\pg} \) and \( \Phi_{\pg}\), respectively. 
\begin{lemma}{\textbf{The optimal Rumi-LSGAN}: \label{Optimal Rumi-LSGAN}} Consider the LSGAN optimization problem defined through Equations \eqref{LS Rumi-GAN discriminator} and \eqref{LS Rumi-GAN generator}. Assume that the labels satisfy \( a \leq \frac{\bp+\bn}{2}\) with \( \bp > \bn \). The optimal discriminator and generator are given as
\begin{align}
D^*(\x) &= \frac{\bp \betap \pdp + \bn\betan\pdn + a\pg }{\betap\pdp + \betan\pdn + \pg} \quad \text{and} \quad & \pg^*(\x) &=  \betap \eta^+ \pdp(\x) + \betan \eta^- \pdn(\x),
\label{pg LS Rumi-GAN}
\end{align} 
respectively, where \( \eta^+ = \left( \frac{(1+\betan)(a - \bp) - \betan(a-\bn)}{\betap(a - \bp) + \betan(a-\bn)}\right) \), and  \( \eta^- = \left( \frac{(1+\betap)(a-\bn) - \betap(a - \bp)}{\betap(a - \bp) + \betan(a-\bn)}\right)\). \end{lemma}
\textbf{Proof}:  As in the case of SGANs, the optimization of the costs in Equations~\eqref{LS Rumi-GAN discriminator} and~\eqref{LS Rumi-GAN generator}, with \(\loss_G^{LS}\) subjected to the constraints \(\Omega_{\pg}\) and \(\Phi_{\pg}\), is carried out point-wise. The detailed proof is presented in Appendix~\ref{AppSec: Rumi-LSGAN} \qedwhite

\subsection{The Optimal Generator} \label{OptimalGen}
Let us now analyze the optimal generator distributions obtained in Lemmas \ref{Optimal Rumi-GAN} and \ref{Optimal Rumi-LSGAN}. The Rumi-SGAN generator \(\pg^* = (1 + \alphan)\pdp - \alphan\pdn\) subject to \( \alphan \geq \alphap - 1,\) and \(\alphap \in [0,1]\) has a family of solutions based on the choice of the weights \( \alphan\) and \(\alphap \). The generator always learns a mixture of \(\pdp\) and \(\pdn\), and it latches on to \( \pdp\) when \( \alphan = 0 \), or \(\pdn\) when \( \alphap = 0\). These corner cases represent the scenarios where the corresponding loss-terms are ignored and the Rumi-SGAN coincides with the standard GAN formulation.\par
The optimal Rumi-LSGAN generator in Equation~\eqref{pg LS Rumi-GAN} shows that the weights and the class labels can be leveraged to control the learnt mixture density. In particular, when \(\betap = \betan\), the separation between \( a, \bp,\) and \(\bn\) decides the mixture weights: when the negative class label \( \bn\) is closer to \(a\), the generator gives prominence to \(\pdp\) and vice versa. On the other hand, when \(a = \frac{\bp + \bn}{2}\), \(\betap\) and \(\betan\) control the mixing. As a special case, when \(a < \frac{\bp + \bn}{2}\), and \(\betap =\frac{a - \bn}{\bn - \bp}\), we have the optimal generator \(\pg^* = \pdp\). This is important as we can now train a discriminator using samples taken from both the positive and the negative classes while still learning the distribution of the positive class only.


\section{Experimental Validation} \label{Sec:Experimental_val}
We conduct experiments on MNIST \cite{MNIST}, Fashion-MNIST \cite{FMNIST}, CelebA \cite{CelebA} and CIFAR-10 \cite{CIFAR10} datasets. The GAN models are coded in TensorFlow 2.0 \cite{TF}. The generator and discriminator architectures are based on deep convolutional GAN \cite{DCGAN}. In all the cases, latent noise is drawn from a 100-dimensional standard Gaussian \(\mathcal{N}(\bm{0}_{100},\mathbb{I}_{100})\). The ADAM optimizer \cite{Adam} with learning rate \( \eta = 10^{-4} \) and exponential decay parameters for the first and second moments \( \beta_1 = 0.50 \text{ and } \beta_2 = 0.999 \) is used for training both the generator and the discriminator. A batch size of 100 is used for all the experiments and all models were trained for 100 epochs, unless stated otherwise. For every step of the Rumi-LSGAN generator, the discriminator performs two updates, one with the positive data and another one with the negative data. In order to make a fair comparison, for every update of the discriminator and generator of Rumi-LSGAN, we perform two updates of the discriminator and one update of the generator on the baseline approaches. This approach of updating the discriminator multiple times per update of the generator is actually in favor of the baseline GANs as shown in \cite{WGAN17,Unrolled17}. As argued in Section~\ref{OptimalGen}, while the Rumi-SGAN learns a mixture of the positive and negative data distributions, the Rumi-LSGAN is capable of latching on to the positive data distribution only. In view of this property, we report comparisons between Rumi-LSGAN and the baseline LSGAN \cite{LSGAN} and ACGAN \cite{ACGAN17}. We set \((\bn,a,\bp,c) = (-1,0,2,1.5)\) for Rumi-LSGAN, and use  \((a,b,c) = (0,1,1)\) for the baseline LSGAN as proposed by Mao \textit{et al.} \cite{LSGAN}. Given the batches of positive, negative, and generated samples: \(\mathcal{D}^{+} = \{\x_i;~\x_i\sim\pdp,~i= 1,2,\ldots,N\}\), \(\mathcal{D}^{-} = \{\x_j;~\x_j\sim\pdn,~j= 1,2,\ldots,N\}\), and \(\mathcal{D}^{g} = \{\x_k;~\x_k\sim\pg,~k= 1,2,\ldots,N\} \), respectively, we train the Rumi-LSGAN models by replacing the expectations in the loss functions with their sample estimates as follows:
\begin{align*}
\loss^{LS}_{D} &= \frac{\betap}{N} \sum_{\x_i \in \mathcal{D}^{+}}  \left( D(\x_i) - \bp \right)^2 + \frac{\betan}{N} \sum_{\x_j \in \mathcal{D}^{-}}  \left( D(\x_j) - \bn \right)^2 + \frac{1}{N} \sum_{\x_k \in \mathcal{D}^{g}} \left( D(\x_k) - a\right)^2,~\text{and} \\
\loss^{LS}_{G} &= \frac{\betap}{N} \sum_{\x_i \in \mathcal{D}^{+}}  \left( D(\x_i) - c \right)^2 + \frac{\betan}{N} \sum_{\x_j \in \mathcal{D}^{-}}  \left( D(\x_j) - c \right)^2 + \frac{1}{N} \sum_{\x_k \in \mathcal{D}^{g}} \left( D(\x_k) - c\right)^2.
\end{align*}
We do not enforce the integral or non-negativity constraints explicitly, as it turns out that these are automatically satisfied during the optimization (cf. Appendix~\ref{AppSec: Rumi-LSGAN}). A comparison between the Rumi-GAN variants is also provided in Appendix~\ref{AppSec: MNIST}.


\subsection{Evaluation Metrics}
The Fr{\'e}chet inception distance (FID) \cite{FID} is a useful metric for comparing the quality of the images generated by GANs. In our experiments, we use the InceptionV3 \cite{InceptionV3} model without the topmost layer, loaded with ImageNet pretrained weights to generate the embeddings over which the FID scores are evaluated. InceptionV3 requires a minimum input dimension of \( 76\times 76\times 3\). Hence, color images (CelebA, CIFAR-10) are rescaled to \(80\times 80\times 3\) using bilinear interpolation. Gray-scale images (MNIST and Fashion MNIST) are rescaled to \(80 \times 80\) and then replicated across three channels.\par 
FID scores provide an objective assessment of the image quality, but not the diversity of the learnt distribution. Hence, we also report Precision-Recall (PR) behaviour, following the method of Sajjadi \textit{et al.}~\cite{PR}. The PR curve, in the context of generative models, gives the precision that a model of a given recall is likely to have, and vice versa. A high precision with a low recall implies that the model is able to generate samples of high quality in comparison with those coming from the target distribution, but at the cost of little diversity. On the other hand, a model that generates diverse images of low quality would have a low precision but a high recall. We use both FID scores and PR curves to analyze the performance of Rumi-LSGAN with respect to the baselines.

\begin{figure*}[t!]
\begin{center}
  \begin{tabular}[b]{p{.02\linewidth}||P{.31\linewidth}|P{.31\linewidth}|P{.31\linewidth}}
  &Even MNIST & Overlapping MNIST & Fashion MNIST \\
  \multirow{2}{*}[4.9em]{\rotatebox{90}{LSGAN}} &
    \includegraphics[width=0.95\linewidth]{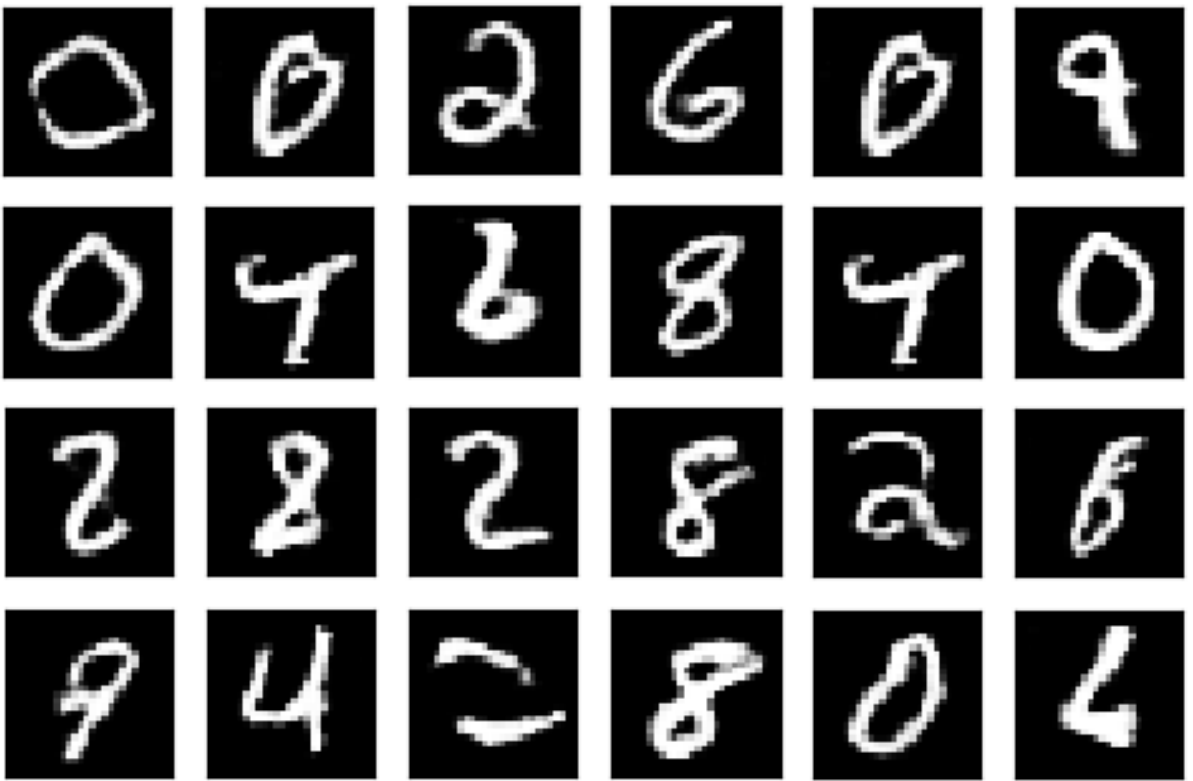} &
     \includegraphics[width=0.92\linewidth]{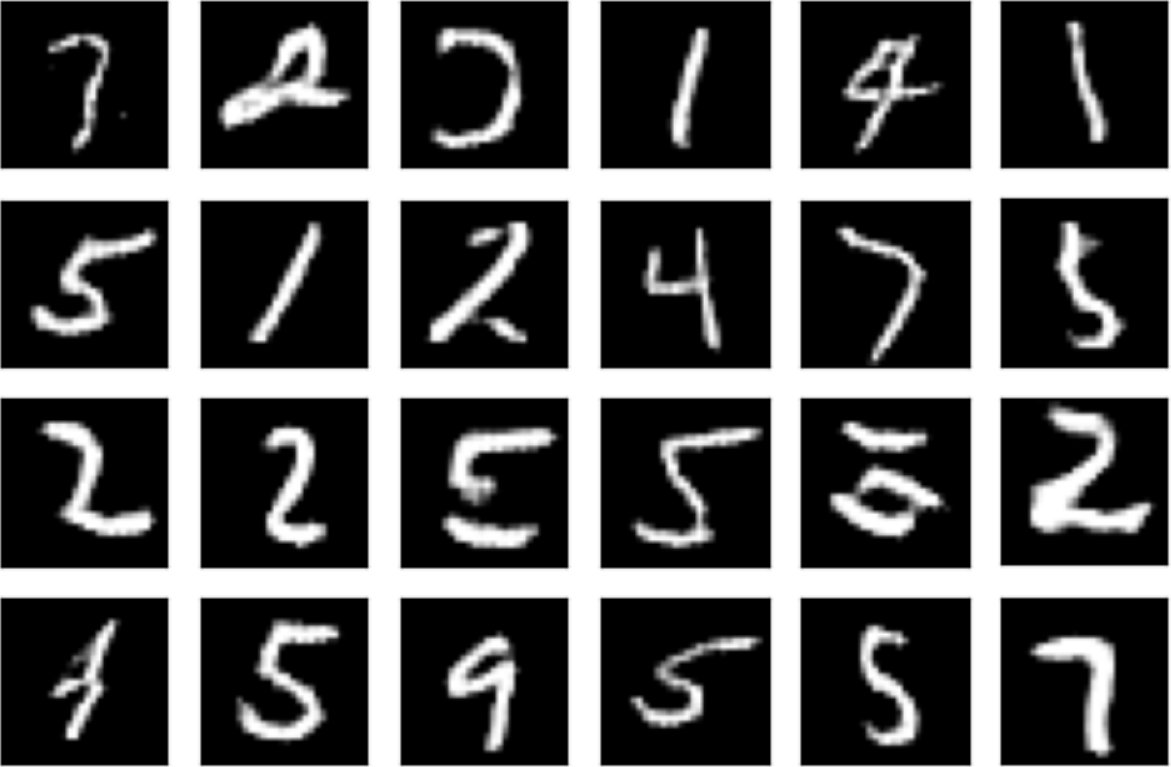} &
      \includegraphics[width=0.95\linewidth]{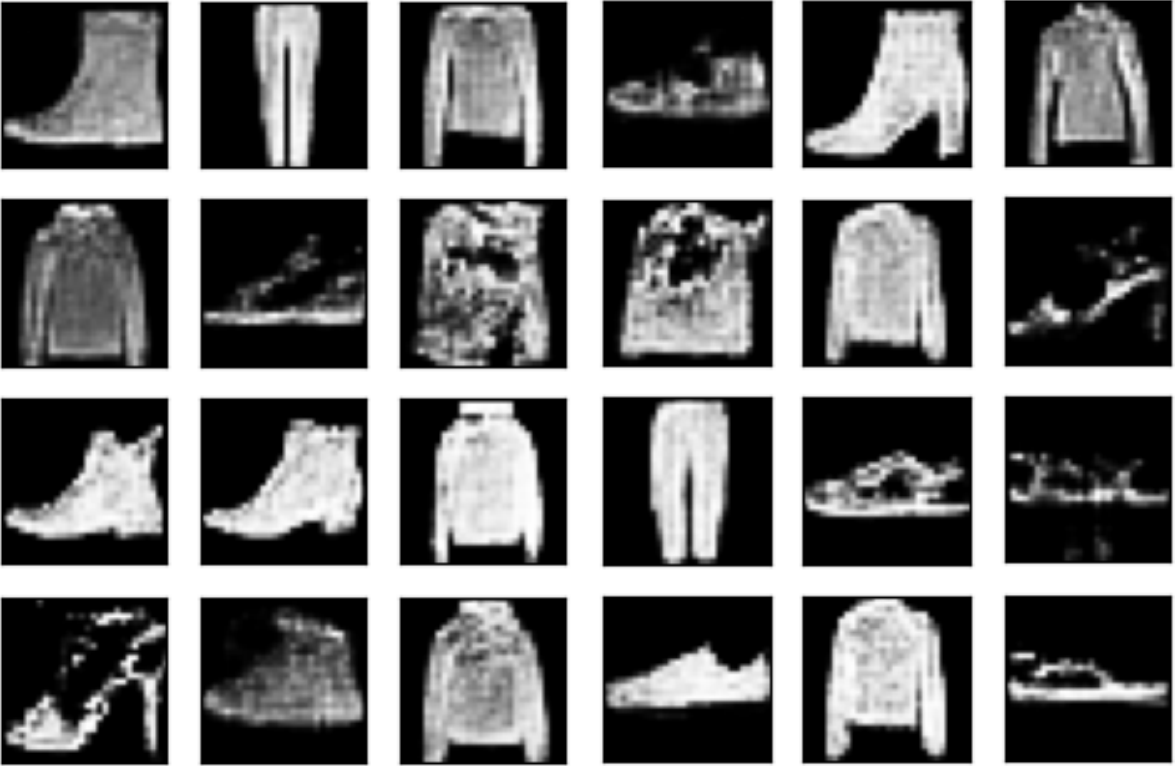} \\[-1pt]
    &(a)  & (d)  & (g) \\[2pt]
    \multirow{2}{*}[4.9em]{\rotatebox{90}{ACGAN}} &
    \includegraphics[width=0.95\linewidth]{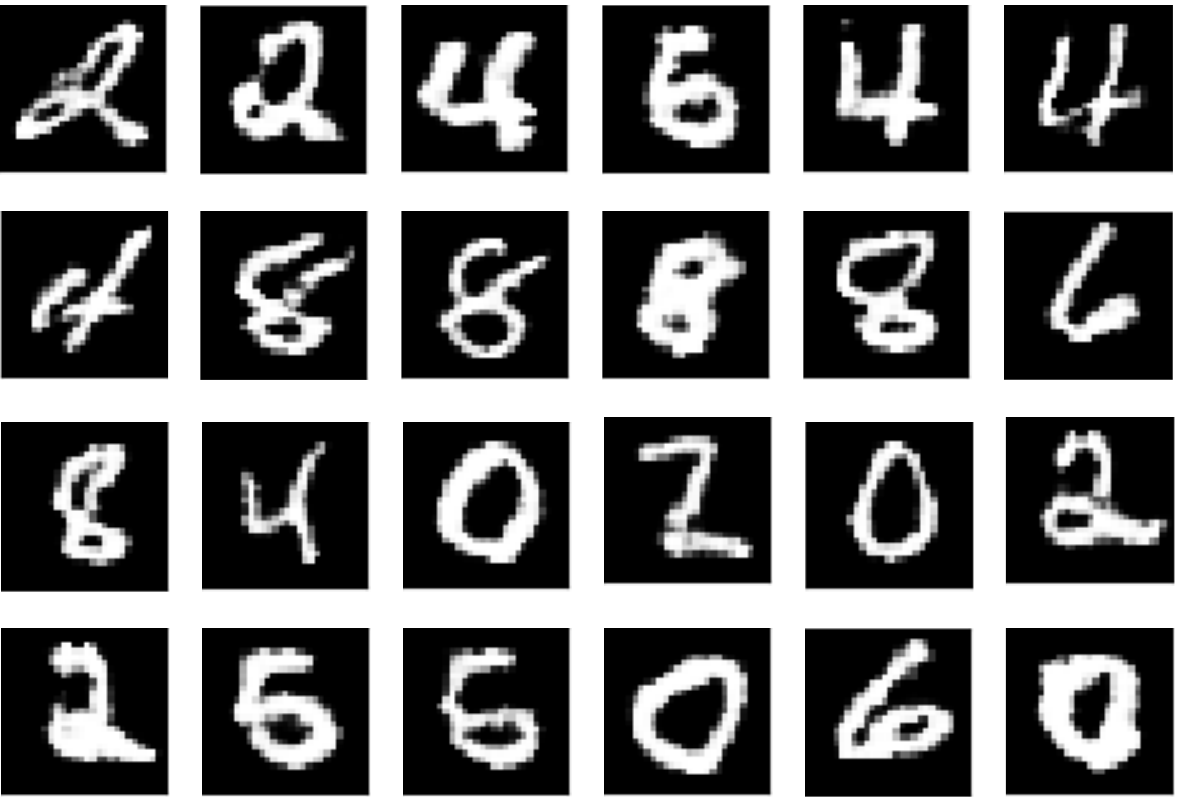} & 
    \includegraphics[width=0.95\linewidth]{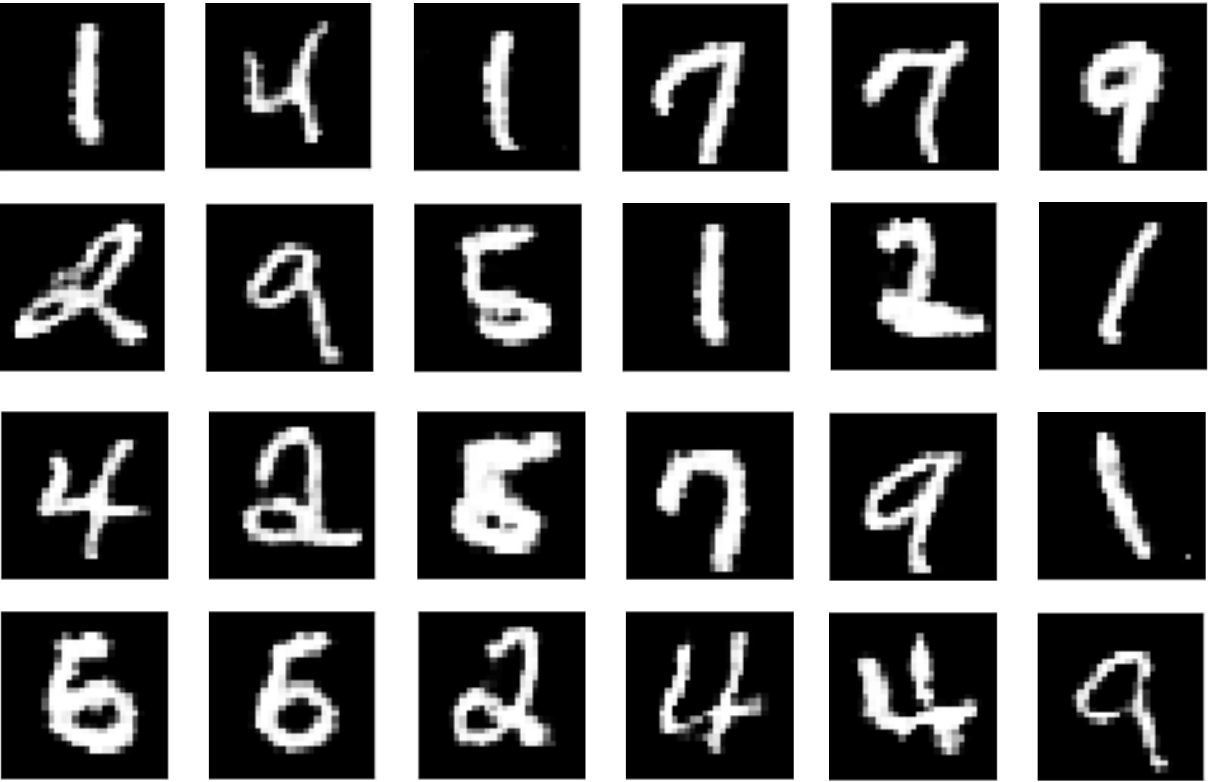} & 
    \includegraphics[width=0.95\linewidth]{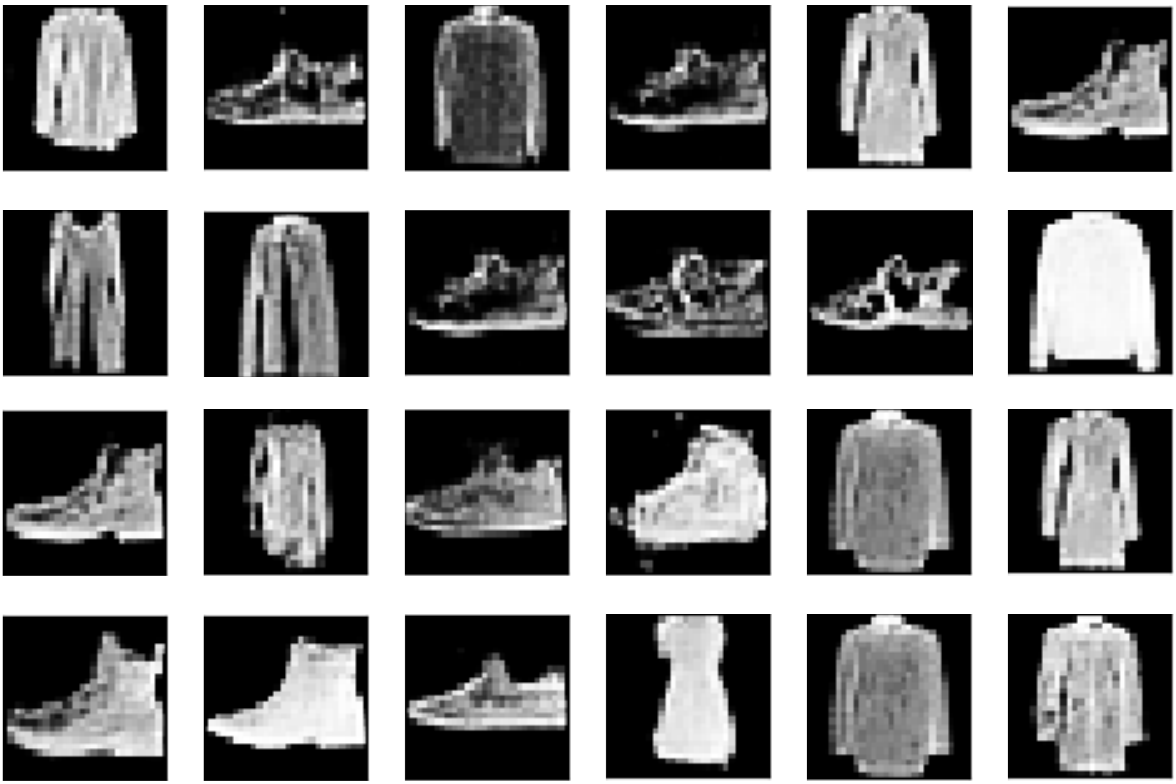} \\[-1pt]
    &(b)  & (e)  & (h) \\[2pt]
    \multirow{2}{*}[6.2em]{\rotatebox{90}{{\bfseries Rumi-LSGAN}}}  &
    \includegraphics[width=0.95\linewidth]{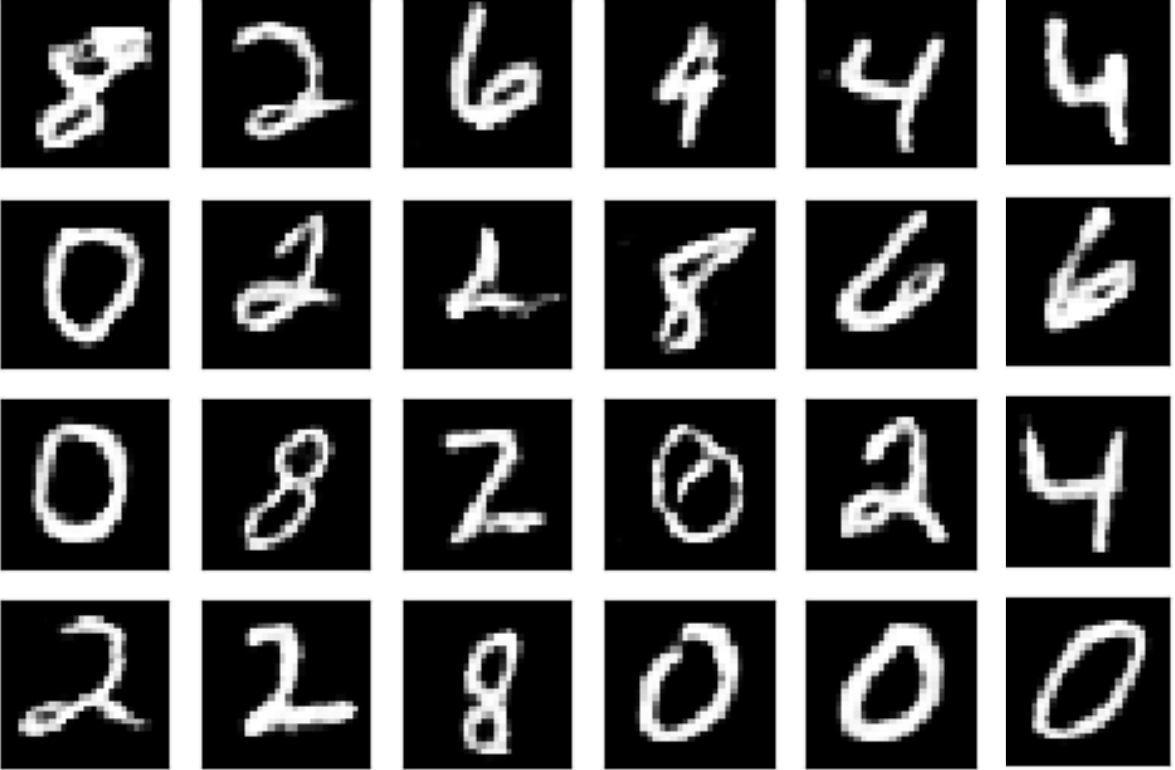} & 
    \includegraphics[width=0.95\linewidth]{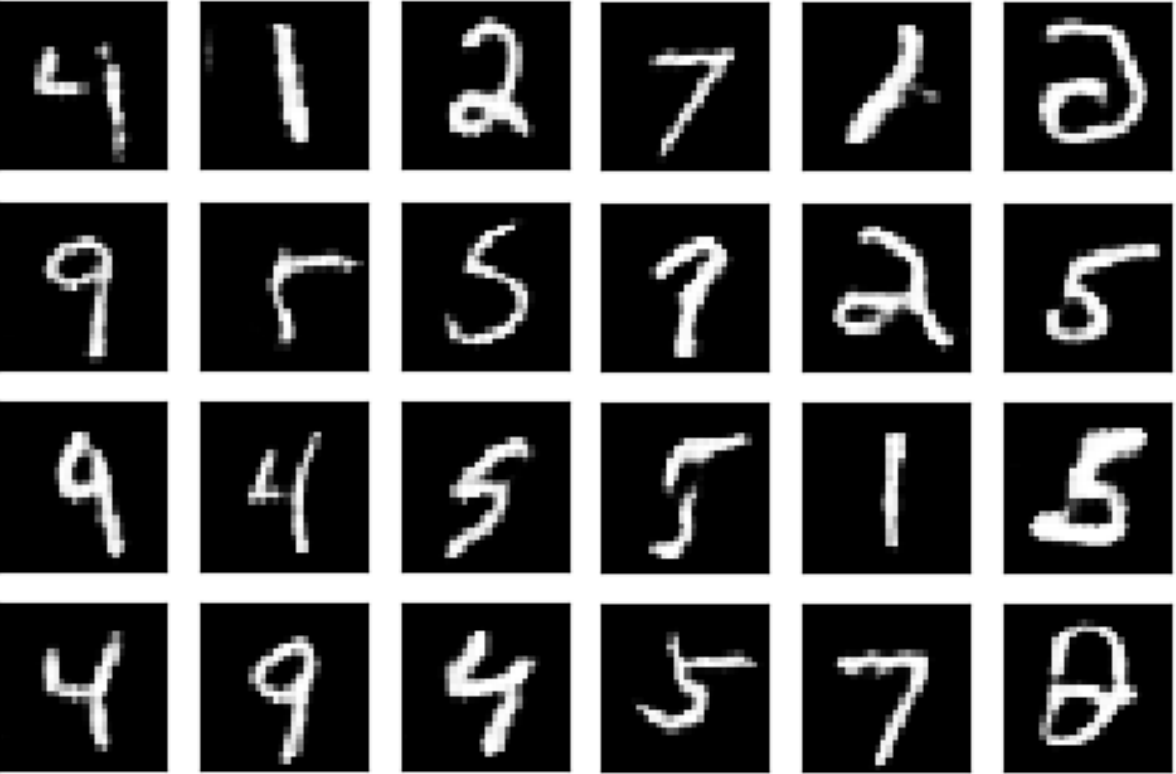} &
     \includegraphics[width=0.95\linewidth]{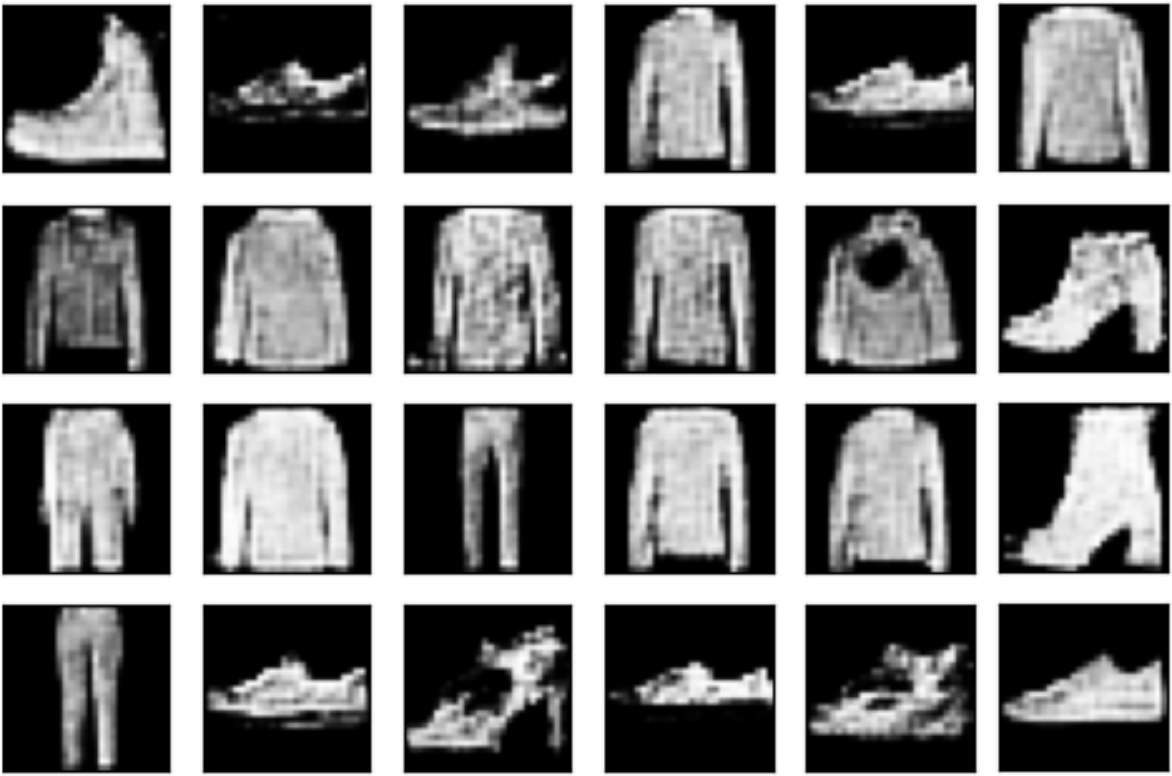} \\[-1pt]
    &(c)  & (f)  & (i)
  \end{tabular} 
  \caption{Comparison of the samples generated by LSGAN, ACGAN, and Rumi-LSGAN on: (a)-(c) MNIST with disjoint positive and negative classes; (d)-(f) MNIST with overlapping positive and negative classes; and (g)-(i) Fashion MNIST with overlapping positive and negative classes.} 
  \label{Rumi-GAN_MNIST_Samples}
  \vspace{-1.5em}
  \end{center}
\end{figure*}

\begin{figure*}[t!]
\begin{center}
  \begin{tabular}[b]{P{.32\linewidth}|P{.32\linewidth}|P{.32\linewidth}}
  Even MNIST & Overlapping MNIST & Fashion MNIST \\
    \includegraphics[width=1\linewidth]{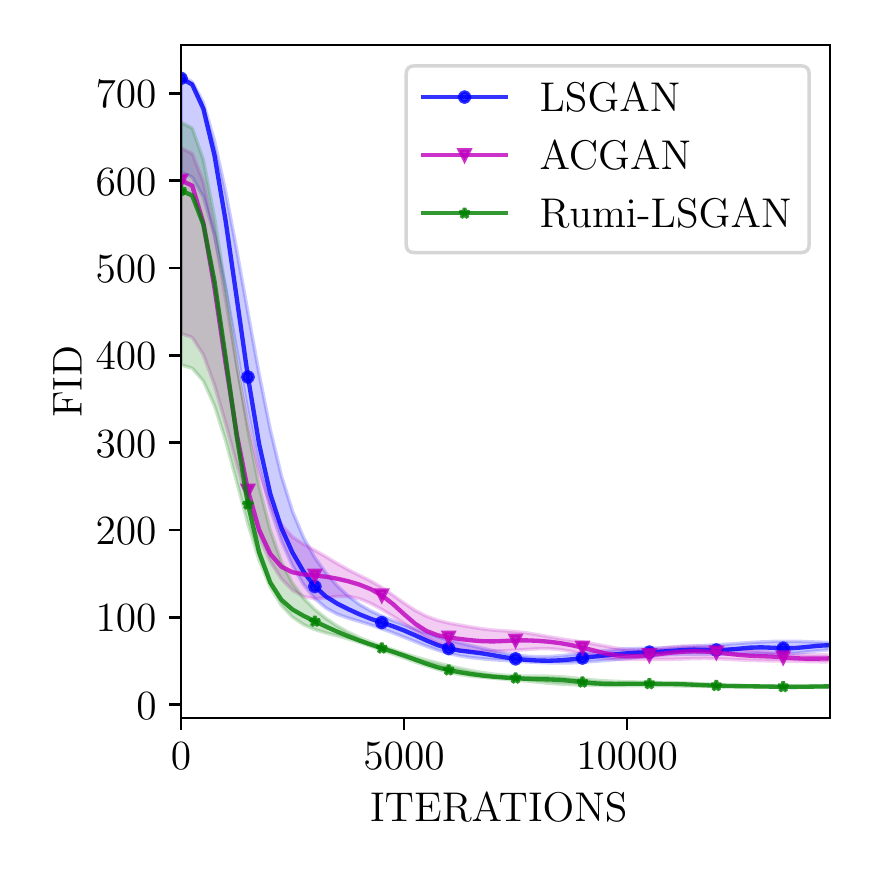} & 
    \includegraphics[width=0.99\linewidth]{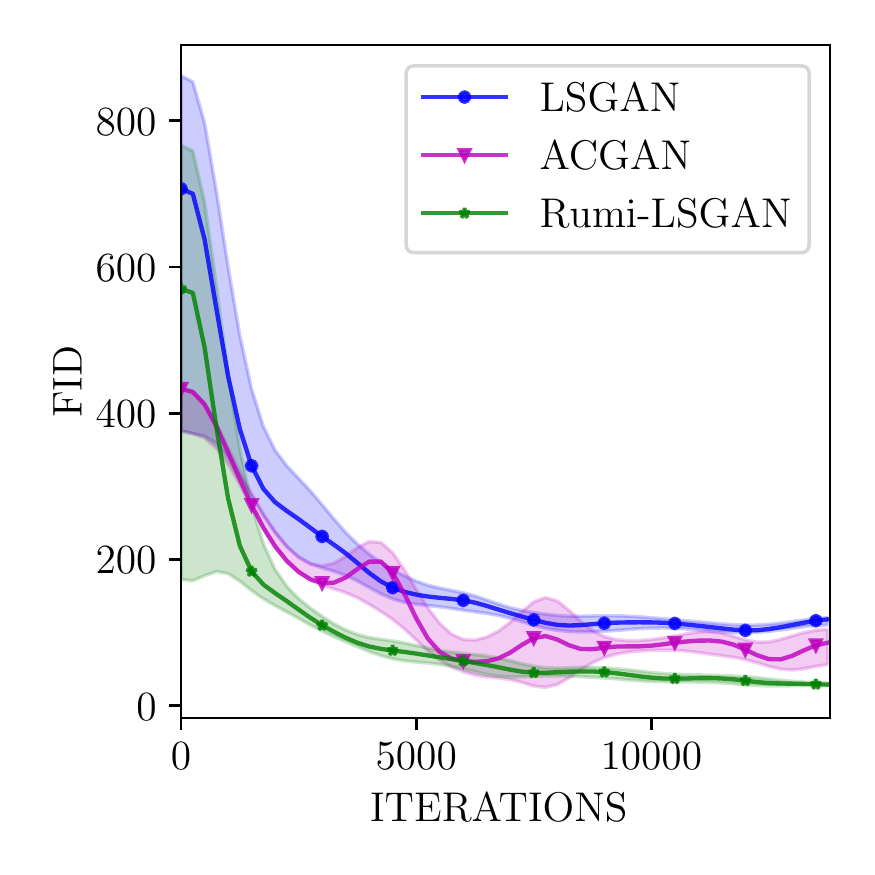} &
    \includegraphics[width=1\linewidth]{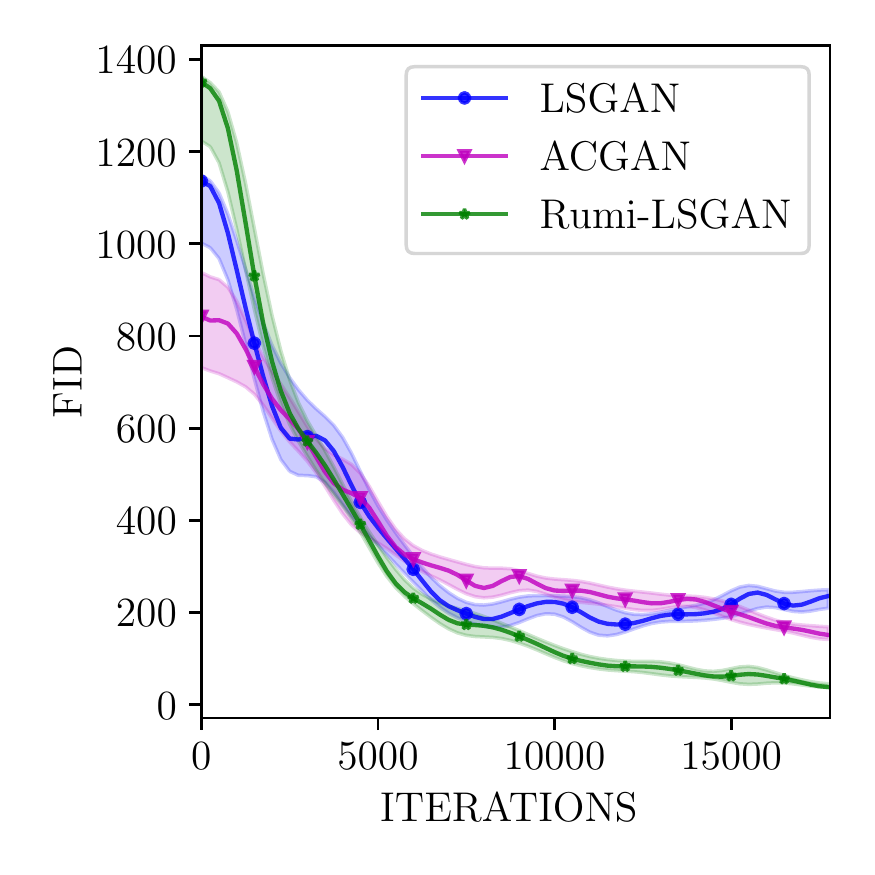} \\[-1pt]
    (a)  & (b) & (c) \\[3pt]
    \includegraphics[width=1\linewidth]{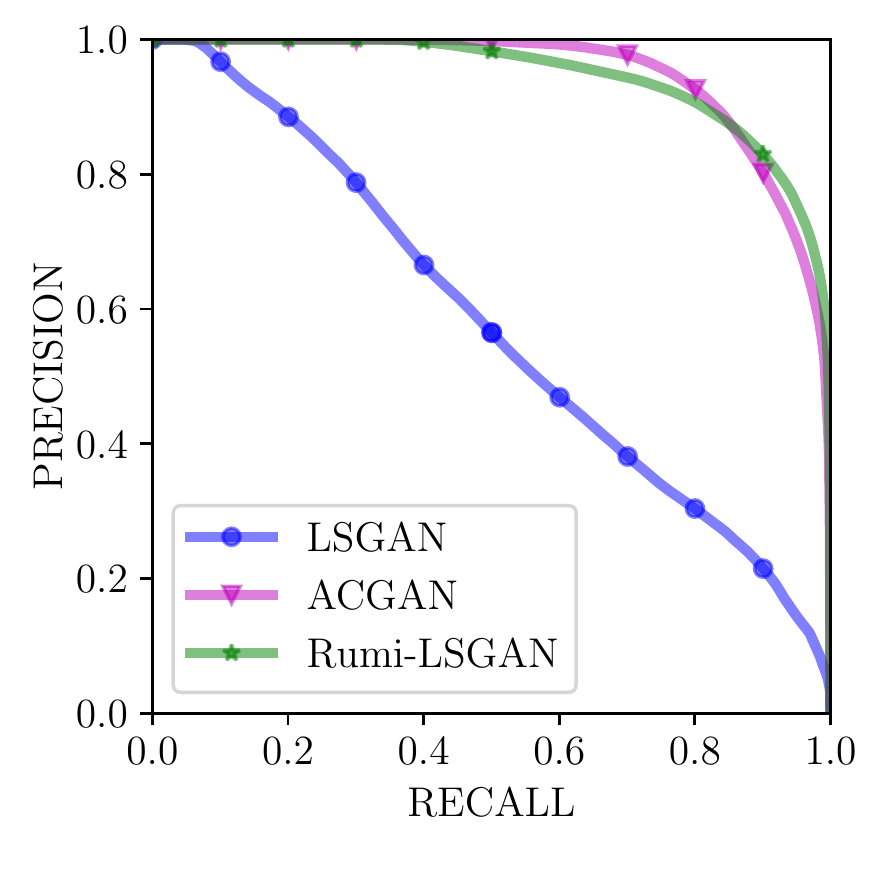} & 
    \includegraphics[width=1\linewidth]{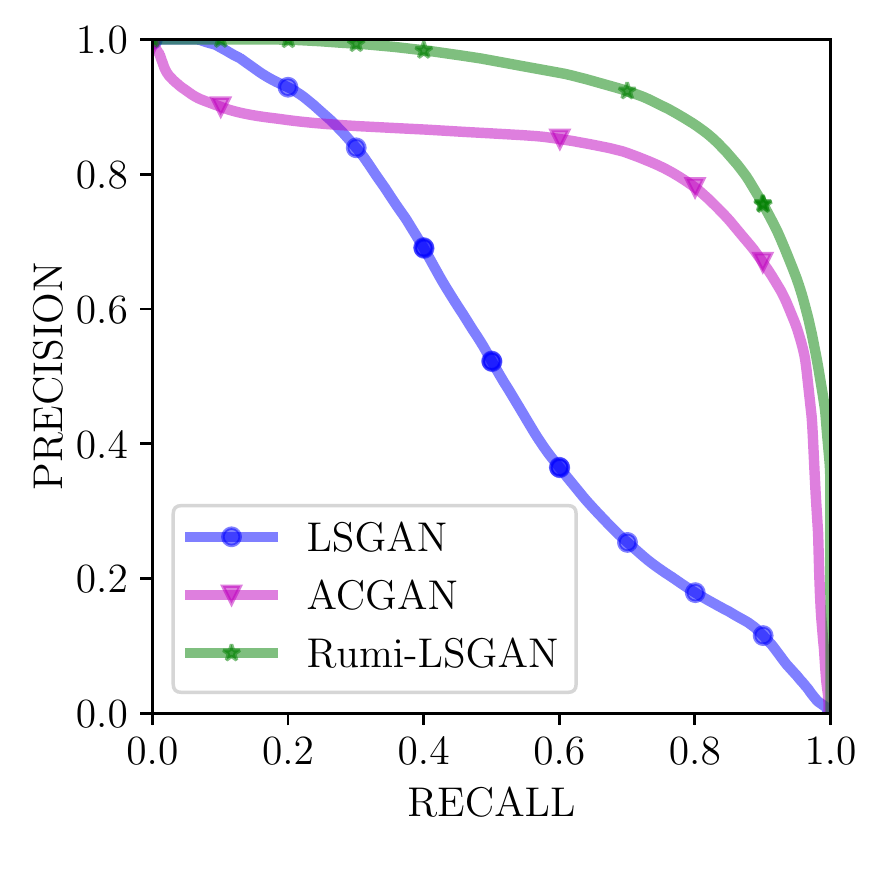} &
    \includegraphics[width=1\linewidth]{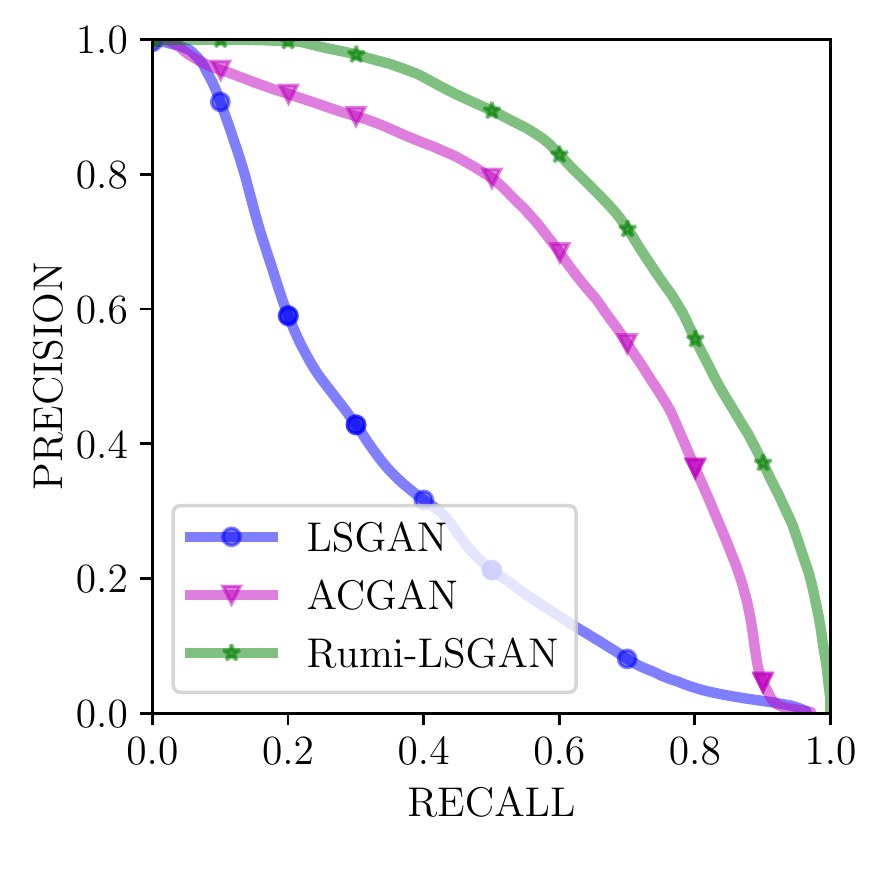} \\[-1pt]
    (d)  & (e) & (f) \\
  \end{tabular} 
  \caption[]{(\includegraphics[height=0.014\textheight]{Rgb.png} Color Online) (a)-(c): A comparison of FID scores as iterations progress on: (a) Even numbers in MNIST; (b) Random subset of MNIST; and (c) Random subset of Fashion-MNIST. (d)-(f): Comparison of precision vs. recall on (d) Even numbers in MNIST; (e) Random subset of MNIST; and (f) Random subset of Fashion MNIST.} 
  \label{Rumi-GAN_MNIST_Compare}
  \vspace{-1.5em}
  \end{center}
\end{figure*}
\vspace{-1em}

\subsection{Experiments on MNIST and Fashion MNIST Datasets} \label{Sec: MNIST_and_FMNIST}
\textit{\textbf{Experimental setup:}} Both baseline LSGAN and the Rumi-LSGAN use identical generator and discriminator networks. For ACGAN, the generator consists of an additional class label embedding on to a \( 7\times7\times1\) layer, which is concatenated with the noise input to the deconvolution layers. For comparison, the baseline model is trained only on the positive subset of data. The FID and PR performance measures are evaluated considering only positive samples as the reference dataset. \par
\textit{\textbf{Results:}} In the first experiment, we pool the five even digit classes of MNIST into the positive class, and the odd ones into the negative class. The LSGAN is trained solely on the positive class data, whereas the ACGAN and Rumi-LSGAN are trained using both positive and negative class data. Figures~\ref{Rumi-GAN_MNIST_Samples}(a)-(c) show the images generated by the three models under comparison. We observe that the Rumi-LSGAN generates sharper images that consistently belong to the positive class unlike the other two models. This is evidence that the Rumi-LSGAN has a superior capability to learn to avoid the negative class. In the next experiment, we consider a configuration of the positive and negative classes with some overlap between them. More specifically, the positive ones are 1, 2, 4, 5, 7, and 9, and the negative ones are 0, 2, 3, 6, 8, and 9. The images generated in this configuration are shown in Figures~\ref{Rumi-GAN_MNIST_Samples}(d)-(f). These images show that, despite the overlap between the positive and negative classes, Rumi-LSGAN consistently generates samples belonging to the positive class only, and of perceptually better quality. This experiment demonstrates that the positive class specification has an over-riding effect in the Rumi formulation, which is a desired feature. \par
Figure~\ref{Rumi-GAN_MNIST_Compare} compares the FID and PR curves of the three models. The shaded tolerance bands indicate the spread in the performance obtained over 10 runs. Rumi-LSGAN was found to settle at a lower FID in both the cases. ACGAN's relatively slower convergence of FID  could be attributed to its more complex generation pipeline. We observe that the unmodified baseline LSGAN, trained only on the desired dataset achieves only average precision and recall values, with a minor improvement in the case of the six random classes, due to the additional samples available for training per epoch. Rumi-LSGAN and ACGAN have comparable PR curves, with the Rumi-LSGAN achieving slightly better precision values due to its ability to strongly latch on to the desired positive class. \par
As an additional experiment, we show results on Fashion MNIST dataset with randomly picked positive and negative classes with overlap. Figures~\ref{Rumi-GAN_MNIST_Samples}(g)-(i) show the images generated by the three GANs. Figures~\ref{Rumi-GAN_MNIST_Compare}(c) and \ref{Rumi-GAN_MNIST_Compare}(f) show that Rumi-GAN converges relatively faster than the baselines with convergence measured in terms of FID and has better PR scores. The ACGAN has a poorer PR performance because it occasionally generates samples belonging to the wrong class. Additional comparisons on CelebA and CIFAR-10 datasets are presented in Appendix~\ref{AppSec: C10CelebA}.

\section{Handling Unbalanced Datasets} \label{Sec:Unbalanced_data}
 We now address a pertinent application of the Rumi-GAN framework --- learning the minority class in unbalanced datasets. We simulate unbalanced data with MNIST, CelebA, and CIFAR-10 datasets by holding out samples from one of the classes and training the models to learn that class. In addition to the baselines considered in Section~\ref{Sec:Experimental_val}, we also compare Rumi-LSGAN with the twin auxiliary classifier GAN (TACGAN)~\cite{TACGAN19} and CGANs with projection discriminator (CGAN-PD)~\cite{CGANPD18}, as they have been shown to generate balanced class distributions. \par 
\textit{\textbf{Experimental Setup:}} As an example, we consider digit 5 from the MNIST dataset to be the held-out class and the rest to be the negative classes. The imbalance is introduced by randomly picking 200 samples out of 6000 (less than \(5\%\)) exemplars of the digit 5 to constitute the minority positive class. The LSGAN is trained on only the 200 positive samples, whereas all other variants are trained on both positive and negative samples.  Each model is trained for \(10^4\) iterations. In the case of CelebA, we present results considering only \(5\%\) of the images in the {\it Female} class to be the minority class, whereas from CIFAR-10, we demonstrate the performance when \(5\%\) of the images in the {\it Horse} class are considered as the minority class. Since CelebA and CIFAR-10 datasets are more sophisticated than MNIST, the models require more iterations to converge. Hence, we trained the models for \(5 \times 10^4\) iterations on these datasets. \par
\textit{\textbf{Evaluation Metrics:}}  We use FID scores and PR to compare the performance of Rumi-LSGAN with the baselines. Only the target class samples are queried from the conditional GAN variants. The FID scores and PR curves are compared with respect to the entire parent class of the held-out samples, which gives us a measure of how well the model has learnt to {\it generalize} and not simply {\it memorize} the training exemplars. We also tabulate the converged FID for all models in Table~\ref{Table:FID_scores}, averaged over multiple test cases. For MNIST and CIFAR-10, we average over results from choosing each class as the minority class. For CelebA, we average over the results obtained by choosing both {\it Males} and {\it Females} as instances of the under-represented class. We evaluate the FID upon convergence by drawing \(10^4\) samples from the target class in the case of CelebA, and the entire target class in the case of MNIST and CIFAR-10 datasets. In all cases, \(10^4\) samples are drawn from the generator. \par

\begin{figure}[H]
\begin{center}
  \begin{tabular}[b]{p{.02\linewidth}||P{.31\linewidth}|P{.31\linewidth}|P{.31\linewidth}}
    &MNIST & CelebA & CIFAR-10 \\
    \multirow{2}{*}[3.6em]{\rotatebox{90}{LSGAN}} &
    \includegraphics[width=1\linewidth]{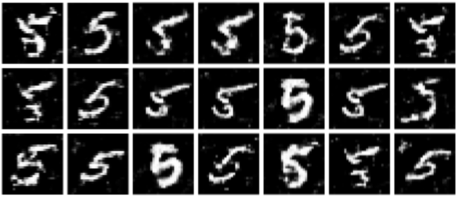} & 
    \includegraphics[width=1\linewidth]{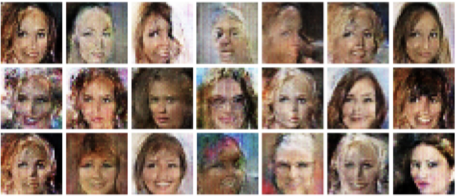} & 
    \includegraphics[width=1\linewidth]{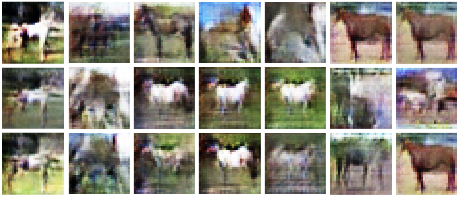} \\[-1pt]
    &(a)  & (h)  & (o) \\[0pt]
     \multirow{2}{*}[3.6em]{\rotatebox{90}{ACGAN}} &
        \includegraphics[width=1\linewidth]{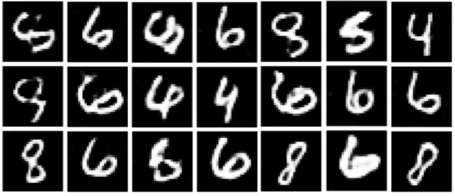} & 
        \includegraphics[width=1\linewidth]{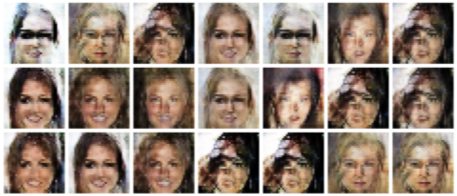} & 
        \includegraphics[width=1\linewidth]{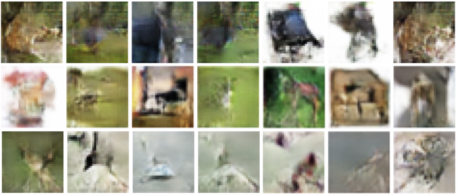} \\[-1pt]
    &(b)  & (i)  & (p) \\[0pt]
     	\multirow{2}{*}[4em]{\rotatebox{90}{ TACGAN }} &
        \includegraphics[width=1\linewidth]{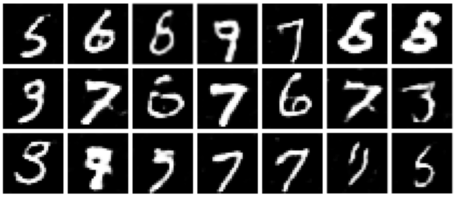} &
        \includegraphics[width=1\linewidth]{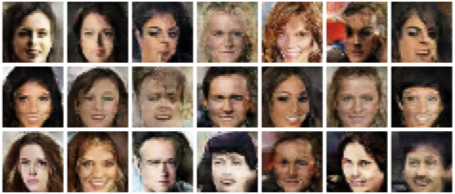} & 
        \includegraphics[width=1\linewidth]{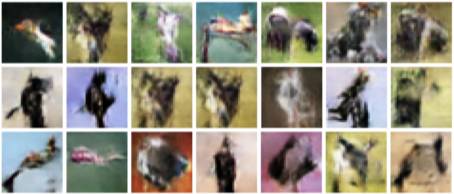} \\[-1pt]
    &(c)  & (j)  & (q) \\[0pt]
        \multirow{2}{*}[4.1em]{\rotatebox{90}{CGAN-PD}} &
        \includegraphics[width=1\linewidth]{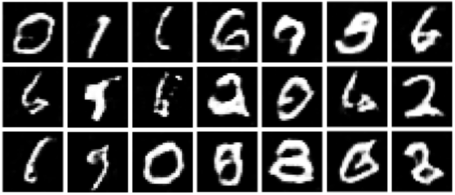} &
        \includegraphics[width=1\linewidth]{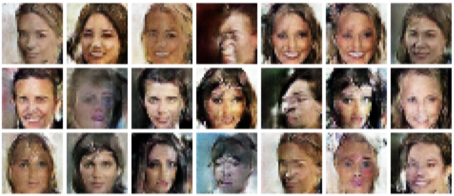} & 
        \includegraphics[width=1\linewidth]{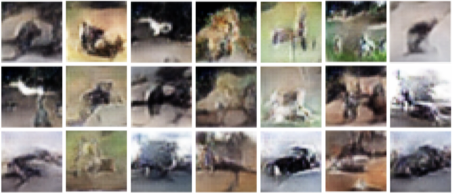} \\[-1pt]
    &(d)  & (k)  & (r) \\[-2pt]
       \multirow{2}{*}[4.3em]{\rotatebox{90}{{\bfseries Rumi-LSGAN}}} &
        \includegraphics[width=1\linewidth]{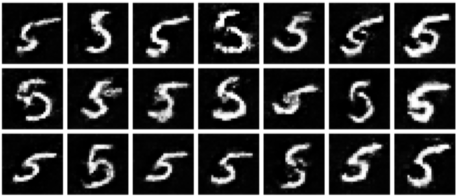} &
        \includegraphics[width=1\linewidth]{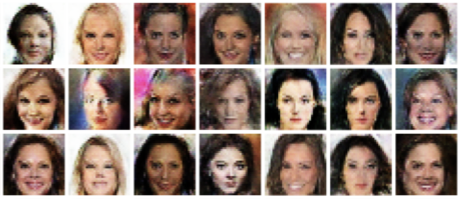} & 
        \includegraphics[width=1\linewidth]{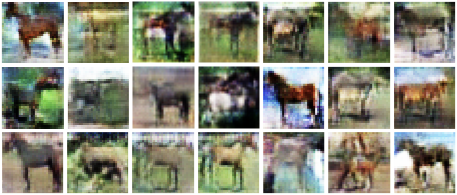} \\[-1pt]
    &(e)  & (l)  & (s) \\[1pt]
     \hline && \\[-10pt]
     \multirow{2}{*}[9em]{\rotatebox{90}{FID vs. Iterations}} &
    \includegraphics[width=0.97\linewidth]{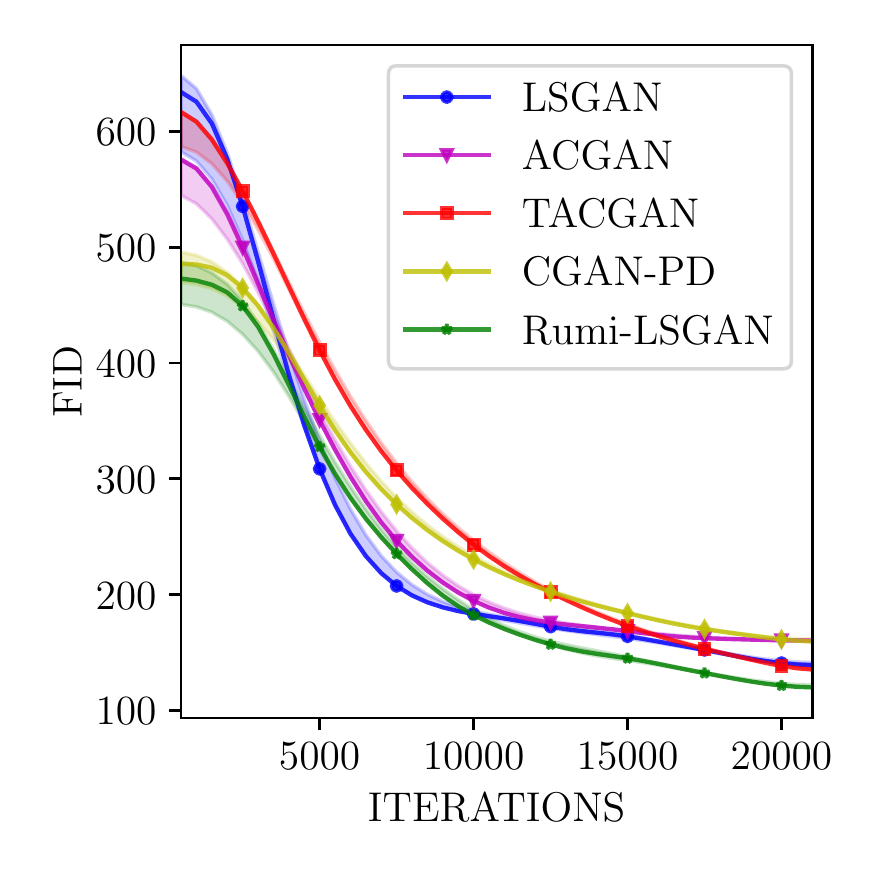} & 
    \includegraphics[width=0.97\linewidth]{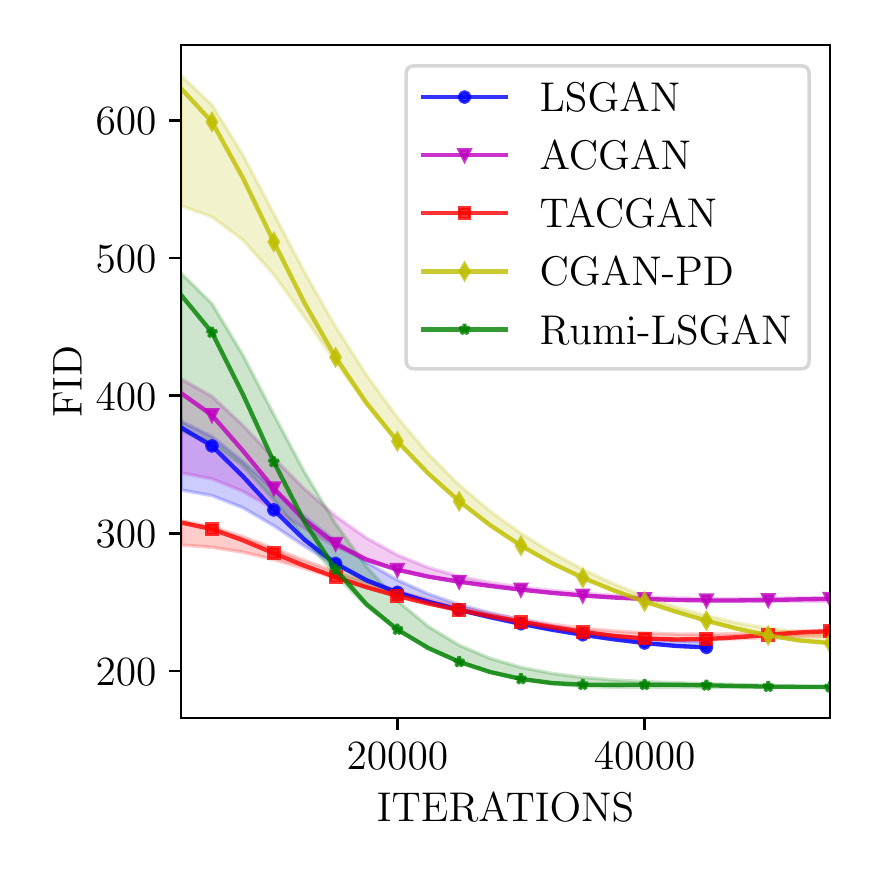} &
    \includegraphics[width=0.96\linewidth]{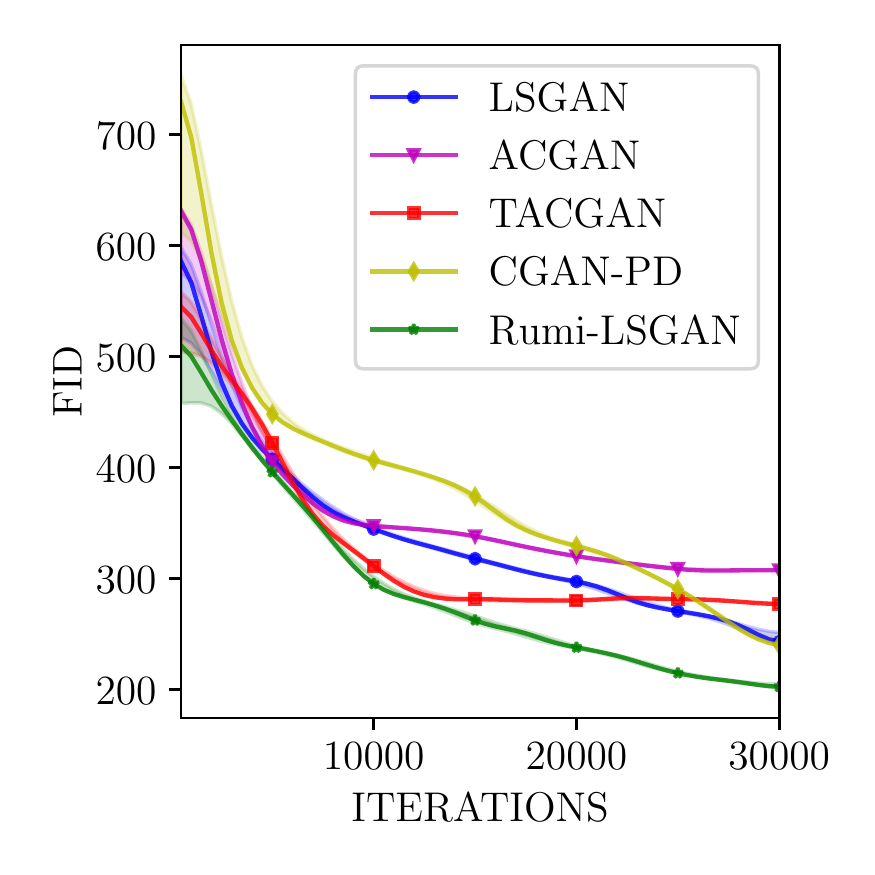} \\[-1pt]
    &(f)  & (m) & (t) \\[0pt]
     \multirow{2}{*}[9.5em]{\rotatebox{90}{ Precision vs. Recall }} &
    \includegraphics[width=0.95\linewidth]{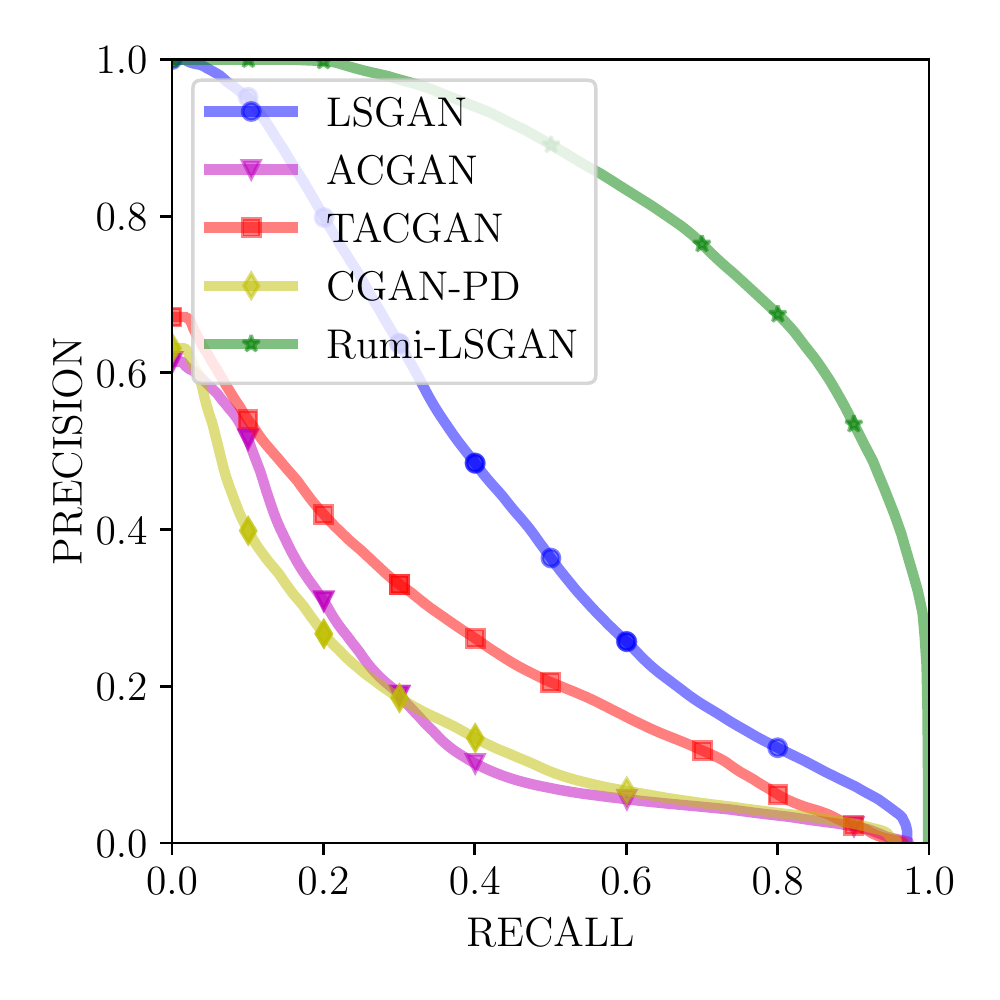} & 
    \includegraphics[width=0.95\linewidth]{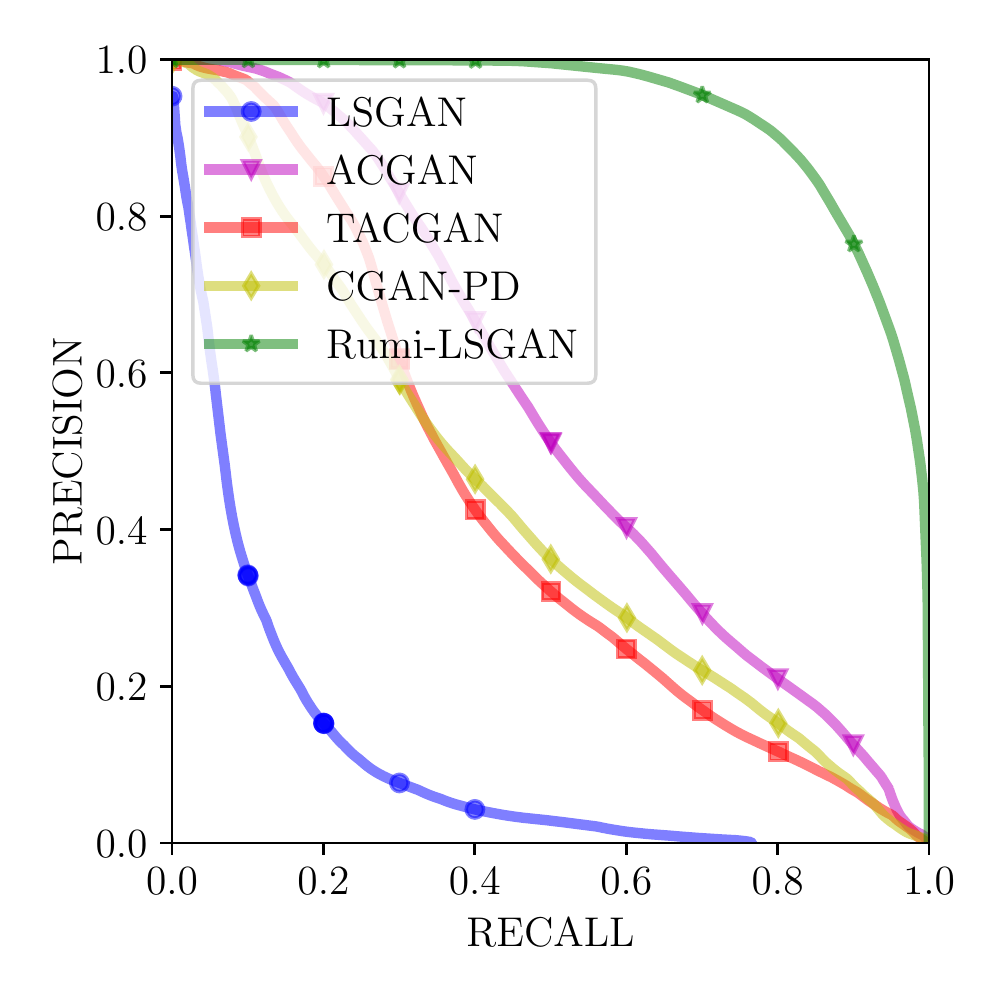} & 
    \includegraphics[width=0.95\linewidth]{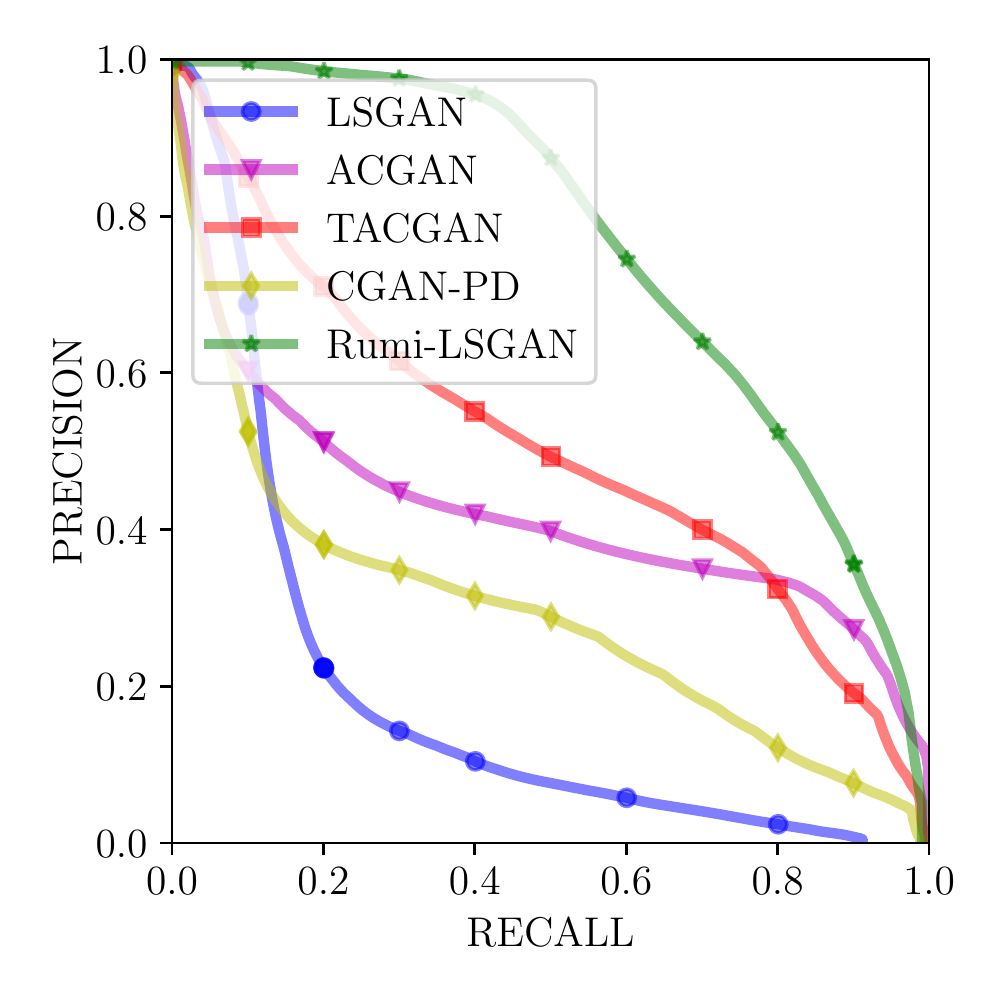}  \\[-1pt]
    &(g)  & (n) & (u) \\[0pt]
  \end{tabular} 
  \caption[]{Results from training various GANs on unbalanced data: A comparison of generated samples on (a)-(e) MNIST with \(5\%\) of digit class \(5\); (h)-(l) CelebA with \(5\%\) of {\it Females} class; (o)-(s)  CIFAR-10 with \(5\%\) of {\it Horses} class. FID and PR curves from training the models on: (f) \& (g) MNIST (digit class 5); (m) \& (n) CelebA (Females); and (t) \& (u) CIFAR-10 ({\it Horse} class). Rumi-LSGAN generates samples of superior quality, while learning only the distribution of the desired target class.} 
   \vspace{-1.2em}
  \label{RumiGAN_Compares}  
  \end{center}
\end{figure}

\textit{\textbf{Results on MNIST Dataset:}}  Figure~\ref{RumiGAN_Compares} presents the samples learnt by the considered GAN variants. From Figures~\ref{RumiGAN_Compares}(a)-(e), we observe that the baseline LSGAN, which received no negative samples, performs better than ACGAN which collapsed on to a few classes (digits 3, 4, 6, and 8). Although TACGAN and CGAN-PD perform better than the baseline ACGAN, they also latch on to similar classes. Rumi-LSGAN produces visually more appealing samples than the baseline LSGAN. This is also reflected in the objective assessment carried out using the FID and PR scores shown in Figures~\ref{RumiGAN_Compares}(f) and~\ref{RumiGAN_Compares}(g). Rumi-LSGAN achieves a lower FID while also exhibiting higher precision and recall than the baseline LSGAN, thereby demonstrating that the model indeed generalized well. Conditional variants, on the other hand, perform poorly as they latch on to the majority classes.  \par
\textit{\textbf{Results on CelebA and CIFAR-10:}} Similar results were obtained in the case of the CelebA and CIFAR-10 datasets. Mariani \textit{et al.} \cite{BAGAN18} showed that the ACGAN suffers from mode collapse when trained on unbalanced data and our experiments also confirmed this behavior. Figures~\ref{RumiGAN_Compares}(h)-(l) and~\ref{RumiGAN_Compares}(o)-(s) show that the Rumi-LSGAN outperforms the baseline conditional models subjectively. In the case of CelebA, ACGAN, TACGAN, and CGAN-PD learn a mixture of both the desired and undesired classes, while on CIFAR-10, these models invariably latched on to the majority class. Table~\ref{Table:FID_scores}, Figures~\ref{RumiGAN_Compares}(m)-(n), and~\ref{RumiGAN_Compares}(t)-(u) further validate these observations through the convergence of FID scores and PR curves. While the converged models have relatively close FID scores, Rumi-LSGAN has the lowest FID score on both datasets. The PR curves indicate that Rumi-LSGAN generalized better to the desired held-out class. The poor PR performance of LSGAN may be attributed to mode collapse. Additional results are included in Appendix~\ref{AppSec: C10CelebA}.
\begin{table}[!t]
\begin{center}
\begin{tabular}{P{3.2cm}||P{1.5cm}|P{1.5cm}|P{1.8cm}|P{1.8cm}|P{2.3cm}}
\toprule
 & LSGAN &  ACGAN &   TAC-GAN  & CGAN-PD &  {\bfseries Rumi-LSGAN}  \\[2pt]
 \midrule[0.5pt]
 MNIST (Averaged) &  128.55 &  127.91 &  121.3 & 151.4 &  {\bfseries118.93} \\[2pt] 
 CelebA (Averaged) &  226.25 &  243.95 &  213.6 & 281.51 &  {\bfseries169.34}\\[2pt] 
CIFAR-10 (Averaged) &  231.02 &  355.06 &  275.43 & 262.3 &  {\bfseries217.15}  \\
\bottomrule
\end{tabular}
\end{center}
 \vspace{-0.5em}
 \caption{Comparison of FID scores on unbalanced datasets. Rumi-LSGAN has the best FID scores.}
  \label{Table:FID_scores}
\end{table}
\section{Conclusions}
We introduced the Rumi formulation for GANs, which aims at generating samples from a desired positive class by learning to ignore the negative class, but with the training relying on samples coming from both the positive and the negative classes. We showed that the Rumi-SGAN generator can learn any weighted combination of the two classes, whereas the Rumi-LSGAN can specifically latch on to the positive class distribution. Validations on standard datasets such as MNIST, Fashion MNIST, CelebA and CIFAR-10 showed that Rumi-LSGAN outperforms the baseline and auxiliary-classifier based models particularly on learning a held-out class in unbalanced datasets. The validations serve to strengthen the philosophy that teaching the generator to avoid fitting to the undesired class distribution indeed boosts its performance on fitting to the desired class distribution. The Rumi framework is applicable to all known GAN flavors and regularized variants such as DRAGAN \cite{DRAGAN17} and not just the standard GAN or LSGAN. For learning high-resolution images, models such as BigGAN \cite{BIGGAN18}, StyleGANs \cite{StyleGAN19}, or progressive growing of GANs (PGGANs) \cite{PGGAN18} can also be reformulated within the Rumi framework.


\newpage

\section{Acknowledgement}
This work is supported by the Qualcomm Innovation Fellowship 2019.

\section{Broader Impact}
Neural network based image classification and supervised image generation are data-intensive tasks. These models, when trained on unbalanced data, for instance, facial image datasets with insufficient racial diversity~\cite{Diversity17}, tend to inherit the implicit biases present in the data. DeVries \textit{et al.}~\cite{Objects19} demonstrated the existence of such biases with sub-par classification performance on images of objects coming from countries with low-income households, compared with those coming from countries with high-income households. The proposed approach could be used to address the imbalance in the data distribution and cater to the under-represented classes. The optimized generator in the proposed Rumi approach could be used to generate more samples of the under-represented classes and thus make the machine learning task more {\it inclusive}. The negative aspect is that one could flip the whole argument around and redefine the desired and undesired classes to serve exactly the opposite objective of favoring a certain class at the expense of the other. While the proposed approach could be used to alleviate the imbalances and biases present in the dataset, it cannot overcome the biases of the data scientist.
\par

\bibliography{RumiGAN_NIPS2020}
\bibliographystyle{ieeetr}

\appendix

\section*{Appendices}
We provide additional analytical and experimental results to support the content presented in the main manuscript. Appendix~\ref{AppSec: Rumi-LSGAN} of this document presents a detailed discussion on the Rumi-LSGAN. In Appendix~\ref{AppSec: Rumi-fGAN}, we impose the Rumi formulation on \(f\)-GANs~\cite{fGAN16}, and in Appendix~\ref{AppSec: Rumi-WGAN}, we generalize it to include integral probability metric (IPM) based GANs such as the Wasserstein GAN (WGAN)~\cite{WGAN17}. In Appendix~\ref{AppSec: MNIST}, we compare the performance of Rumi-SGAN, Rumi-LSGAN, and Rumi-WGAN on the MNIST dataset. Finally, in Appendix~\ref{AppSec: C10CelebA}, we provide additional results and comparisons on CelebA and CIFAR-10 datasets.
\par


\section{Rumi-LSGAN} \label{AppSec: Rumi-LSGAN}
 Recall the Rumi-LSGAN formulation:
 \begin{align*}
\loss^{LS}_{D} &= \betap \E_{\x \sim \pdp} \left[  \left( D(\x) - \bp \right)^2 \right] + \betan \E_{\x \sim \pdn} \left[  \left( D(\x) - \bn \right)^2 \right] + \E_{\x \sim \pg} \left[ \left( D(\x) - a\right)^2 \right], \\ 
\loss^{LS}_{G} &= \betap \E_{\x \sim \pdp} \left[  \left( D^*(\x) - c \right)^2 \right] + \betan \E_{\x \sim \pdn} \left[  \left( D^*(\x) - c \right)^2 \right] + \E_{\x \sim \pg} \left[ \left( D^*(\x) - c\right)^2 \right], 
\end{align*}
where \(\loss_G^{LS}\) must be subjected to the integral and non-negativity constraints
\begin{align*}
 \Omega_{\pg}: \int_{\mathcal{X}\subseteq\mathbb{R}^n} \pg(\x)~ \mathrm{d}\x = 1, \quad \quad   \text{and} \quad \quad  \Phi_{\pg} : \pg(\x) \geq 0, ~\forall~ \x,
\end{align*}


respectively. Incorporating the constraints using a Lagrangian formulation and expressing the expectations as integrals results in
 \begin{align}
\loss^{LS}_{D} &= \int_{\mathcal{X}} \left(\betap  \left( D(\x) - \bp \right)^2\pdp + \betan \left( D(\x) - \bn \right)^2\pdn +  \left( D(\x) - a\right)^2\pg\right)~\mathrm{d}\x \label{LS Rumi-GAN discriminator integral},~\text{and}\\
\loss^{LS}_{G} &=   \int_{\mathcal{X}} \left(\left( D^*(\x) - c \right)^2  ( \betap\pdp + \betan\pdn + \pg ) + \lambda_p\pg + \mu_p(\x)\pg\right)~\mathrm{d}\x - \lambda_p~, \label{LS Rumi-GAN generator integral}
\end{align}
where \(\lambda_p\) and \(\mu_p(\x)\) are the Karush-Kuhn-Tucker (KKT) multipliers. The cost functions in Equations~\eqref{LS Rumi-GAN discriminator integral} and \eqref{LS Rumi-GAN generator integral} have to be optimized with respect to the discriminator $D$ and the generator $\pg$, respectively. Since it is a functional optimization problem, we have to invoke the {\it Calculus of Variations}. If the integrand is continuously differentiable everywhere over the support  \( \mathcal{X} =  \mathrm{Supp}(\pdp)\cup\mathrm{Supp}(\pdn)\cup\mathrm{Supp}(\pg)\subseteq \mathbb{R}^n\), then the optimization of the integral cost carries over point-wise to the integrand. The optimal discriminator turns out to be
\begin{align*}
D^*(\x) &= \frac{\bp \betap \pdp + \bn\betan\pdn + a\pg }{\betap\pdp + \betan\pdn + \pg}.
\end{align*}
The optimal generator turns out to be the solution to a quadratic equation with roots
\begin{align}
\pg^* &= \betap \left( \frac{ \pm (a-\bp)}{\sqrt{(a-c)^2 + \lambda_p + \mu_p}} - 1\right) \pdp + \betan \left( \frac{ \pm(a-\bn)}{\sqrt{(a-c)^2 + \lambda_p + \mu_p}} - 1\right) \pdn.
\label{pg quadratic}
\end{align}
The positive root is the minimizer of the cost. The optimal KKT multiplier \( \mu_p^*(\x)\) is the one that satisfies the complementary slackness condition \(  \mu_p^* \pg^* = 0,~\forall~\x\in\mathcal{X}\), and the feasibility criterion \(\mu_p^*(\x) \leq 0,~\forall \x \in \mathcal{X}\). Since \(\pg^*\) is a weighted mixture of \(\pdp\) and \(\pdn\), the support of the solution could be split into three regions corresponding to: (i) \(\pdp>0\) and \(\pdn>0\); (ii) \(\pdp>0\) and \(\pdn=0\); and (iii) \(\pdp=0\) and \(\pdn>0\). Enforcing the complementary slackness condition in each region, with suitable assumptions on the class labels and weights such as \( \ds a \leq \frac{\bp+\bn}{2}\) and \( \betap >0, \betan>0\), yields \(\mu_p^*(\x) = 0,~\forall \x \in \mathcal{X}\), as the only feasible solution with a consistent value for \(\lambda_p^*\) over all three regions constituting the support. Enforcing the integral constraint \(\Omega_{\pg}\) and solving for \( \lambda_p^*\) gives
\begin{align*}
\lambda_p^* = \left( \frac{\betap(a-\bp) + \betan(a-\bn)}{1 + \betap + \betan} \right)^2 - (a-c)^2.
\end{align*}
Substituting for \(\mu_p^*\) and \(\lambda_p^*\) in~\eqref{pg quadratic} yields the optimal generator:
\begin{align*}
\pg^*(\x) &=  \betap \eta^+ \pdp(\x) + \betan \eta^- \pdn(\x),
\end{align*} 
where \( \eta^+ = \left( \frac{(1+\betan)(a - \bp) - \betan(a-\bn)}{\betap(a - \bp) + \betan(a-\bn)}\right)\), and \( \eta^- = \left( \frac{(1+\betap)(a-\bn) - \betap(a - \bp)}{\betap(a - \bp) + \betan(a-\bn)}\right).\)
The solutions for \(\mu_p^*\) and \(\lambda_p^*\) are also intuitively satisfying as the optimal generator obtained under these conditions automatically satisfies the non-negativity constraint. With this choice of the class labels and weights, the solution obtained by applying only the integral constraint \(\Omega_{\pg}\) automatically satisfies the non-negativity constraint. \par
As a special case, observe that setting \(\ds \betap = \frac{a - \bn}{\bn - \bp} \) results in \( \eta^- = 0\). Subsequently, any choice of \(a, \bp, \bn,~\text{and}~\betan\) such that \(\betap\eta^+ = 1\) gives \(\pg^* = \pdp\). Similarly, setting \(\ds \betan = \frac{a - \bp}{\bp - \bn} \) and \(\betan\eta^- = 1\)  yields \( \pg^* = \pdn\). Lastly, when \( \ds a = \frac{\bp + \bn}{2} \), we have $$ \ds \pg^* = \left( \frac{\betap( 1 - 2\betan)}{\betap + \betan} \right) \pdp + \left( \frac{\betan( 1 + 2\betap )}{\betap + \betan} \right) \pdn,$$ which is a mixture of \(\pdp\) and \(\pdn\). 

\section{The Rumi-\(f\)-GANs} \label{AppSec: Rumi-fGAN}
The Rumi formulation is extendable to all the \(f\)-GAN variants presented by Nowozin \textit{et al.} \cite{fGAN16}. Consider a GAN that minimizes the generalized divergence metric
\begin{align*}
D_f(p_g,p_d) = \int_\mathcal{X} p_d(\x) f\left(\frac{p_g(\x)}{p_d(\x)} \right) \mathrm{d}\x \text{,}
\end{align*}
which was shown to be equivalent to minimizing the loss functions \cite{fGAN16}:
\begin{align*}
\mathcal{L}^f_D &=  -\mathbb{E}_{\x \sim \pd} [T(\x)] + \mathbb{E}_{\x \sim \pg}[f^{c}(T(\x))], \quad \text{and}\\
\mathcal{L}^f_G &= \mathbb{E}_{\x \sim \pd}[T(\x)]-\mathbb{E}_{\x \sim \pg}[f^{c}(T(\x))],
\end{align*}
with respect to \(D(\x)\) and \(\pg\), respectively, where \(f^c\) is the Fenchel conjugate of the divergence \(f\), and \(T(\x) = g(D(\x))\) with \(g\) explicitly representing the activation function employed at the output of the discriminator network. The Rumi-\(f\)-GAN minimizes \(D_f^+ = D_f(\pg,\pdp)\), while maximizing \(D_f^- = D_f(\pg,\pdn)\), weighted by \(\gammap\) and \(\gamman\), respectively, which is given as
\begin{align}
\loss^{Rf}_D &=  - \gammap \mathbb{E}_{\x \sim \pdp} [T(\x)] + \gamman \mathbb{E}_{\x \sim \pdn} [T(\x)] +  \mathbb{E}_{\x \sim \pg}[f^{c}(T(\x))],~\text{and}  \label{f-GAN discriminator}\\
 \loss^{Rf}_G &= \gammap \mathbb{E}_{\x \sim \pdp} [T(\x)] - \gamman \mathbb{E}_{\x \sim \pdn} [T(\x)] - \mathbb{E}_{\x \sim \pg}[f^{c}(T(\x))], \label{f-GAN generator}
\end{align}
where \( \gammap - \gamman = 1 \), and \( \loss^{Rf}_G\) is subjected to the integral and non-negativity constraints \( \Omega_{\pg} \) and \( \Phi_{\pg}\), respectively. As shown in Appendix~\ref{AppSec: Rumi-LSGAN}, we enforce only \( \Omega_{\pg} \), showing that the optimal solution satisfies \( \Phi_{\pg}\) without having to enforce it explicitly.

\begin{lemma}{\textbf{The optimal Rumi-\(f\)-GAN}: \label{Optimal Rumi-fGAN}} Consider the \(f\)-GAN optimization problem defined through Equations \eqref{f-GAN discriminator} and \eqref{f-GAN generator}. Assume that the weights satisfy \( \gammap - \gamman = 1 \). The optimal discriminator and generator are the solutions to 
\begin{align}
\fracpartial{f^c}{T} &= \frac{\gammap \pdp - \gamman \pdn}{\pg}, \text{and} \label{D_star Rumi-fGAN}\\
 f^c(T^*) &= \lambda_p,
\label{pg_star Rumi-fGAN}
\end{align} 
respectively, where \(T^* = g(D^*)\). \end{lemma}
\textbf{Proof:} The integral Rumi-\(f\)-GAN costs can be optimized as in the case of Rumi-LSGAN. Optimization of the integrand in Equation~\eqref{f-GAN discriminator} yields the necessary condition that the optimal discriminator \(D^*(\x)\) must satisfy, which gives us Equation~\eqref{D_star Rumi-fGAN}. Differentiating the integrand in \eqref{f-GAN generator}, we get
\begin{align*}
\left( (\gammap \pdp - \gamman \pdn) - \pg \fracpartial{f^c}{T^*} \right) \fracpartial{T^*}{\pg} - f^c(T^*) + \lambda_p = 0.
\end{align*}
Enforcing the condition given in \eqref{D_star Rumi-fGAN} yields the necessary condition that \(\pg^*\) must satisfy.  \qedwhite
  \par
Table~\ref{Table. Rumi-fGANs} shows the optimal discriminator and generator functions obtained in the case of each of the \(f\)-GAN variants presented by Nowozin \textit{et al.} \cite{fGAN16}. We observe that all \(f\)-GANs learn a weighted mixture of \(\pdp\) and \(\pdn\), akin to the Rumi-SGAN presented in the main manuscript. The weights are a function of \(\gammap, \gamman\), and \(\lambda_p\). The optimal \(\lambda_p^*\) in each case can be found by enforcing the integral constraint \(\Omega_{\pg}\). Also observe that the Rumi-Pearson-\(\chi^2\) GAN in Table~\ref{Table. Rumi-fGANs} is a special case of the Rumi-LSGAN, where \( a-c = 1 \), \( (a - \bp) =  \sqrt{\lambda_p +1}  + 1  \) and \( (a - \bn) =  \sqrt{\lambda_p +1}  - 1\). \par
Finally, we note that setting \(\gammap \in [0,1] \) in addition to \(\gammap - \gamman =1 \) result in solutions for all \(f\)-GANs that automatically satisfy the non-negativity constraint.

\begin{table}[t]
\caption{Rumi formulation of \(f\)-GANs: The optimal Rumi discriminator \(D^*\) and generator \(\pg^*\) for a given \(f\)-GAN defined through its activation function \(g\) and Fenchel conjugate \(f^c\) of the divergence metric. $T=g(D)$.  } \label{Table. Rumi-fGANs}
\begin{center}
\begin{tabular}{p{2cm}||p{2cm}|p{2cm}||p{3.5cm}|p{3.5cm}}
\toprule 
\small{\(f\)-divergence} & \(g(D)\) & \(f^c(T)\) &\( D^*(\x)\) & \(\pg^*(\x)\)   \\ \midrule 
\small{ Kullback-Leibler (KL)} & \(D\) & \( e^{T-1} \) & \(  1 + \log \left( \frac{\gammap\pdp - \gamman \pdn  }{\pg}\right) \) & \(  \frac{\gammap}{\log(\lambda_p)} \pdp - \frac{\gamman}{\log(\lambda_p)} \pdn \) \\[10pt]
\small{ Reverse KL} & \(-e^{-D}\) & \( -1-\log(-T) \) & \(  \log \left( \frac{\gammap\pdp - \gamman \pdn  }{\pg}\right) \) & \(  \frac{\gammap}{e^{\lambda_p+1}} \pdp - \frac{\gamman}{e^{\lambda_p+1}} \pdn \) \\[10pt]
\small{ Pearson-\(\chi^2\)} & \(D\) & \( \frac{1}{4}T^2+T \) & \(  2 \left( \frac{\gammap\pdp - \gamman \pdn  -\pg }{\pg}\right) \) & \(  \frac{\gammap}{\sqrt{\lambda_p+1}} \pdp - \frac{\gamman}{\sqrt{\lambda_p+1}} \pdn \) \\[10pt]
\small{ Squared-Hellinger} & \(1 - e^{-D}\) & \( \frac{T}{1-T}\) & \(  \frac{1}{2}\log \left( \frac{\gammap\pdp - \gamman \pdn  }{\pg}\right) \) & \(  \frac{\gammap}{(\lambda_p+1)^2} \pdp - \frac{\gamman}{(\lambda_p+1)^2} \pdn \) \\[10pt]
\small{ SGAN} & \( \scriptstyle -\log(1-e^{-D})\) & \( \scriptstyle -\log(1 - e^T)\) & \(  \log \left( \frac{\gamman\pdn - \gammap \pdp  }{\pg}\right) \) & \(  \frac{\gammap \lambda_p}{1 - \lambda_p} \pdp - \frac{\gamman\lambda_p}{1 - \lambda_p} \pdn \) \\[10pt]
 \bottomrule
\end{tabular}
\end{center}
\end{table}

\section{Rumi-WGAN} \label{Rumi-WGAN} \label{AppSec: Rumi-WGAN}
All integral probability metric (IPM) based GANs~\cite{WGAN17,MMDGAN17,FisherGAN17} can be reformulated under the Rumi framework. As an example, we consider the Rumi flavor of the Wasserstein GAN (Rumi-WGAN). The Rumi-WGAN minimizes the earth-mover distance (EMD) between \(\pdp\) and \(\pg\), while maximizing the EMD between \(\pdn\) and \(\pg\). Both these terms can be independently brought to the Kantarovich-Rubinstein dual-form as in WGANs \cite{WGAN17}:
\begin{align*}
D^* &= \argmax_{D, \|D\|_L \leq 1} \left( \gammap\left( \E_{\x \sim \pdp} [D(\x)] - \E_{\x \sim \pg} [D(\x)] \right) + \gamman \left( \E_{\x \sim \pdn} [D(\x)] - \E_{\x \sim \pg} [D(\x)] \right) \right), \\
\pg^* &= \argmin_{\pg} \left( \gammap\left( \E_{\x \sim \pdp} [D(\x)] - \E_{\x \sim \pg} [D(\x)] \right) - \gamman \left( \E_{\x \sim \pdn} [D(\x)] - \E_{\x \sim \pg} [D(\x)] \right) \right),
\end{align*}
which directly result in the Rumi-WGAN costs:
\begin{align*}
\loss^W_D &= -\gammap \left( \E_{\x \sim \pdp} [D(\x)] - \E_{\x \sim \pg} [D(\x)] \right) - \gamman \left( \E_{\x \sim \pdn} [D(\x)] - \E_{\x \sim \pg} [D(\x)] \right), \text{ and} \\ 
\loss^W_G &= \gammap \left( \E_{\x \sim \pdp} [D^*(\x)] - \E_{\x \sim \pg} [D^*(\x)] \right) - \gamman \left( \E_{\x \sim \pdn} [D^*(\x)] - \E_{\x \sim \pg} [D^*(\x)] \right),
 \end{align*}
with the constraint that \(D(\x)\) is Lipschitz-1. Similar to WGAN-GP \cite{WGANGP17}, we enforce the Lipschitz constraint through the gradient penalty: \((\|\nabla_{\x}D(\x)\|_2 - 1)^2 \), which is evaluated in practice by a sum over points interpolated between \(\pdp\) and \(\pg\), and \(\pdn\) and \(\pg\):
\begin{align*}
\Omega_{GP}: \sum_{\hat{\x}} (\|\nabla_{\hat{\x}}D(\hat{\x})\|_2 - 1)^2 + \sum_{\tilde{\x}} (\|\nabla_{\tilde{\x}}D(\tilde{\x})\|_2 - 1)^2,
\end{align*}
where \( \hat{\x} = (1 - \xi)\x^g + \xi\x^+ \) and \( \tilde{\x} = (1 - \zeta) \x^g + \zeta \x^-\) such that \( \x^g \sim \pg, \x^+ \sim \pdp \), \( \x^- \sim \pdn\), and \(\xi, \zeta\) are uniformly distributed over \([0,1] \).

\section{Comparison of Rumi-GAN Variants} \label{AppSec: MNIST}
In this section, we compare the performance of Rumi-SGAN, Rumi-LSGAN, and Rumi-WGAN-GP with their baseline variants on the MNIST dataset. We consider the following test scenarios: 
\begin{enumerate}
    \item Even digits as the positive class;
    \item Overlapping positive and negative classes as described in the main manuscript (Section~\ref{Sec: MNIST_and_FMNIST});
    \item One vs. rest learning where the positive class comprises all samples from the digit class 5, with the rest of the digit classes representing the negative class data.
\end{enumerate}
Scenario 3 above simulates an important variant of learning from unbalanced data. \par

\textit{\textbf{Experimental Setup:}} The network architectures and hyper-parameters are as described in the main manuscript (Section 4). For Rumi-SGANs, we set \(\alphap = 0.8\) and \(\alphan = -0.2\). Rumi-LSGAN uses class labels \( (a,\bn,c,\bp) = (0,0.5,1,2) \) with weights \(\betap = 1\) and \(\betan = 0.5\). For Rumi-WGAN-GP, we use \(\gammap = 5\) and \(\gamman=1\). \par
\begin{figure}[H]
\begin{center}
  \begin{tabular}[b]{P{.5\linewidth}|P{.5\linewidth}}
  {\bf Baseline} & {\bf Rumi counterpart} \\[3pt]
    \includegraphics[width=1\linewidth]{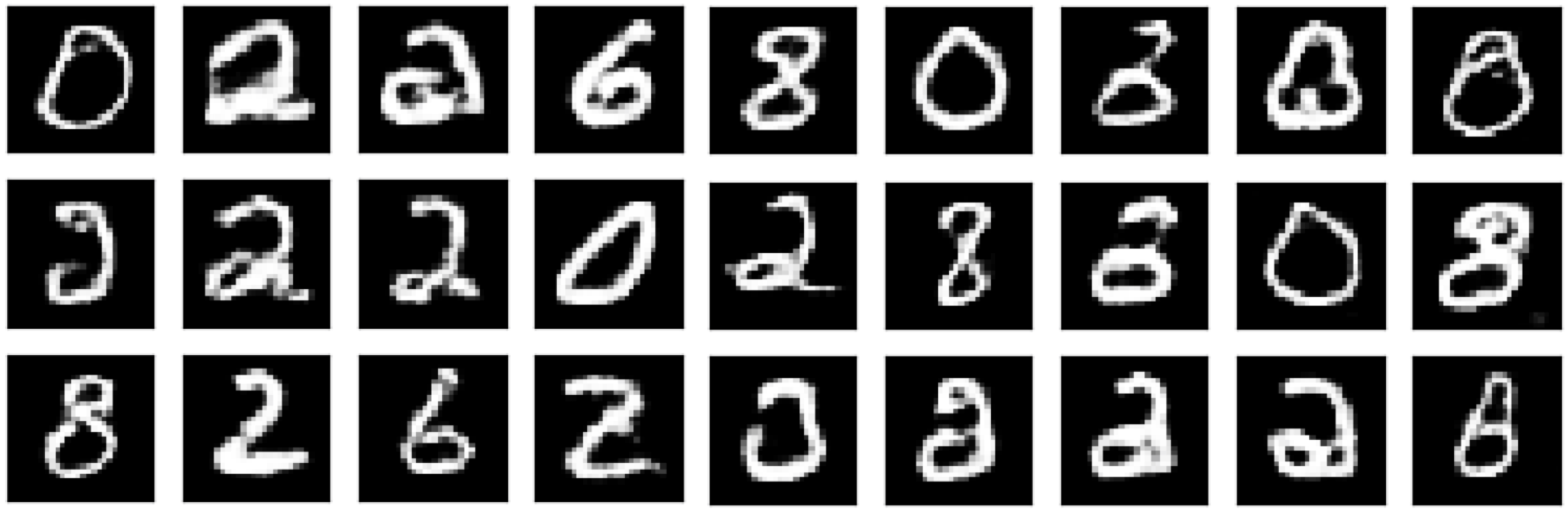} & 
     \includegraphics[width=1\linewidth]{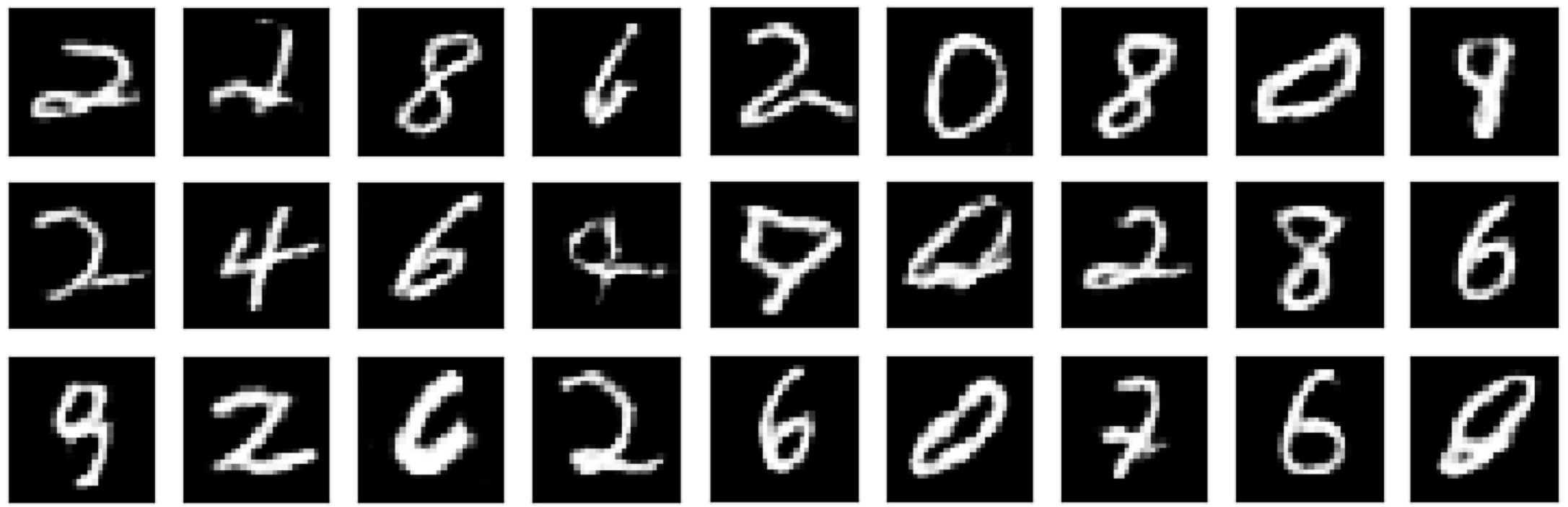}  \\[-1pt]
     (a) SGAN  & (b) Rumi-SGAN  \\[3pt] \midrule
     \includegraphics[width=1\linewidth]{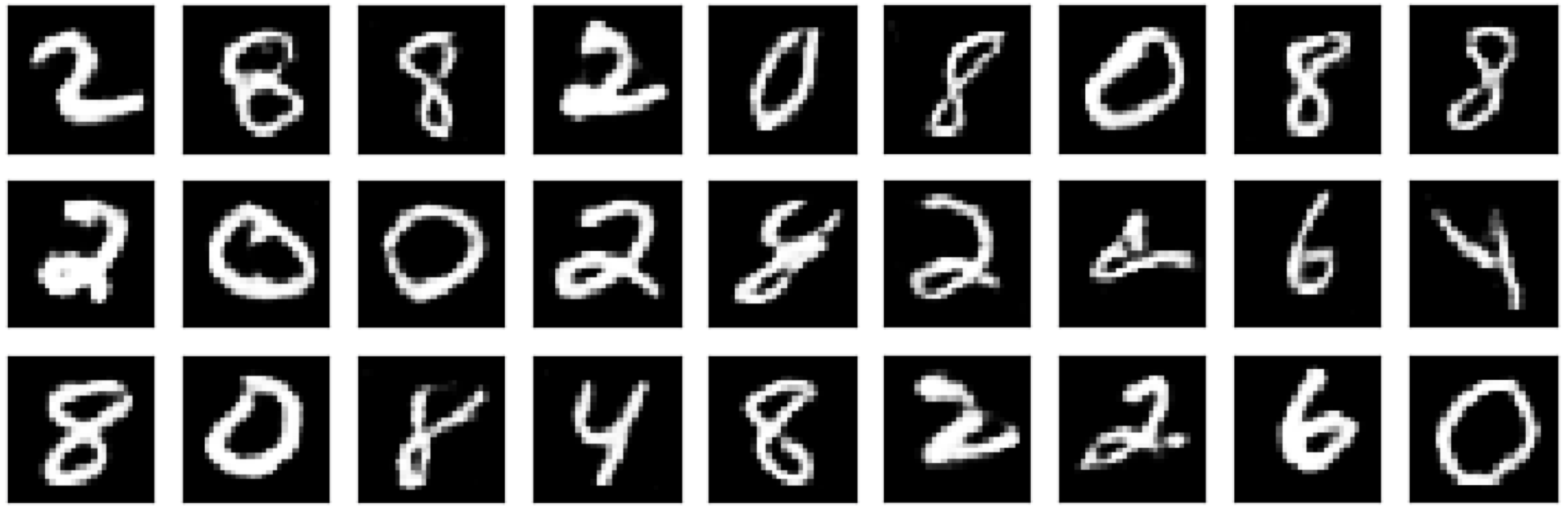} & 
     \includegraphics[width=1\linewidth]{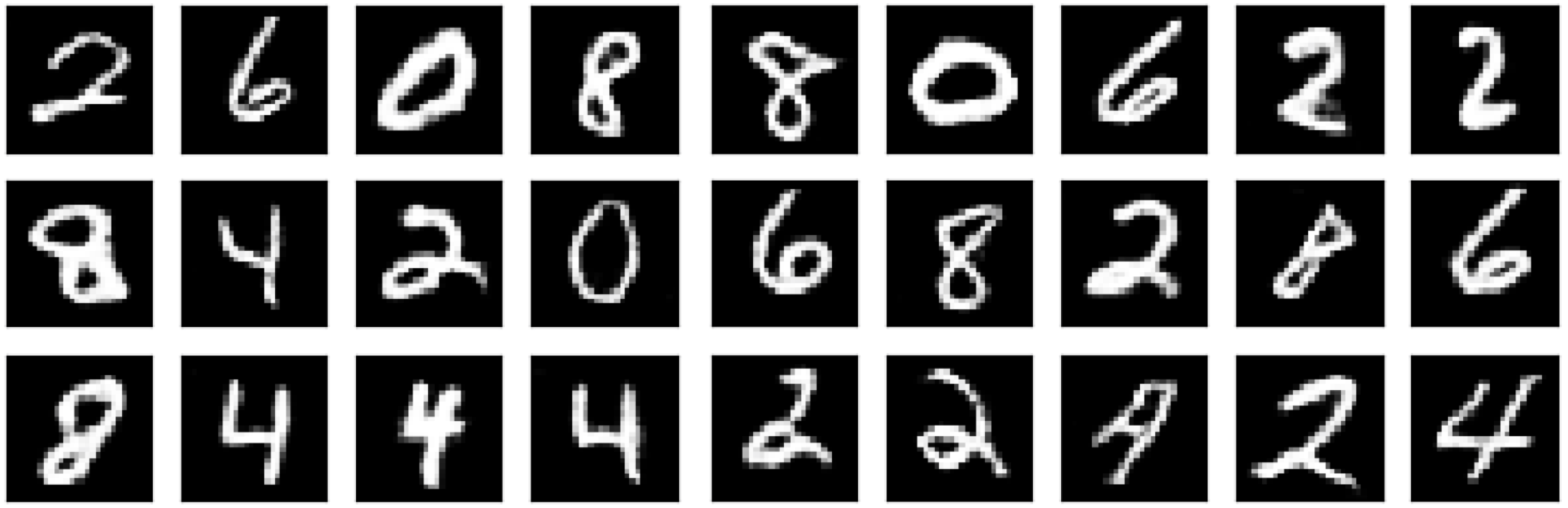}  \\[-1pt]
     (c) LSGAN  & (d) Rumi-LSGAN  \\[3pt] \midrule
     \includegraphics[width=1\linewidth]{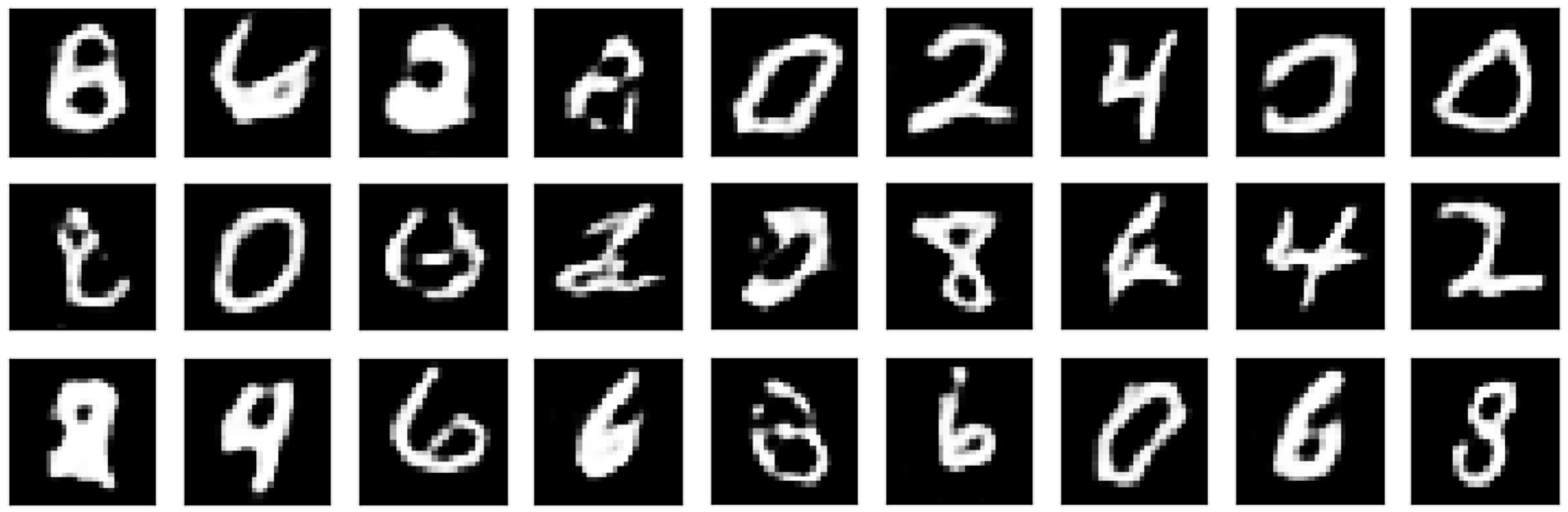} &
     \includegraphics[width=1\linewidth]{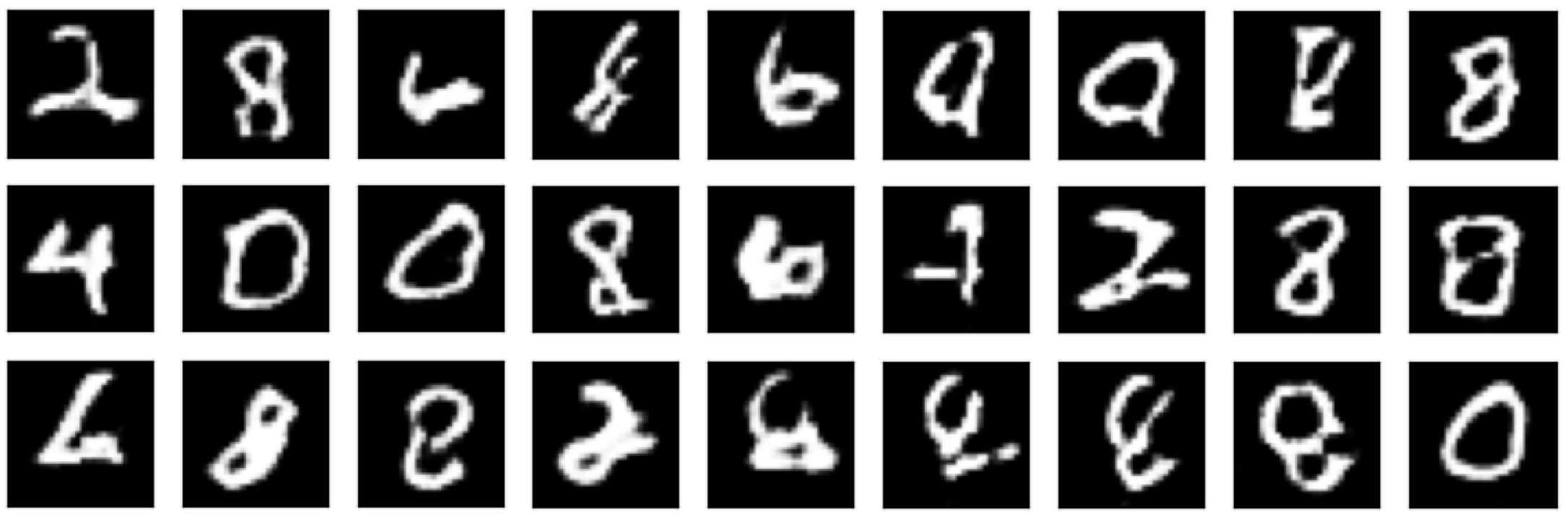} \\[-1pt]
	(e) WGAN  & (f) Rumi-WGAN \\[7pt]
      \end{tabular} 
  \caption[]{ Illustrating the strength of the {\it Rumi framework}. {\bf Scenario 1}: Training GANs to generate even digits from the MNIST dataset. The Rumi counterparts learn qualitatively better images than the corresponding baselines. The SGAN seems to have mode-collapsed unlike its Rumi counterpart.} 
  \vspace{-1em}
  \label{Even_MNIST}
  \end{center}
\end{figure}
\begin{figure}[H]
\begin{center}
  \begin{tabular}[b]{P{.5\linewidth}|P{.5\linewidth}}
  {\bf Baseline} & {\bf Rumi counterpart}  \\[3pt]
    \includegraphics[width=1\linewidth]{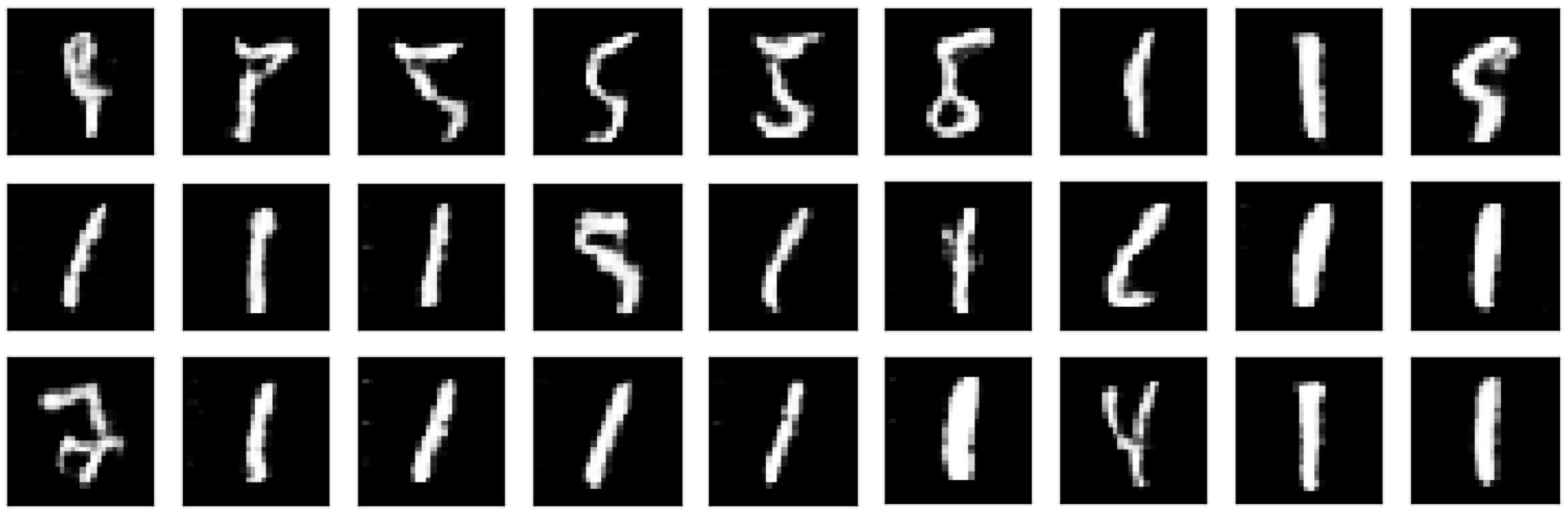} & 
     \includegraphics[width=1\linewidth]{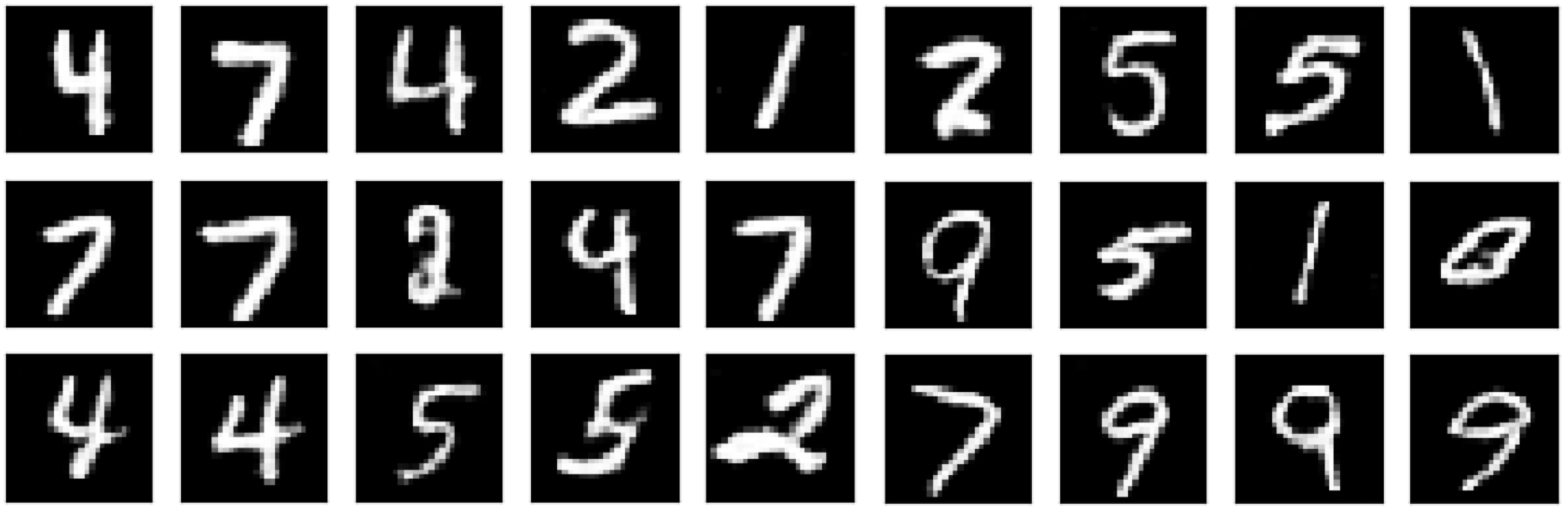}  \\[-1pt]
     (a) SGAN  & (b) Rumi-SGAN  \\[3pt] \midrule
     \includegraphics[width=1\linewidth]{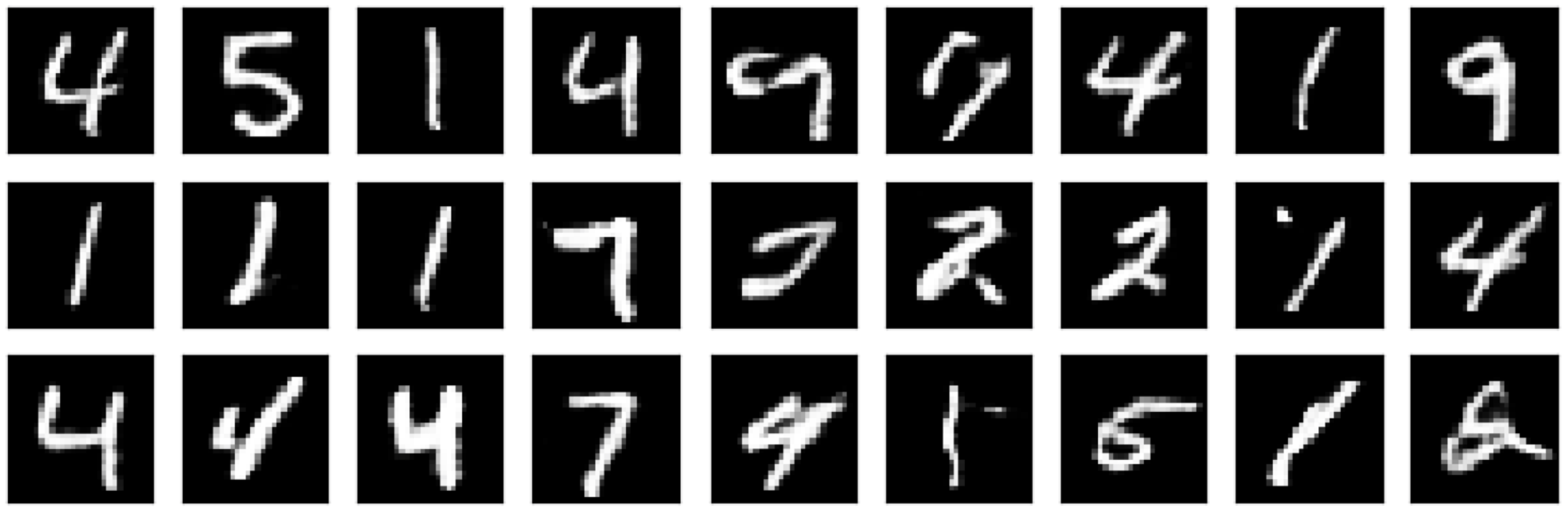} & 
     \includegraphics[width=1\linewidth]{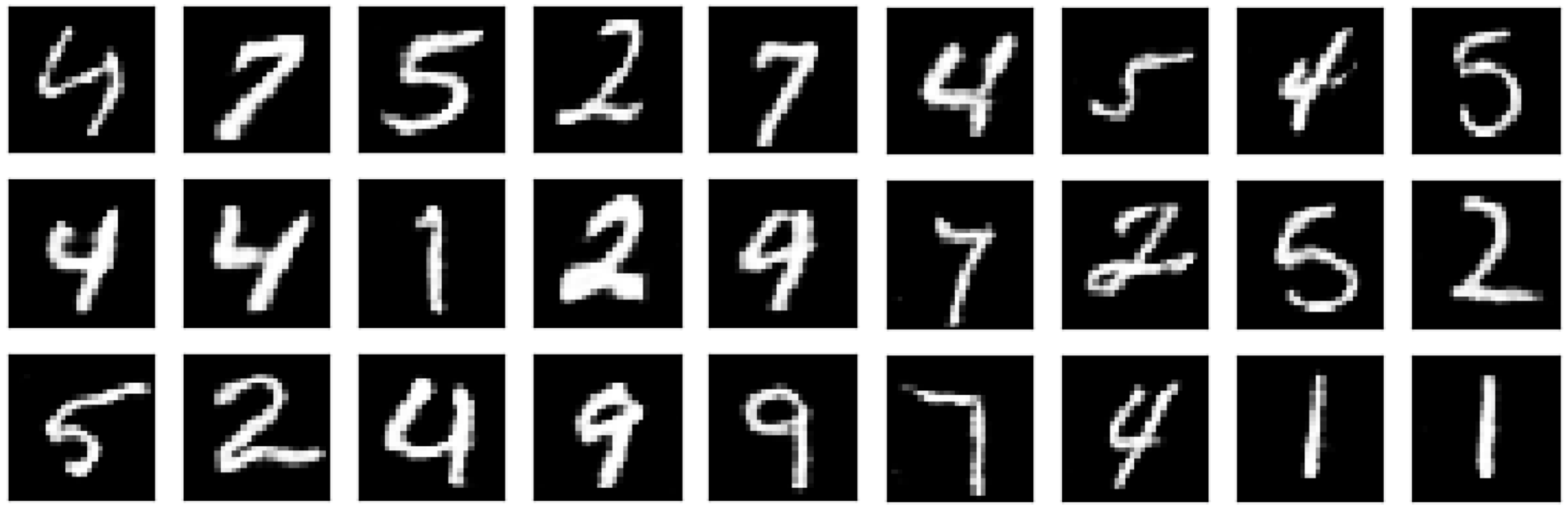}  \\[-1pt]
     (c) LSGAN  & (d) Rumi-LSGAN  \\[3pt] \midrule
     \includegraphics[width=1\linewidth]{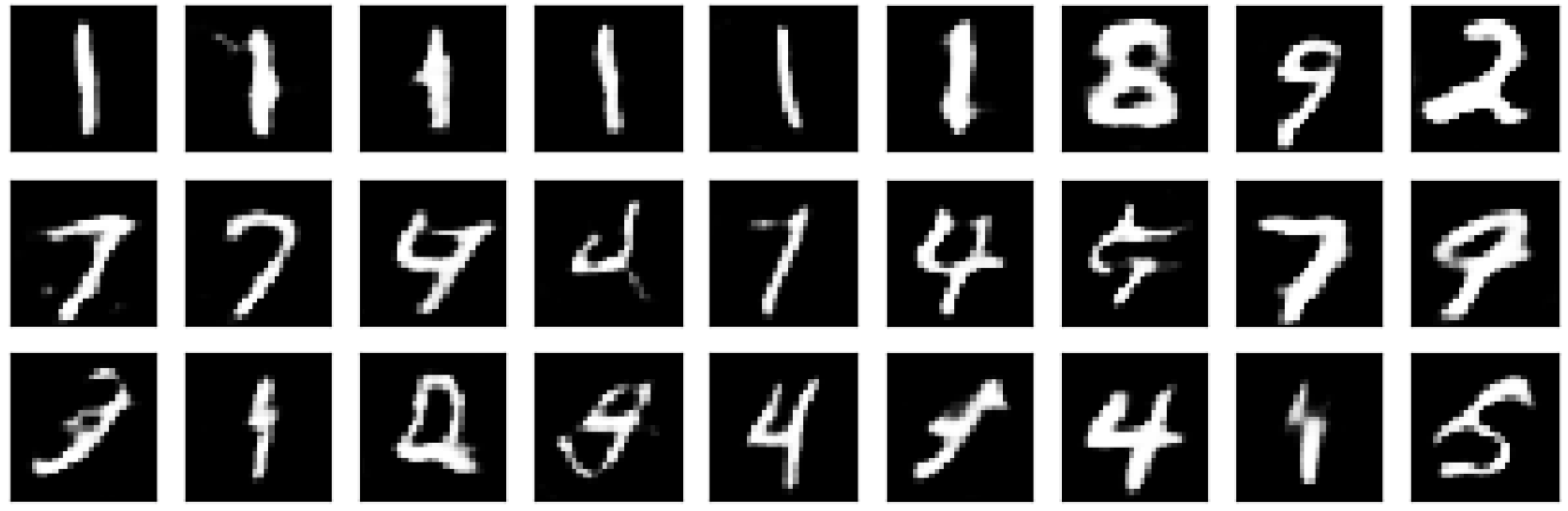} &
     \includegraphics[width=1\linewidth]{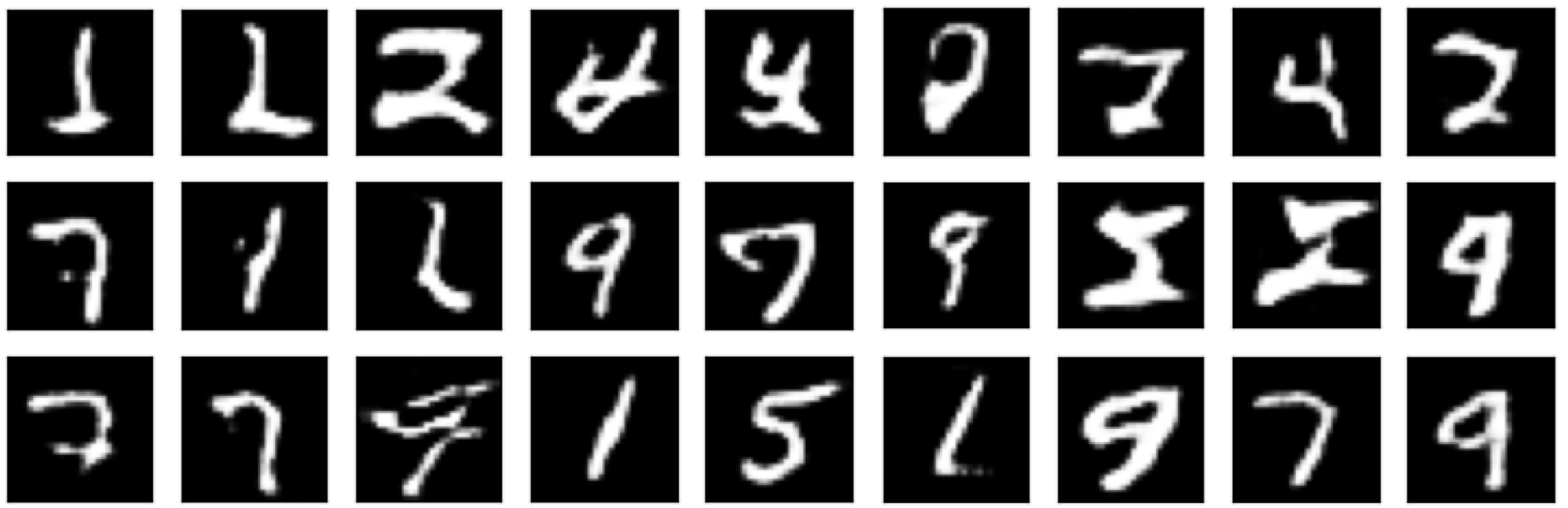} \\[-1pt]
	 (e) WGAN  & (f) Rumi-WGAN \\[7pt]
      \end{tabular} 
  \caption[]{ {\bf Scenario 2}: Training GANs on  overlapping MNIST classes. The Rumi counterparts generate visually better images than the baselines. The {\it mode-collapse} effect is prominent in the baselines unlike the Rumi counterparts.} 
  \vspace{-1em}
  \label{Random_MNIST}
  \end{center}
\end{figure}
\begin{figure}[H]
\begin{center}
  \begin{tabular}[b]{P{.5\linewidth}|P{.5\linewidth}}
  {\bf Baseline} & {\bf Rumi counterpart} \\
    \includegraphics[width=1\linewidth]{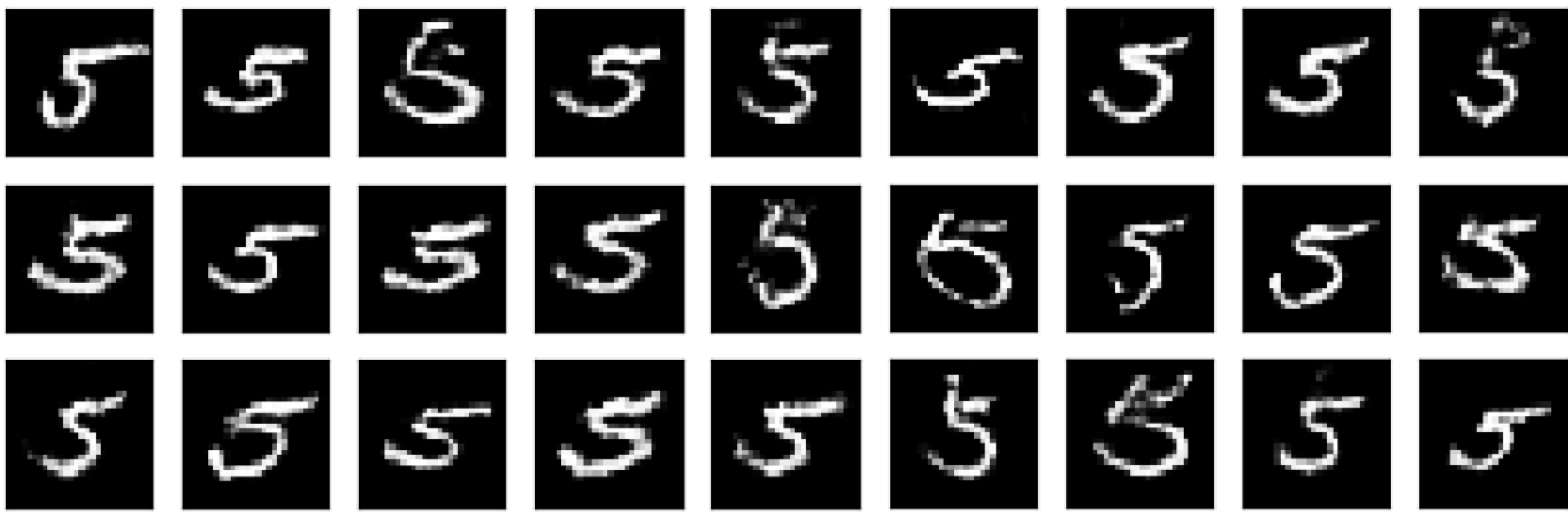} & 
     \includegraphics[width=1\linewidth]{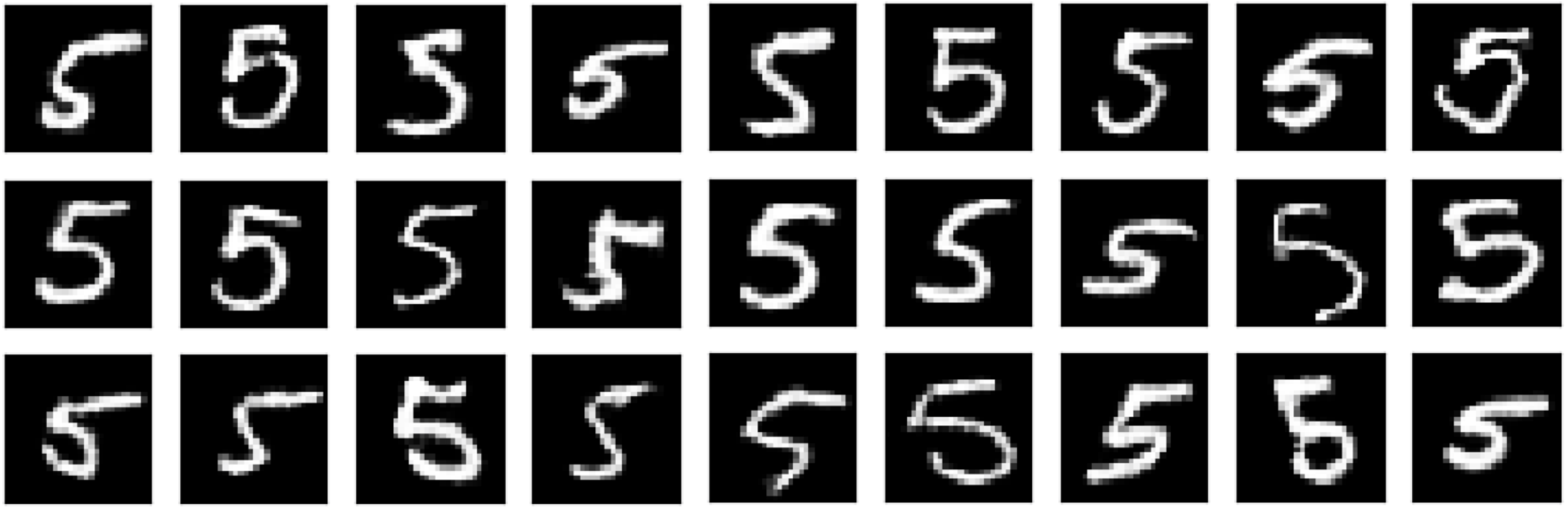}  \\[-1pt]
     (a) SGAN  & (b) Rumi-SGAN  \\[1pt] \midrule
     \includegraphics[width=1\linewidth]{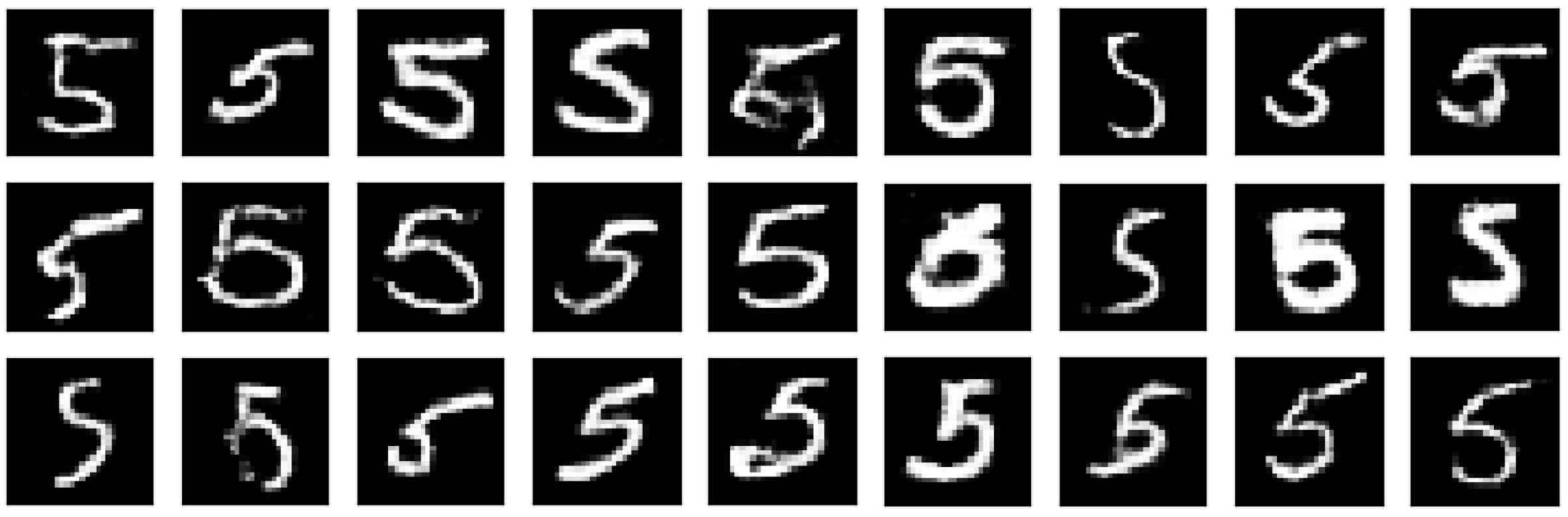} & 
     \includegraphics[width=1\linewidth]{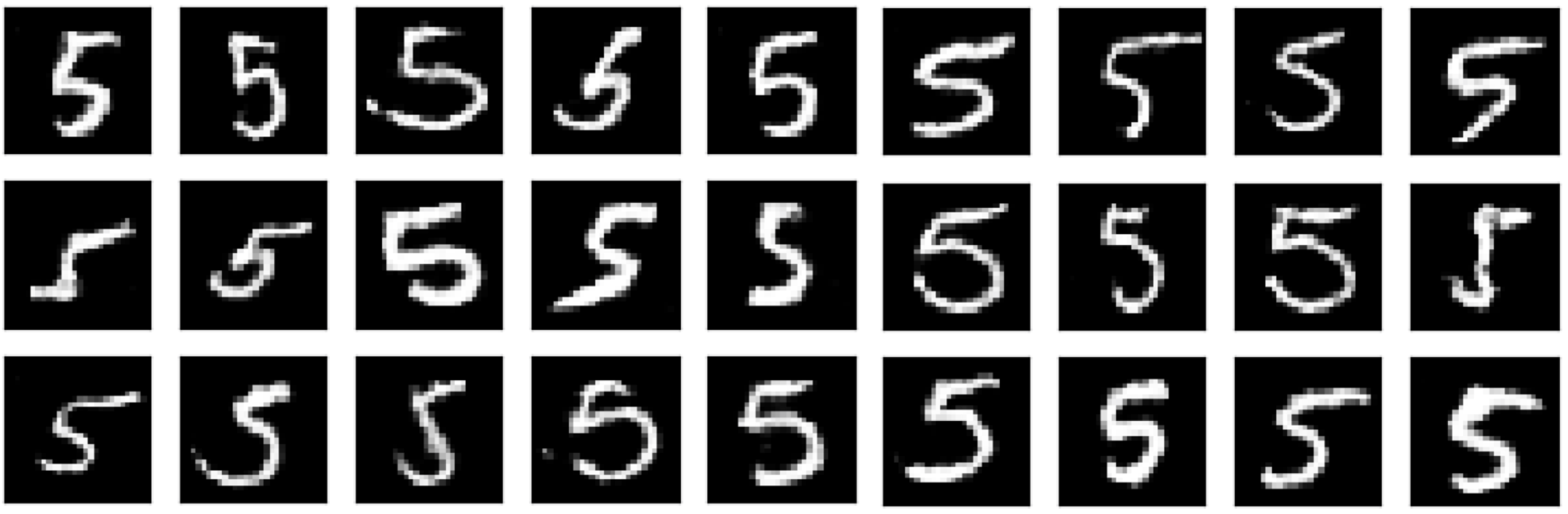}  \\[-1pt]
     (c) LSGAN  & (d) Rumi-LSGAN  \\[-1pt] \midrule
     \includegraphics[width=1\linewidth]{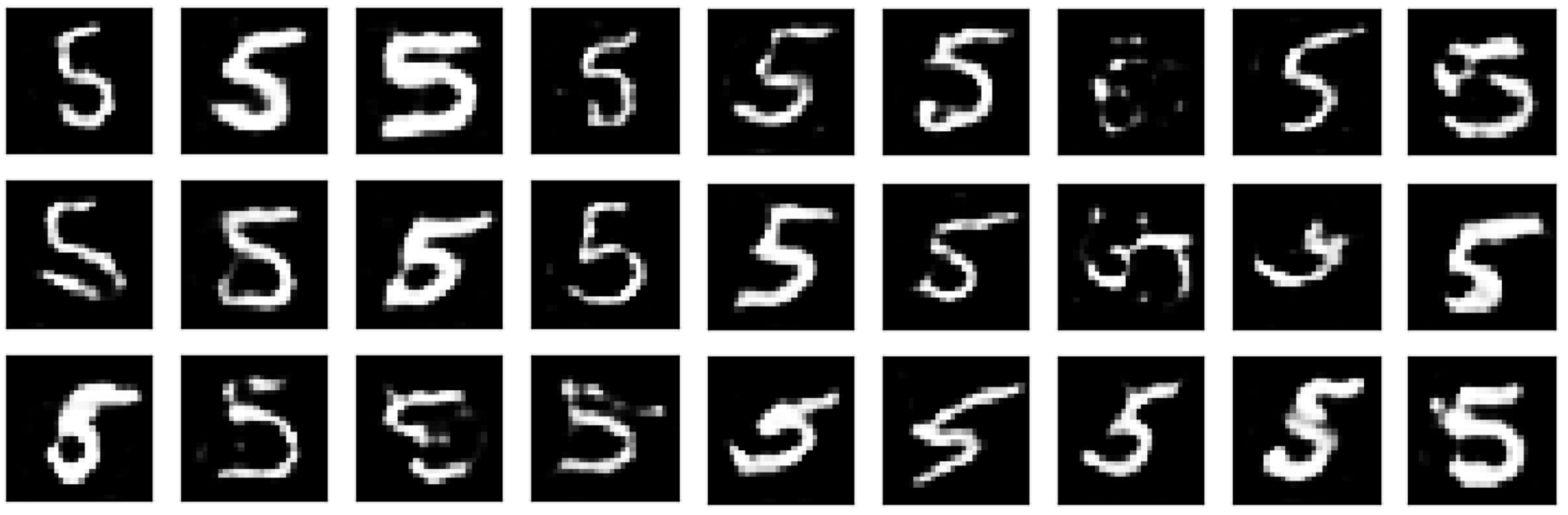} &
     \includegraphics[width=1\linewidth]{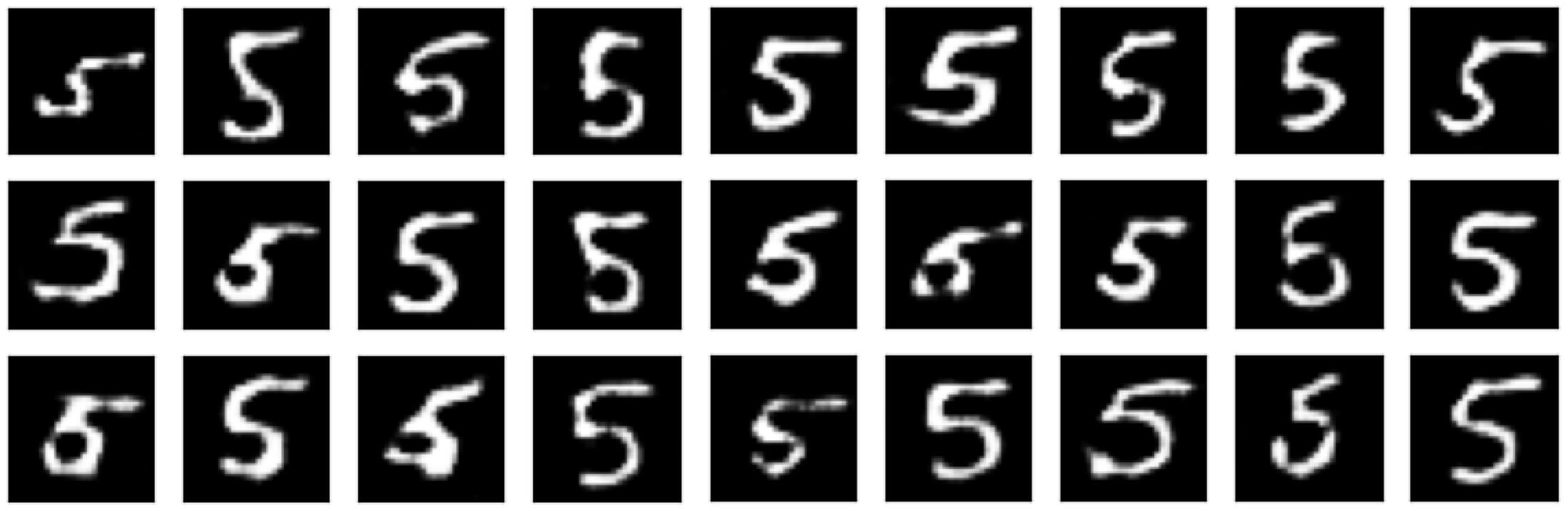} \\[-1pt]
	 (e) WGAN  & (f) Rumi-WGAN \\[-1pt]
      \end{tabular} 
  \caption[]{ {\bf Scenario 3}: Training GANs to learn the digit class 5. The Rumi variants generate sharper images compared with the corresponding baselines.}
  \vspace{-1em}
  \label{Single_MNIST}
  \end{center}
\end{figure}
\begin{figure}[h]
\begin{center}
  \begin{tabular}[b]{P{.32\linewidth}|P{.32\linewidth}|P{.32\linewidth}}
    Even MNIST & Overlapping MNIST & Single digit MNIST\\
    \includegraphics[width=0.98\linewidth]{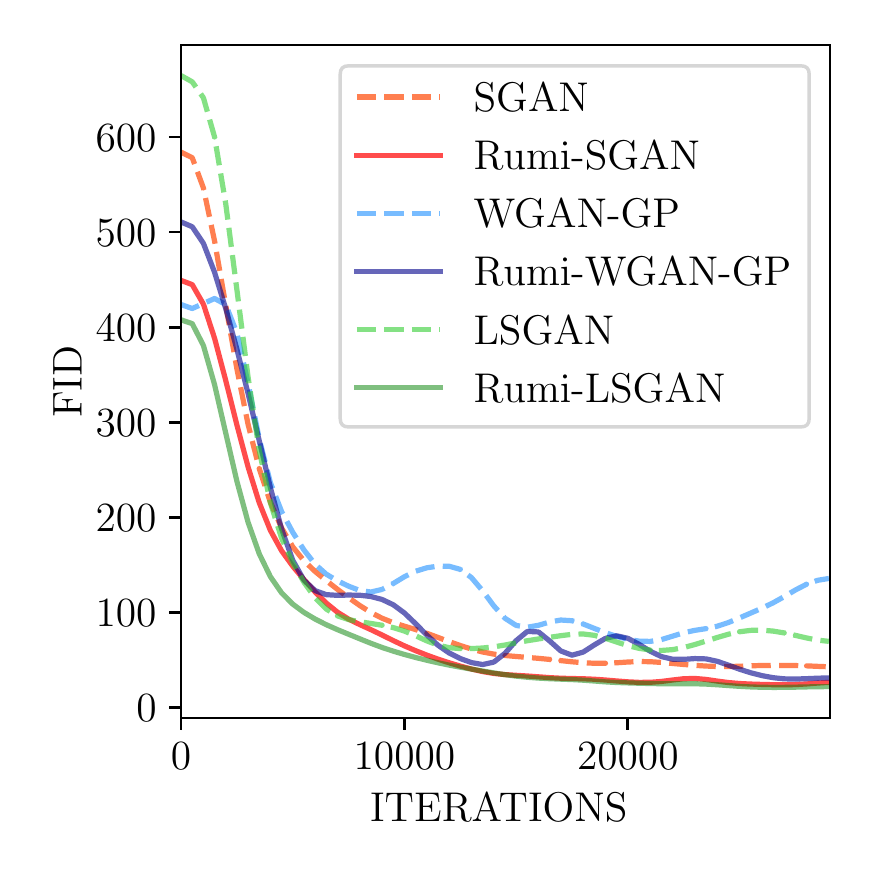} & 
    \includegraphics[width=0.98\linewidth]{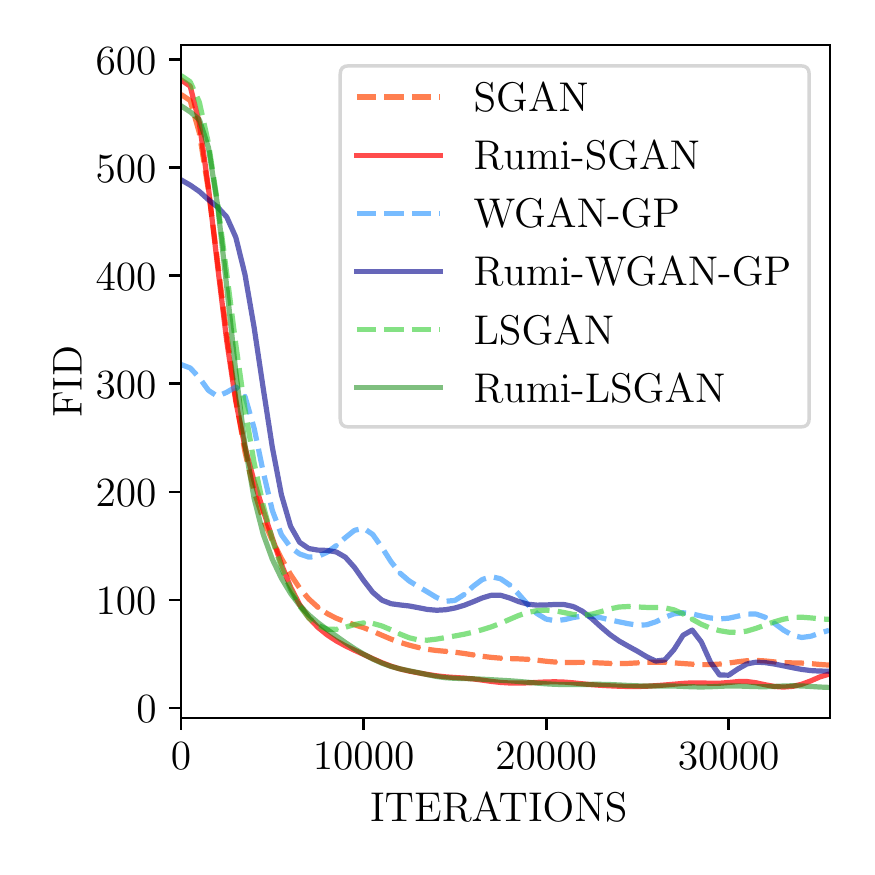} &
    \includegraphics[width=0.96\linewidth]{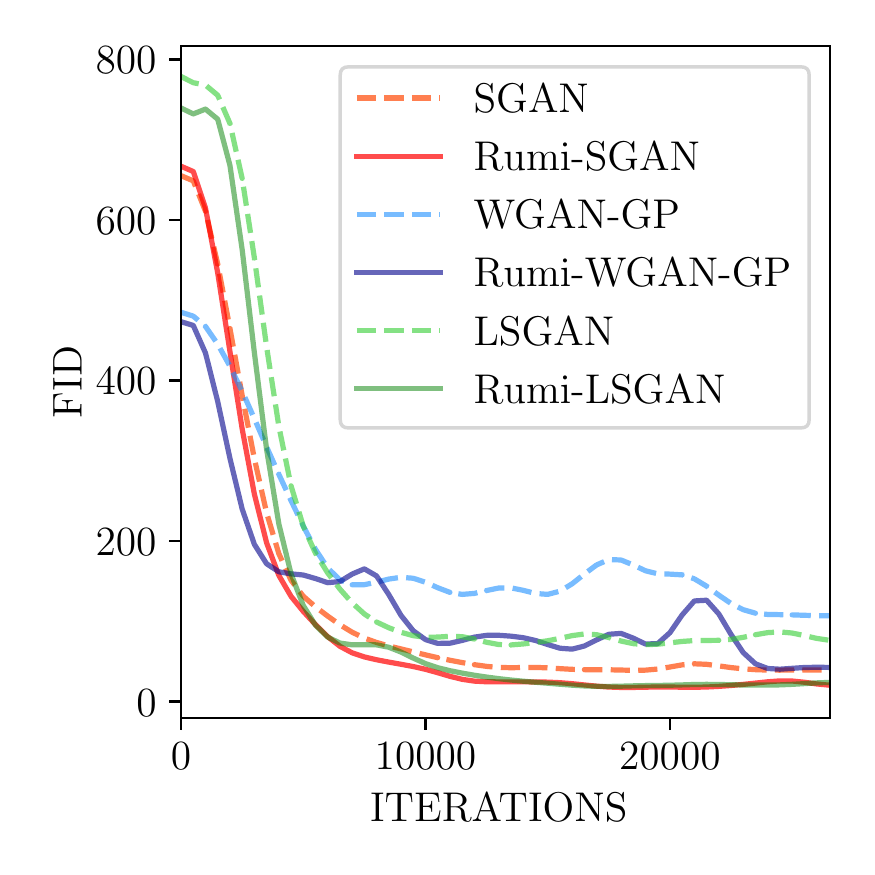} \\[-1pt]
    (a)  & (c) & (e) \\[-1pt]
    \includegraphics[width=0.98\linewidth]{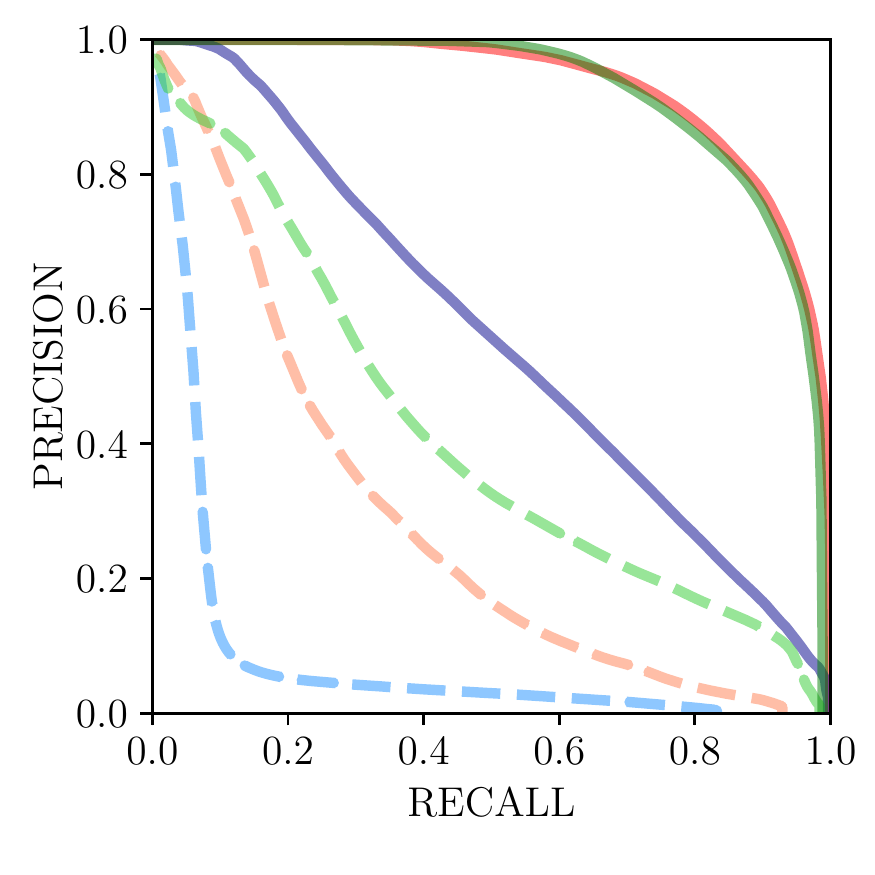} & 
    \includegraphics[width=0.98\linewidth]{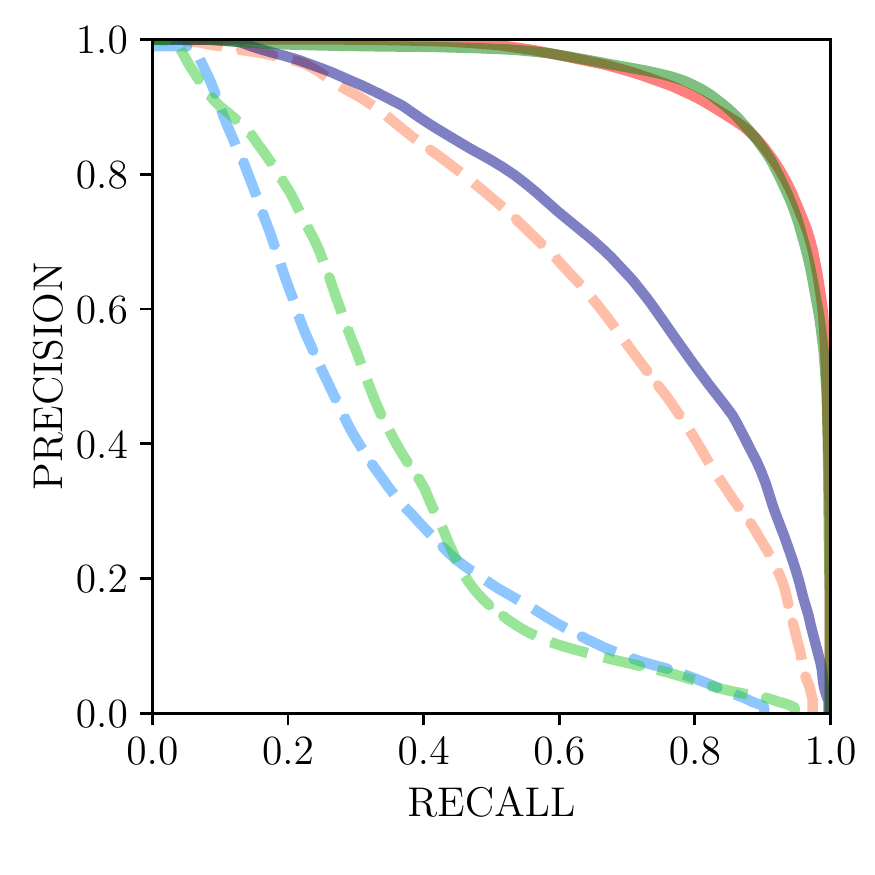} & 
    \includegraphics[width=0.98\linewidth]{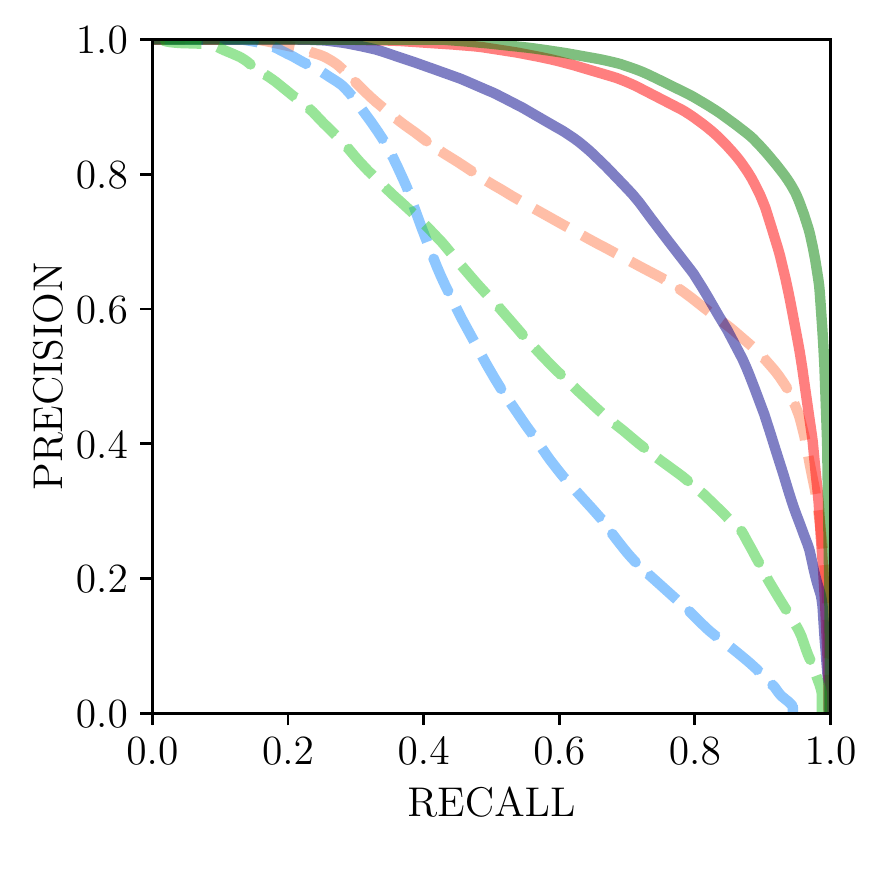}  \\[-1pt]
    (b)  & (d) & (f) \\[-1pt]
  \end{tabular} 
  \caption[]{(\includegraphics[height=0.014\textheight]{Rgb.png} Color Online) Comparison of FID vs. iterations and PR curves for various GANs trained on the MNIST dataset with positive class data being: (a) \& (b) Even numbers; (c) \& (d) Overlapping subsets; and (e) \& (f) Single digit class 5. Rumi variants possess better precision and recall characteristics, and also achieve better FID values than the baseline models.} 
  \vspace{0em}
  \label{MNIST_FID_PR}
  \end{center}
\end{figure}

\textit{\textbf{Results:}} Figures~\ref{Even_MNIST}, \ref{Random_MNIST}, and \ref{Single_MNIST} show the samples generated by the various GANs under consideration. From Figure~\ref{Even_MNIST}, we observe that the Rumi formulation always results in an improvement in the visual quality of the images generated. In the case of SGAN, the baseline approach experienced mode collapse (Fig.~\ref{Even_MNIST}(a)), while its Rumi counterpart (Fig.~\ref{Even_MNIST}(b)) learnt the target distribution accurately. Observe that the Rumi-SGAN and Rumi-WGAN variants learn to generate a few random samples from the negative class as well --- this validates our claim that these variants always learn a mixture of the positive and negative class densities. Similar improvements in the visual quality of images are also seen in Scenario 2, which considers overlapping classes (Fig.~\ref{Random_MNIST}), or Scenario 3, which considers {\it one vs. the rest} learning (Fig.~\ref{Single_MNIST}). The Fr\'echet inception distance (FID) plots and precision-recall (PR) curves in Figure~\ref{MNIST_FID_PR} quantitatively validate our findings. The Rumi variants achieve a higher precision and recall, and saturate to better FID values than their respective baselines. Also, Rumi-SGAN and Rumi-LSGAN show equivalent performance. As training the LSGAN is relatively more stable than training the SGAN, we prefer the Rumi-LSGAN to Rumi-SGAN when considering complex datasets such as CelebA.


\vspace{1em}

\section{Validation on CIFAR-10 and CelebA Datasets} \label{AppSec: C10CelebA}
We now present additional experimental results on CIFAR-10 and CelebA datasets. The experimental setup is identical to that explained in the main manuscript (Sections~\ref{Sec:Experimental_val} and~\ref{Sec:Unbalanced_data}). All the models are trained for \(10^5\) iterations. CelebA images are rescaled to \( 32 \times 32 \times 3\) unless stated otherwise.

\textit{\textbf{Experimental Setup:}} On the CIFAR-10 dataset, we consider the case of learning disjoint positive and negative classes. The {\it animal} classes are labelled  as positive, and the {\it vehicle} classes as negative. This is a scenario where the negative class samples have very little resemblance to those in the positive class. In the main manuscript, for CelebA, we presented results of learning the class of {\it female} celebrities as the positive class while setting the {\it males} to be the negative class. Here, we consider the converse situation --- positive class of {\it male celebrities} and the negative class of {\it female celebrities}. We also present additional experimental results on learning minority classes in CelebA. Splitting the data based on the \textit{bald} or the \textit{hat} class label of CelebA gives about 5,000 positive samples and 195,000 negative class samples in each case. \par
\begin{figure}[H]
\begin{center}
  \begin{tabular}[b]{P{.31\linewidth}|P{.31\linewidth}|P{.31\linewidth}}
    CIFAR-10 - \textit{Animals}& CelebA - \textit{Males} & CelebA - \textit{Bald}\\
    \includegraphics[width=0.98\linewidth]{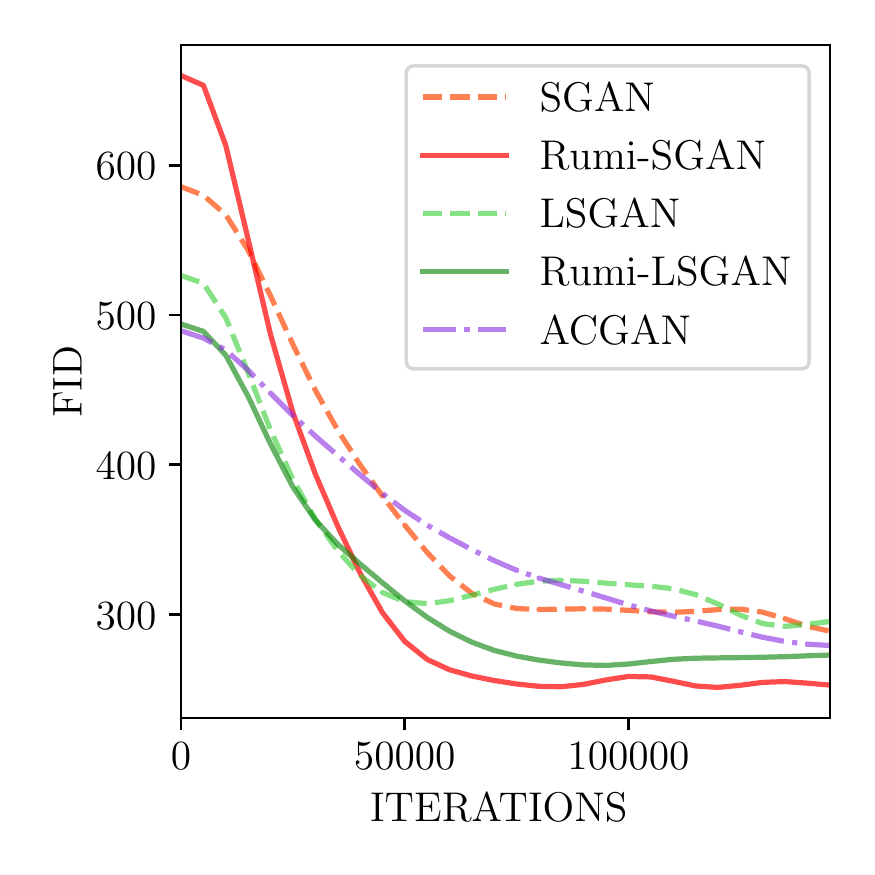} & 
    \includegraphics[width=0.93\linewidth]{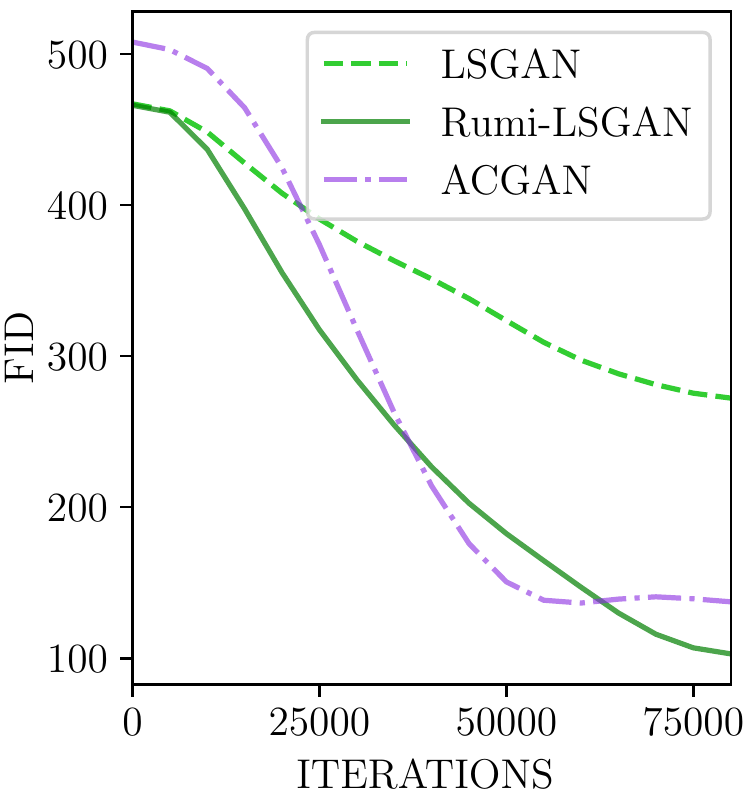} &
    \includegraphics[width=0.97\linewidth]{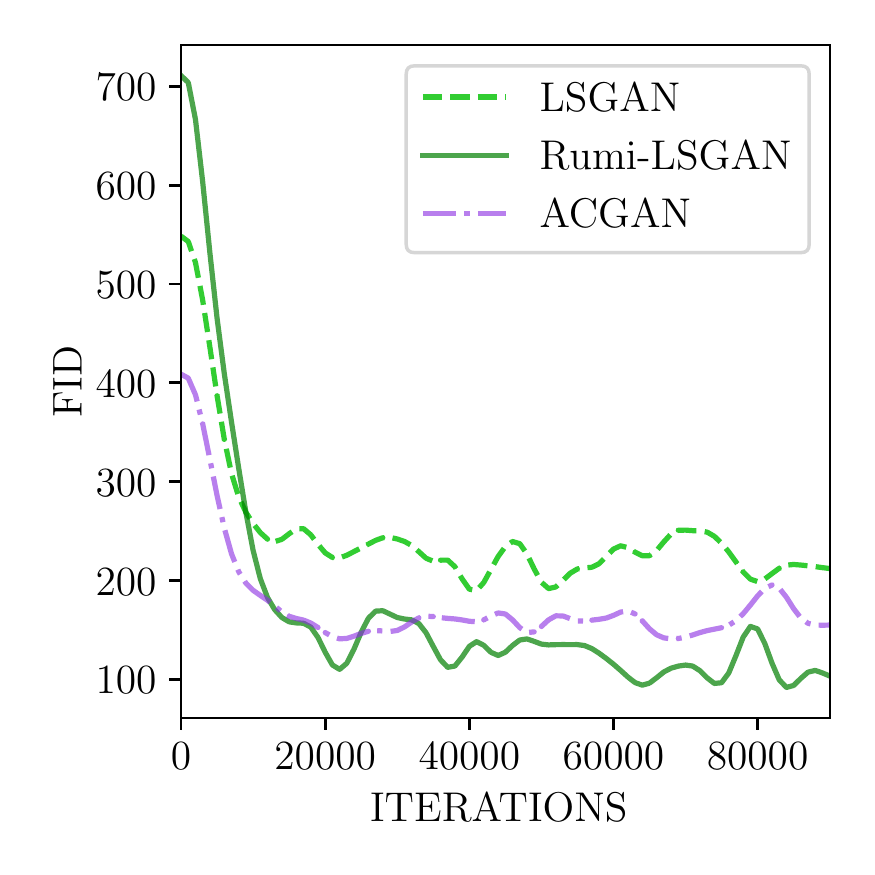} \\[1pt]
    (a)  & (c) & (e) \\[1pt]
    \includegraphics[width=0.98\linewidth]{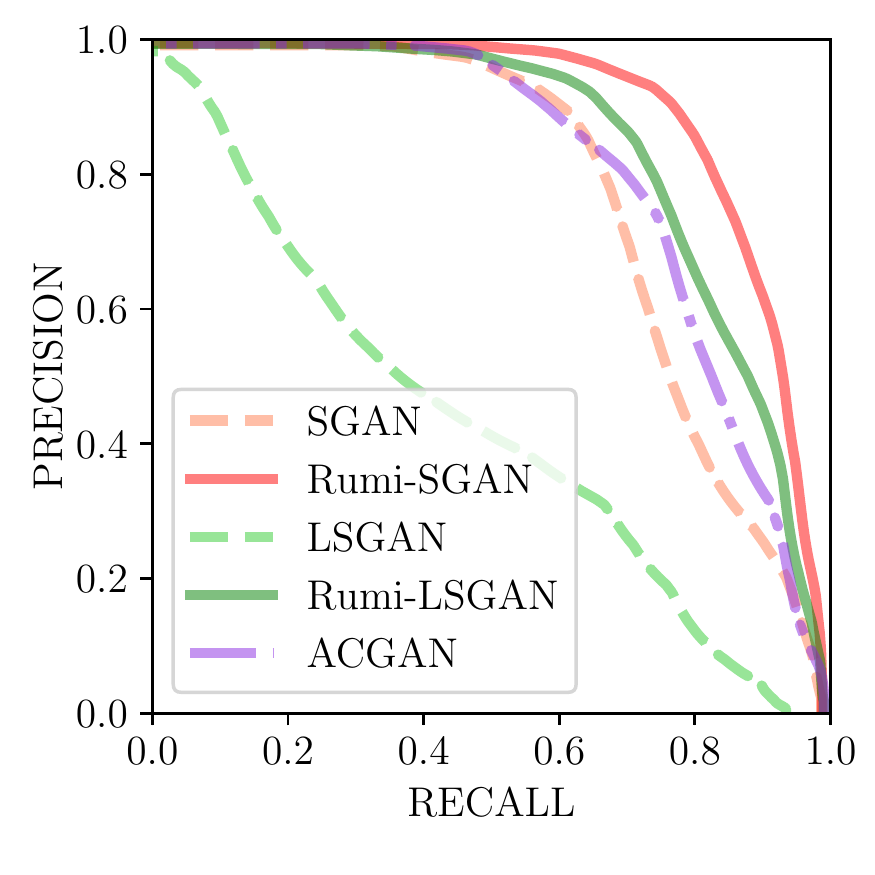} & 
    \includegraphics[width=0.98\linewidth]{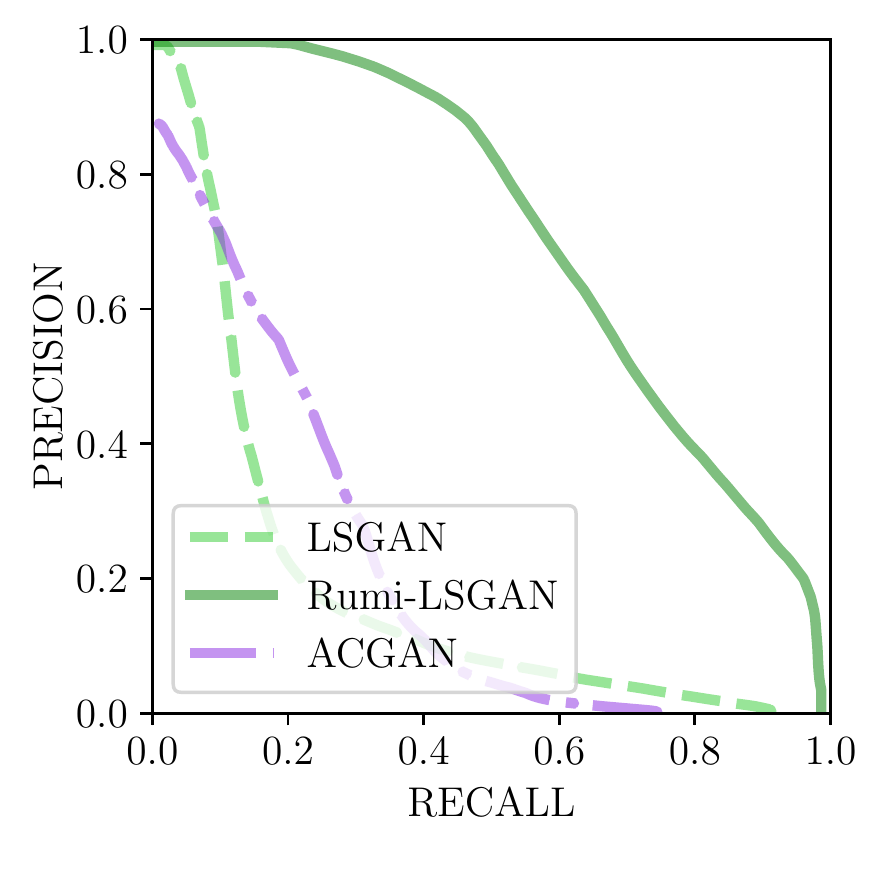} & 
    \includegraphics[width=0.98\linewidth]{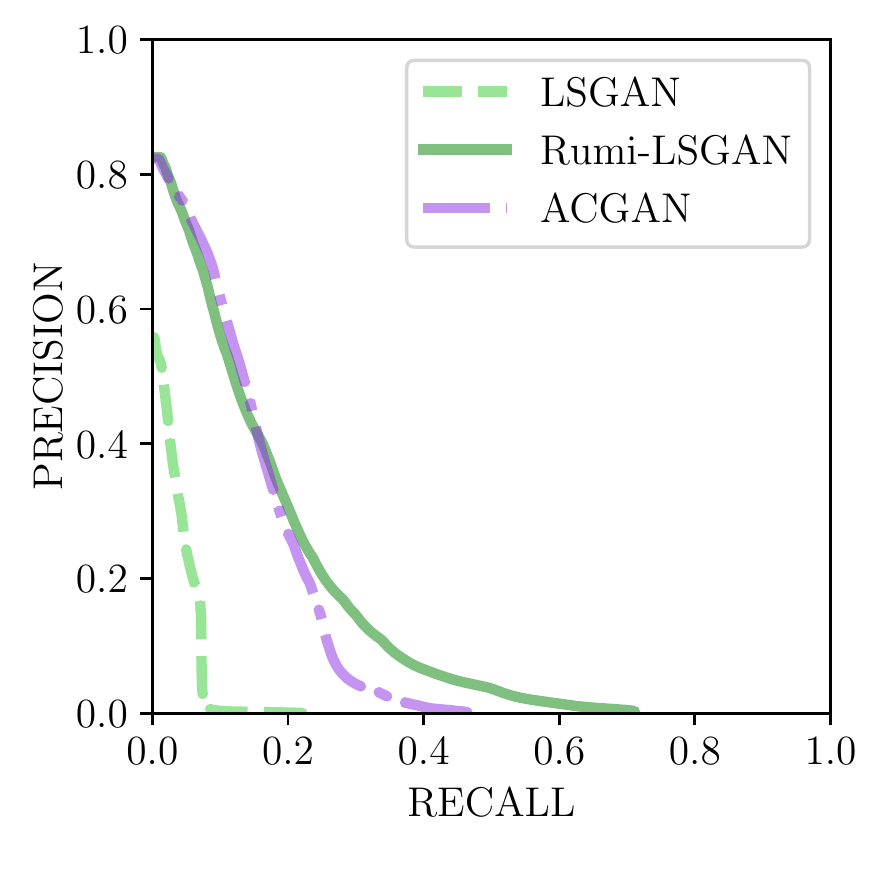}  \\[1pt]
    (b)  & (d) & (f) \\[1pt]
  \end{tabular} 
  \caption[]{(\includegraphics[height=0.014\textheight]{Rgb.png} Color Online) Comparison of FID vs. iterations and PR curves of various GANs when the training is carried out on: (a) \& (b) CIFAR-10 \textit{animal} classes; (c) \& (d) CelebA \textit{male} class; and (e) \& (f) CelebA \textit{bald} class. Rumi-LSGAN outperforms the baselines in terms of FID and PR values. } 
  \vspace{-1em}
  \label{CIFAR10_CelebA_FID_PR}
  \end{center}
\end{figure}
\begin{sidewaysfigure}
\begin{center}
  \begin{tabular}[b]{P{.19\linewidth}|P{.19\linewidth}||P{.19\linewidth}|P{.19\linewidth}||P{.19\linewidth}}
  SGAN & Rumi-SGAN & LSGAN & Rumi-LSGAN & ACGAN \\
    \includegraphics[width=0.94\linewidth]{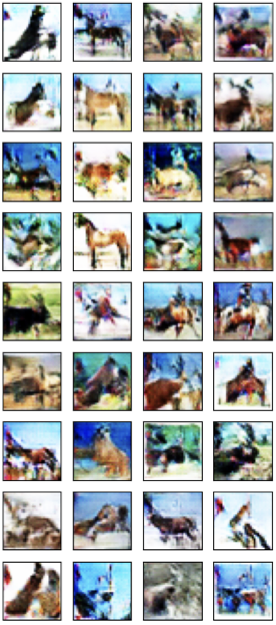} &
    \includegraphics[width=0.94\linewidth]{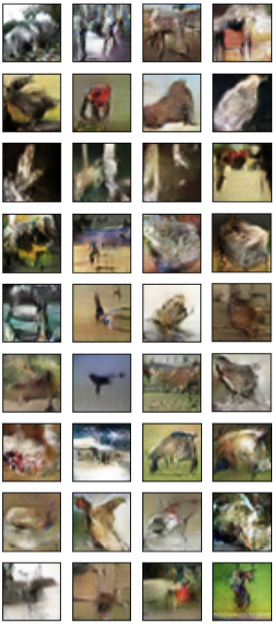} &
     \includegraphics[width=0.94\linewidth]{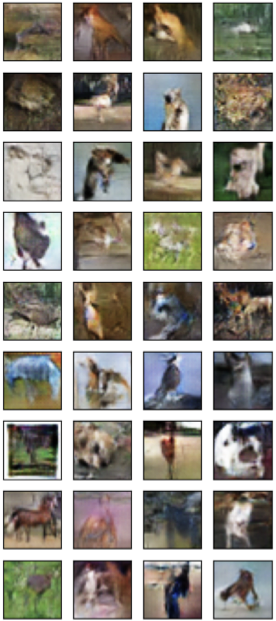} &
     \includegraphics[width=0.94\linewidth]{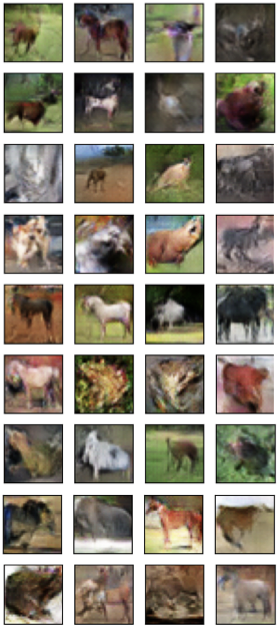} &
     \includegraphics[width=0.94\linewidth]{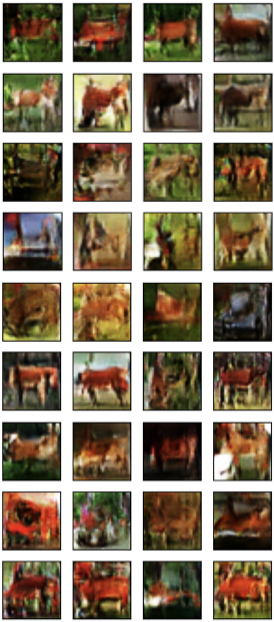} \\[-1pt]
 (a)  & (b)  & (c) & (d)  & (e) \\[1pt]
      \end{tabular} 
  \caption[]{(\includegraphics[height=0.014\textheight]{Rgb.png} Color Online) A comparison of the samples generated by the various GANs on \textit{animals} from CIFAR-10. The Rumi counterparts generate qualitatively better images than the baselines.}
  \vspace{-1.2em}
  \label{CIFAR10_Images}
  \end{center}
\end{sidewaysfigure}

\textit{\textbf{Results:}} Figure~\ref{CIFAR10_Images} presents the samples generated by SGAN, LSGAN, and their Rumi counterparts, alongside those generated by sampling an ACGAN trained purely on the {\it animal classes} of CIFAR-10. We observe that the Rumi-SGAN and Rumi-LSGAN generate images of superior visual quality. The Rumi variants also have better FID and PR performance than the baselines (Figures~\ref{CIFAR10_CelebA_FID_PR}(a) \& (b)). From Figure~\ref{CIFAR10_Images}(b), we observe that, in the context of Rumi-SGAN, no visually discernible features from the negative class of {\it vehicles} are present. This shows that although, in principle, the optimal Rumi-SGAN learns a mixture of the positive and negative classes, in practice, on real-world image datasets such as CIFAR-10, the influence of the negative class in the mixture is negligible. \par
\begin{sidewaysfigure}
\begin{center}
  \begin{tabular}[b]{P{.32\linewidth}|P{.32\linewidth}|P{.32\linewidth}}
  LSGAN & ACGAN & Rumi-LSGAN \\
    \includegraphics[width=0.94\linewidth]{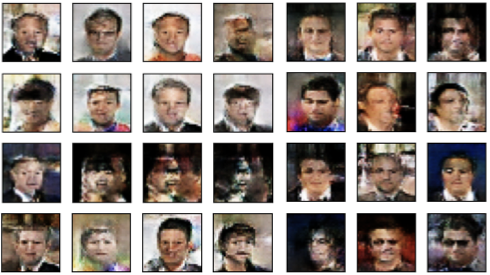} &
    \includegraphics[width=0.94\linewidth]{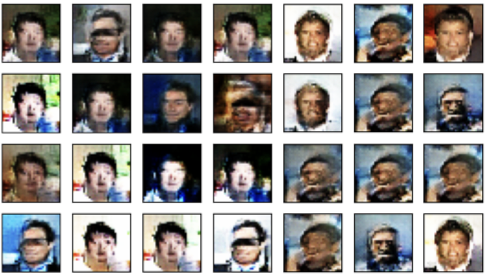} &
    \includegraphics[width=0.94\linewidth]{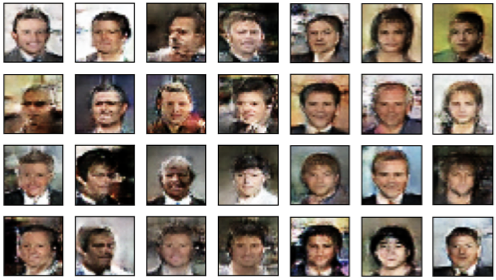} \\[-1pt]
    (a)  & (b)  & (c) \\[1pt]
        \includegraphics[width=0.94\linewidth]{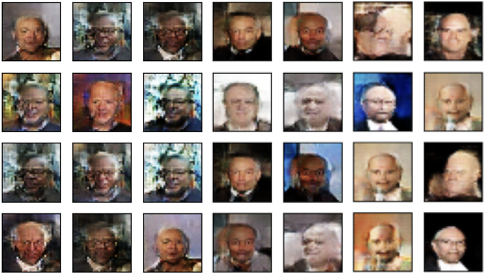} &
        \includegraphics[width=0.94\linewidth]{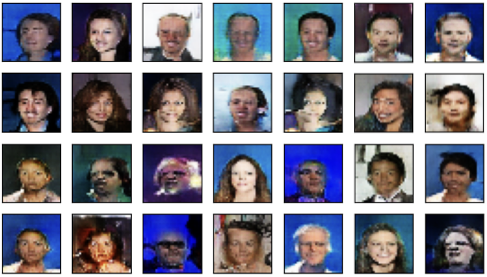} &
        \includegraphics[width=0.94\linewidth]{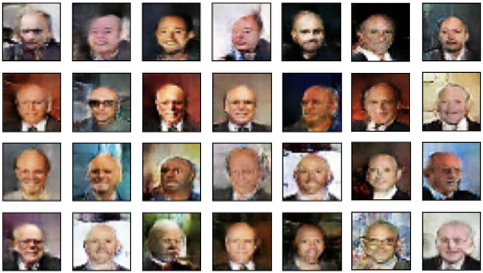} \\[-1pt]
    (d)  & (e)  & (f) \\[1pt]
        \includegraphics[width=0.94\linewidth]{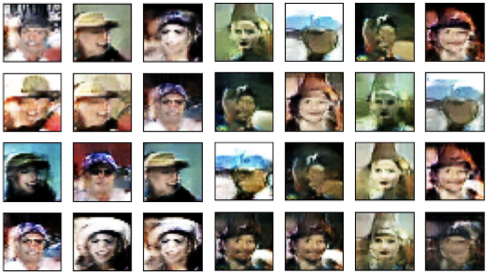} &
        \includegraphics[width=0.94\linewidth]{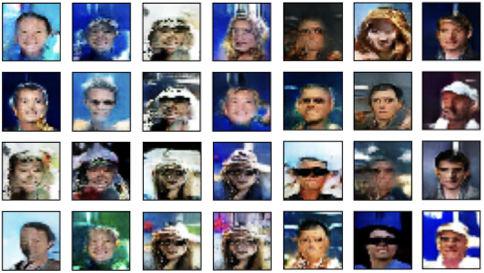} & 
        \includegraphics[width=0.94\linewidth]{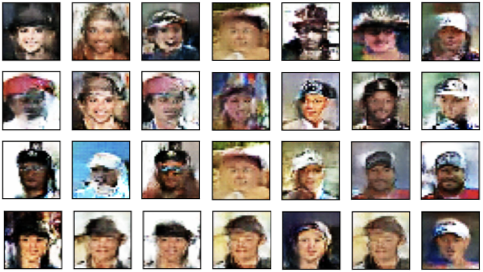} \\[-1pt]
    (g)  & (h)  & (i) \\[1pt]
      \end{tabular} 
  \caption[]{(\includegraphics[height=0.014\textheight]{Rgb.png} Color Online) A comparison of the  samples generated by various GANs on the CelebA dataset with positive samples drawn from: (a)-(c) the class of \textit{male} celebrities; (d)-(f) the class of \textit{bald} celebrities; (g)-(i) the class of celebrities wearing a \textit{hat}. On unbalanced data, ACGAN latched onto the majority classes, while Rumi-LSGAN generated samples from the correct class.}
  \vspace{-1.2em}
  \label{CelebA_Images}
  \end{center}
\end{sidewaysfigure}

The images generated by LSGAN, Rumi-LSGAN and ACGAN on CelebA are given in Figure~\ref{CelebA_Images}. Rumi-LSGAN generates visually better images than the baseline LSGAN and ACGAN in all the three cases considered. In the case on unbalanced data, the ACGAN latches on to the more-represented negative class. whereas Rumi-LSGAN is able to generate images exclusively from the target positive class. The FID and PR curves are presented in Figures~\ref{CIFAR10_CelebA_FID_PR}(c)-(f). In the case of unbalanced data, although ACGAN has a comparable PR score as that of Rumi-LSGAN, a majority of the samples generated correspond to the undesired (negative) class --- the positive/negative class label is something that the embeddings used to evaluate the PR measure are oblivious to. We attribute the relatively poorer PR performance of all models on the CelebA {\it Bald} dataset to insufficient reference positive class samples, resulting in poorer estimates of the model statistics.  \par
\begin{figure}[b]
\begin{center}
  \begin{tabular}[b]{P{0.03\linewidth}||P{0.89\linewidth}}
    \multirow{1}{*}[11.9em]{\rotatebox{90}{CelebA ({\it Females} class)}} & 
     \includegraphics[width=1\linewidth]{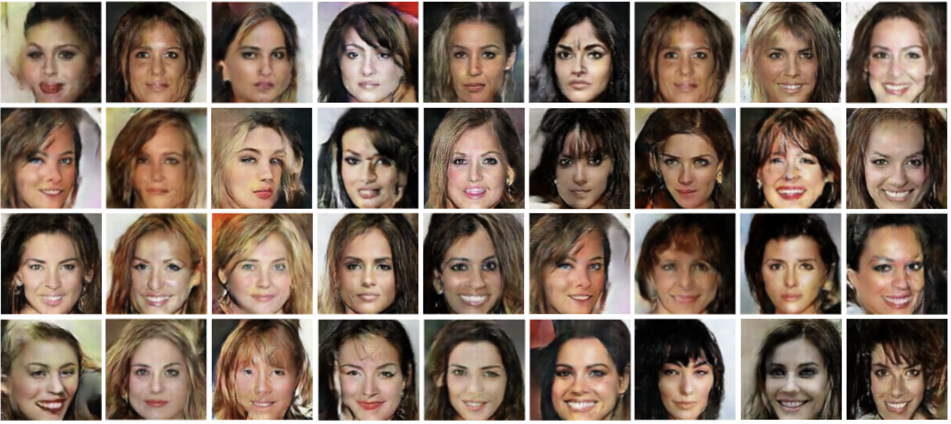}  \\[1pt] \hline
     \multirow{1}{*}[11.3em]{\rotatebox{90}{CelebA ({\it Males} class)}} & 
     \includegraphics[width=1\linewidth]{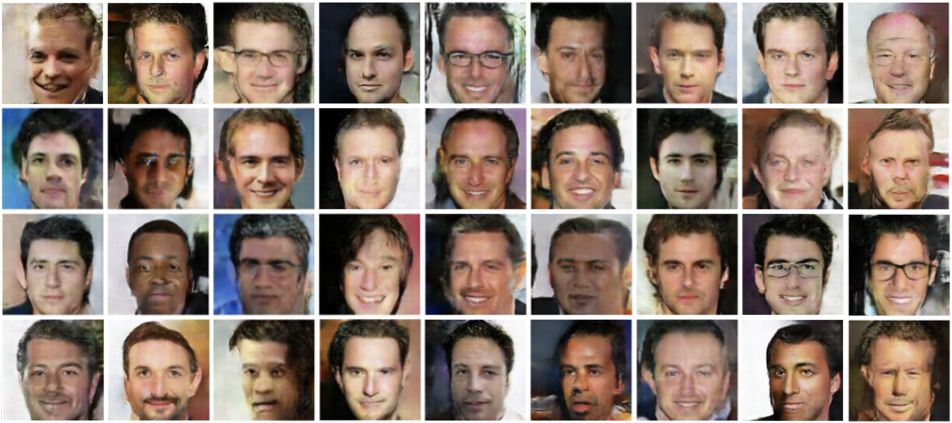}  \\[1pt] \hline
     \multirow{1}{*}[11.2em]{\rotatebox{90}{CelebA ({\it Hat} class)}} &
     \includegraphics[width=1\linewidth]{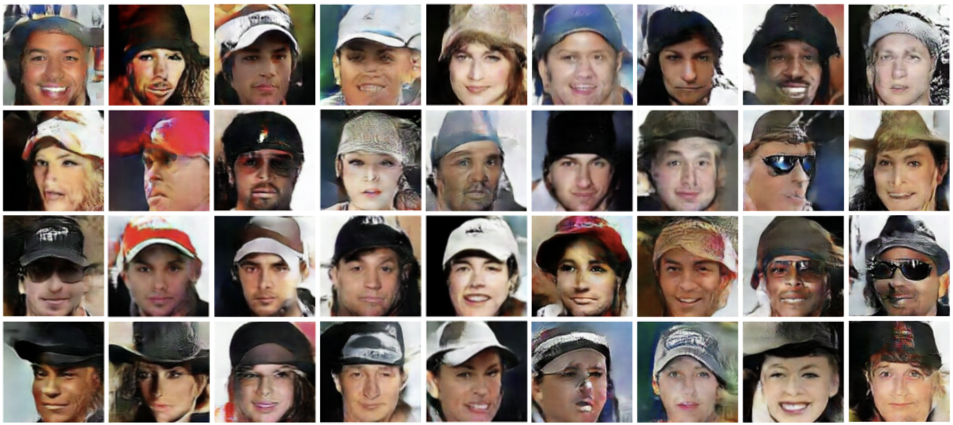} \\[1pt] \hline
      \multirow{1}{*}[11.3em]{\rotatebox{90}{CelebA ({\it Bald} class)}} &
     \includegraphics[width=1\linewidth]{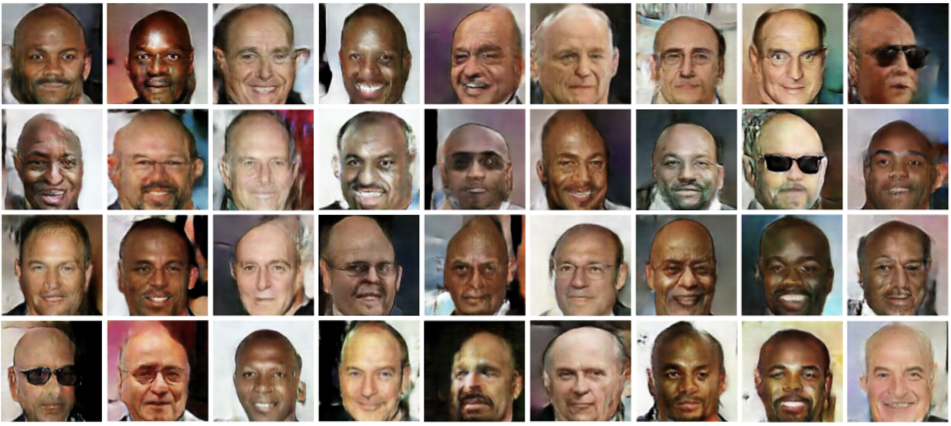} \\[1pt]
      \end{tabular} 
  \caption[]{(\includegraphics[height=0.014\textheight]{Rgb.png} Color Online) High-resolution CelebA images generated by Rumi-LSGAN.} 
  \vspace{-1.5em}
  \label{CelebA_HighRes}
  \end{center}
\end{figure}

 Finally, we show results on high-resolution CelebA images~\((128\times 128 \times 3)\). We train Rumi-LSGAN on both the simulated unbalanced class data ({\it Females}/{\it Males} classes with \(5\%\) positive samples and the other class negative) and true unbalanced classes ({\it Bald} and {\it Hat} classes). From the results shown in Figure~\ref{CelebA_HighRes}, we observe that Rumi-LSGAN generalizes well to the high-resolution scenario as well, generating a diverse set of images from the desired positive class in all cases.

\section*{Source Code}
The TensorFlow source code and models for all the experiments reported in this paper are available at the following GitHub Repository:~\url{https://github.com/DarthSid95/RumiGANs.git}.

 \vspace{8em}
\newpage

\newpage
\newpage

\end{document}